\documentclass{article} % For LaTeX2e
\usepackage{iclr2022_conference,times}

\usepackage{amsmath,amsfonts,bm}

\def\eqref#1{equation~\ref{#1}}
\def\1{\bm{1}}

\DeclareMathAlphabet{\mathsfit}{\encodingdefault}{\sfdefault}{m}{sl}
\SetMathAlphabet{\mathsfit}{bold}{\encodingdefault}{\sfdefault}{bx}{n}

\newcommand{\R}{\mathbb{R}}
\newcommand{\V}{\mathbb{V}}
\newcommand{\PP}{\mathbb{P}}

\DeclareMathOperator*{\argmin}{arg\,min}

\newcommand{\commentout}[1]{}
\newcommand*{\gauss}{\mathsf{gauss}}
\newcommand*{\cluster}{\mathsf{cluster}}
\newcommand*{\hierl}{\mathsf{hierl}}
\newcommand*{\hierd}{\mathsf{hierd}}
\newcommand*{\rbk}{\mathsf{rb}}
\newcommand*{\milky}{\mathsf{milky}}
\newcommand*{\milkyo}{\mathsf{milkyo}}

\newcommand*{\ext}[1]{\mathsf{ext{#1}}}
\newcommand*{\mulmod}{\mathsf{mulmod}}

\newcommand*{\nl}{\mathrm{nl}}
\newcommand*{\mm}{\mathrm{mm}}

\newcommand*{\cD}{\mathcal{D}}

\newcommand*{\code}[1]{{\mathtt{#1}}}

\newcommand*{\nn}{\mathit{nn}}

\newcommand*{\sample}{\mathrm{sa}}
\newcommand*{\observe}{\mathrm{ob}}
\newcommand*{\cond}{\mathrm{if}}

\newcommand*{\decode}{\mathrm{de}}
\newcommand*{\assign}{{:=}}
\newcommand*{\integral}{\mathrm{intg}}

\newcommand*{\defeq}{=}

\newcommand*{\db}[1]{\ensuremath{\llbracket #1 \rrbracket}}

\newcommand*{\KL}{\mathrm{KL}}

\DeclareMathOperator*{\EE}{\mathbb{E}}

\newcommand*{\postinfer}{\textsc{infer}}

\usepackage[utf8]{inputenc} % allow utf-8 input
\usepackage[T1]{fontenc}    % use 8-bit T1 fonts
\usepackage{hyperref}       % hyperlinks
\usepackage{url}            % simple URL typesetting
\usepackage{graphicx}
\usepackage{natbib}
\graphicspath{ {./figures/} }
\usepackage{booktabs}       % professional-quality tables
\usepackage{amsfonts}       % blackboard math symbols
\usepackage{nicefrac}       % compact symbols for 1/2, etc.
\usepackage{microtype}      % microtypography
\usepackage{xcolor}         % colors
\usepackage{caption,subcaption}
\usepackage{pgffor}
\usepackage{cleveref}
\usepackage{mathtools}
\usepackage{stmaryrd}
\usepackage{bbm}
\usepackage{paralist}
\usepackage{proof}
\usepackage{tikz}
\usetikzlibrary{bayesnet}
\usetikzlibrary{arrows}
\usepackage{multirow}
\usepackage{adjustbox}

\allowdisplaybreaks

\title{Meta-Learning an Inference Algorithm for Probabilistic Programs}

\author{%
	Gwonsoo Che \\
	KAIST \\
	Daejeon, Korea \\
	\texttt{gche@kaist.ac.kr} \\
	\And
	Hongseok Yang \\
	KAIST \\
	Daejeon, Korea \\
	\texttt{hongseok.yang@kaist.ac.kr} \\
}

\iclrfinalcopy % Uncomment for camera-ready version, but NOT for submission.
\begin{document}

\maketitle

\begin{abstract}
We present a meta-algorithm for learning a posterior-inference algorithm for restricted probabilistic programs. Our meta-algorithm takes a training set of probabilistic programs that describe models with observations, and attempts to learn an efficient method for inferring the posterior of a similar program.
A key feature of our approach is the use of what we call a white-box inference algorithm that extracts information directly from model descriptions themselves, given as programs. Concretely, our white-box inference algorithm is equipped with multiple neural networks, one for each type of atomic command, and computes an approximate posterior of a given probabilistic program by analysing individual atomic commands in the program using these networks. The parameters of the networks are learnt from a training set by our meta-algorithm. We empirically demonstrate that the learnt inference algorithm generalises well to programs that are new in terms of both parameters and model structures,
%\gc{Structure here includes two: dependency and the position of a nonlinear function. I know the term ``dependency structure'' here is for concreteness, but hope it will not confuse the reader. \hy{I changed it to model structure}}
and report cases where our approach %may be preferable to a state-of-the-art inference algorithm such as HMC.
achieves greater test-time efficiency than alternative approaches such as HMC.
%has advantages over alternative approaches such as HMC in terms of test-time efficiency.
The overall results show the promise as well as remaining challenges of our approach.
\end{abstract}

\section{Introduction}
\label{sec:intro}

One key objective of probabilistic programming is to automate reasoning about probabilistic models from diverse %problem 
domains~\citep{ritchie2015controlling,perov2016automatic,baydin2019efficient,%lew2020leveraging,
schaechtle2016time,cusumano2017probabilistic,saad2016probabilistic,%ouyang2018webppl,
kulkarni2015picture,young2019reconstruction,jager2020inference}. As a way to realize this goal, researchers have extensively worked on the development of posterior-inference or parameter-learning algorithms that are efficient and universal; the algorithms can be applied to all or nearly all models written in probabilistic programming languages (PPLs). This line of research has led to performant probabilistic programming systems~\citep{goodman2008church,wood-aistats-2014,%tolpin2016design,
mansinghka2014venture,InferNET18, narayanan2016probabilistic,salvatier2016probabilistic,carpenter2017stan,tran2016edward,ge2018turing,bingham2018pyro}. Yet, it also revealed the difficulty of achieving efficiency and universality simultaneously, and the need for equipping PPLs with mechanisms for customising inference or learning algorithms to a given %problem 
domain. In fact, recent PPLs include constructs for specifying conditional independence in a model~\citep{bingham2018pyro} or defining proposal or variational distributions \citep{ritchie2015controlling,siddharth2017learning,bingham2018pyro,tran2018edward2,cusumano-towner2019gen}, all enabling users to help inference or learning algorithms.

In this paper, we explore a different approach. We present a meta-algorithm for learning a posterior-inference algorithm itself from a given set of restricted probabilistic programs, which specifies a class of probabilistic models, such as hierarchical or clustering models.
The meta-algorithm aims at constructing a
customised inference algorithm for the given set of models, while ensuring universality to the extent
that the constructed algorithm can generalise: it works well for models not in the training set, as long as the models are similar to the ones in the set.
%\gc{We talk about customisation to a model class, but at the same time we will show extrapolation empirically as a benefit. It may be better to refine or improve our arguments slightly in this respect. 2021/05/19: our inference algorithm gets customised, but it also pursues universality to such an extent that, for some cases, it extrapolates to similar but different/unseen program structures. Maybe we want to
%change the flow from 100\% customisation to 80\% customisation plus 20\% universality (extrapolation).}

%\begin{figure}
%\begin{align*}
%&
%z \sim \mathcal{N}(0,5);
%\\
%&
%\code{obs}(\mathcal{N}(z,1),2) 
%\ \, \texttt{/\!\!/}\  \text{Observation $x=2$ for $x \sim \mathcal{N}(z,1)$}
%\end{align*}
%\caption{Probabilistic program for a simple Gaussian model. The second command expresses that an (un-named) random variable $x$ is drawn from $\mathcal{N}(z,1)$ and its value is observed to be $2$.}
%\label{fig:simple-example}
%\end{figure}

The distinguished feature of our approach is the use of what we call a white-box inference algorithm, which extracts information directly from model descriptions themselves, given as programs in a PPL. Concretely, our white-box inference algorithm is equipped with multiple neural networks, one for each type of atomic command in a PPL, and computes an approximate posterior for a given program by analysing (or executing in a sense) individual atomic commands in it using these networks. For instance, given the probabilistic program in Fig.~\ref{fig:example-galaxy}, which describes a simple model on the Milky Way galaxy, the white-box inference algorithm analyses the program as if an RNN handles a sequence or an interpreter executes a program. Roughly, the algorithm regards the program as a sequence of the five atomic commands (separated by the ``;'' symbol), initialises its internal state $h \in \R^m$ with $h_0$, and transforms the state over the sequence. The internal state $h$ is the encoding of an approximate posterior at the current program point, which corresponds to an approximate filtering distribution of a state-space model. How to update this state over each atomic command is directed by neural networks. 
%A useful analogy is to understand this neural network as a non-standard interpreter of probabilistic programs, which interprets each atomic command as an update operation on the internal-state vector $h$. 
Our meta-algorithm trains the parameters of these networks by trying to make the inference algorithm compute accurate posterior approximations over a training set of probabilistic programs. 
%The reader may find it useful to 
One can also view our white-box inference algorithm as a message-passing algorithm in a broad sense where transforming the internal state $h$ corresponds to passing a message, and understand our meta-algorithm as a method for learning how to pass a message for each type of atomic commands.

\begin{figure}
\hrule
\vspace{-2.5mm}
\begin{align*}
& \mathit{mass} \sim \mathcal{N}(5,10); \ \texttt{/\!\!/}\ \text{log of the mass of Milky Way}
        \\[-0.4ex]
& g_1 \sim \mathcal{N}(\mathit{mass} \,{\times}\, 2,5);\,
\code{obs}(\mathcal{N}(g_1,1),10);   \texttt{/\!\!/}\ \text{observed velocity $\mathit{vel}_1{=}10$ of the first satellite galaxy}
        \\[-0.4ex]
& g_2 \sim \mathcal{N}(\mathit{mass}\,{+}\,5,2); \, 
\code{obs}(\mathcal{N}(g_2,1), 3) \ \texttt{/\!\!/}\ \text{observed velocity $\mathit{vel}_2{=}3$ of the second satellite galaxy}
\end{align*}
\vspace{-4.5mm}
\hrule
\vspace{1mm}
\caption{Probabilistic program for a model for Milky Way and its two satellite galaxies. The $\code{obs}$ statements refer to the observations of (unnamed) random variables $\mathit{vel}_1$ and $\mathit{vel}_2$.}
\label{fig:example-galaxy}
\vspace{-7mm}
\end{figure}

This way of exploiting model descriptions for posterior inference has two benefits. First, it ensures that even after customisation through the neural-network training, the inference algorithm does not lose its universality and can be applied to any probabilistic programs. Thus, at least in principle, the algorithm has a possibility to generalise beyond the training set; its accuracy degrades gracefully as the input probabilistic program diverges from those in the training set. Second, our way of using model descriptions guarantees the efficiency of the inference algorithm (although it does not guarantee the accuracy). The algorithm scans the input program only once, and uses neural networks whose input dimensions are linear in the size of the program. As a result, its time complexity is quadratic over the size of the input program. Of course, the guaranteed speed also indicates that the customisation of the algorithm for a given training set, whose main goal is to achieve good accuracy for probabilistic programs in the set, is a non-trivial process.

%Our white-box inference algorithm relies on a neural network, but does not entirely so. For some cases, it uses prior knowledge about how each program statement transforms the marginal likelihood of observations processed so far. Judiciously combining the flexibility of a neural network with prior knowledge on inference enables the efficient learning of good neural-network parameters of the algorithm. 

%We empirically evaluate our approach on classes of probabilistic models expressed as probabilistic programs, and describe the promise and remaining challenges revealed by the evaluation. 

Our contributions are as follows: (i) we present a white-box posterior-inference algorithm, which works directly on model description and can be customised to a given model class; (ii) we describe a meta-algorithm for learning the parameters of the inference algorithm; (iii) we empirically analyse our approach with different model classes, and show the promise as well as the remaining challenges.

%\hy{Add more contextual information. Isn't it just another message-passing algorithm? We should provide some answer for it. The intro does not talk much about the importance and the benefit of our approach. It just attempts to describe the approach. Fix this.}
%\gc{Diff-1: Message passing algorithms compute the local quantities, e.g., marginal posteriors, directly, while our inference algorithm computes internal states that encode the local quantities. Ours is generalizable to a set of problems, but message passing algorithms are usually problem-specific.}

\noindent{\bf Related work}\ \
The difficulty of developing an effective posterior-inference algorithm is well-known, and has motivated active research on learning or adapting key components of an inference algorithm. Techniques for adjusting an MCMC proposal~\citep{andrieu08adaptive} or an HMC integrator~\citep{hoffman14pnut} to a given inference task were %developed and 
implemented in popular tools. Recently, methods for meta-learning these techniques themselves from a collection of inference tasks have been developed~\citep{wang18metalearning,gong19metalearning}. The meta-learning approach also features in the work on stochastic variational inference where a variational distribution receives information about each inference task in the form of its dataset of observations and is trained with a collection of datasets~\citep{wu2020meta,gordon19metalearning,iakovleva20metalearning}. For a message-passing-style variational-inference algorithm, such as expectation propagation~\citep{minka01,wainwright08vibook}, \citet{jitkrittum15ep} studied the problem of learning a mechanism to pass a message for a given \emph{single} inference task. A natural follow-up question is how to meta-learn such a mechanism from a dataset of \emph{multiple} inference tasks that can generalise to \emph{unseen} models. Our approach provides a partial answer to the question; our white-box inference algorithm can be viewed as a message-passing-style variational inference algorithm that can meta-learn the representation of messages and a mechanism for passing them for given probabilistic programs. 
%a given dataset of probabilistic programs. 

Amortised inference and inference compilation
~\citep{gershman2014amortized,le2017inference,paige2016inference,stuhlmuller2013learning,kingma2013auto,mnih2014neural,rezende2014stochastic,ritchie2016deep,marino2018iterative}
are closely related to our approach in that they also attempt to learn a form of a posterior-inference algorithm.
However, the learnt algorithm by them and that by ours have different scopes. The former is designed to work for 
unseen inputs or observations of a \emph{single} model, while the latter %is designed to work 
for \emph{multiple} 
models with different structures. The relationship between these two algorithms is similar to the one
between a compiled program (to be applied to multiple inputs) and a compiler (to be used for multiple programs).

The idea of running programs with learnt neural networks also appears in the work on training neural networks to execute programs~\citep{zaremba14execute,bieber20executeprograms,reed15npi}. As far as we know, however, we are the first to frame the problem of learning a posterior-inference algorithm as the one of learning to execute.% (in a non-standard execution model). 
%One interesting future research direction is to use the recent findings in this line of research about handling complex control-flow structures~\citep{bieber20executeprograms} and extend our results to a more expressive language.

%Such a non-standard notion of execution appears commonly in the work on static program analysis~\citep{cc:popl:77}, and has also been used to train a robust classifier based on neural networks~\citep{mirman18icml}.

\commentout{
\section{Conclusion}

In the paper, we presented a new posterior-inference algorithm for probabilistic programs, which computes approximate posterior and marginal likelihood estimate by executing a given program in a non-standard way using a collection of trainable neural networks. This non-standard execution exploits the structure of the program, such as the sequencing of atomic commands in the program and the types of those commands. We described a meta-algorithm for learning the parameters of the neural networks from a training set of probabilistic programs. Our experimental results show the promise as well as the remaining challenges of our approach. 

One high-level message of this work is that the very description of a typical statistical model contains a large amount of information, and learning how to extract this information and then exploiting the extracted information may lead to efficient inference on the model. We hope that our work gives the reader an idea about how to instantiate this high-level idea concretely, and encourages further exploration of this research direction.
}

\section{Setup}
\label{sec:setup}

\begin{figure}[t]
\hrule
\vspace{-2.5mm}
\begin{align*}
& \mathit{u} \,{:=}\, 0;\, \mathit{v} \,{:=}\, 5;\, \mathit{w} \,{:=}\, 1;\,
\mathit{z_1} \,{\sim}\, \mathcal{N}(\mathit{u},v); \, \mathit{z_2} \,{\sim}\, \mathcal{N}(\mathit{u},v); 
\\[-0.3ex]
& \mathit{z_3} \,{\sim}\, \mathcal{N}(\mathit{u},w); \ \mathit{\mu_3} \,{:=}\, \code{if}\, (z_3 \,{>}\, \mathit{u})\, z_1\, \code{else}\, z_2;  \code{obs}(\mathcal{N}(\mu_3,w), -1.9); \, \texttt{/\!\!/}\ \text{$x_1  \,{\sim}\, \mathcal{N}(\mu_3,w)$, $x_1\,{=}\,{-1.9}$} 
\\[-0.3ex]
& \mathit{z_4} \,{\sim}\, \mathcal{N}(\mathit{u},w); \ \mathit{\mu_4} \,{:=}\, \code{if}\, (z_4 \,{>}\, \mathit{u})\, z_1\, \code{else}\, z_2;  \code{obs}(\mathcal{N}(\mu_4,w), -2.2); \, \texttt{/\!\!/}\ \text{$x_2  \,{\sim}\, \mathcal{N}(\mu_4,w)$, $x_2\,{=}\,{-2.2}$} 
\\[-0.3ex]
& \mathit{z_5} \,{\sim}\, \mathcal{N}(\mathit{u},w); \ \mathit{\mu_5} \,{:=}\, \code{if}\, (z_5 \,{>}\, \mathit{u})\, z_1\, \code{else}\, z_2; \code{obs}(\mathcal{N}(\mu_5,w), 2.4);\, \texttt{/\!\!/}\ \text{$x_3  \,{\sim}\, \mathcal{N}(\mu_5,w)$, $x_3\,{=}\,{2.4}$} 
\\[-0.3ex]
& \mathit{z_6} \,{\sim}\, \mathcal{N}(\mathit{u},w); \ \mathit{\mu_6} \,{:=}\, \code{if}\, (z_6 \,{>}\, \mathit{u})\, z_1\, \code{else}\, z_2; \code{obs}(\mathcal{N}(\mu_6,w), 2.2)\,  \texttt{/\!\!/}\ \text{$x_4  \,{\sim}\, \mathcal{N}(\mu_6,w)$, $x_4\,{=}\,{2.2}$}
\end{align*}
\vspace{-4.5mm}
\hrule
\vspace{1mm}
\caption{
	Probabilistic program for a simple clustering model on four data points.
}
\label{fig:example-mixture}
\vspace{-7mm}
\end{figure}

Our results assume a 
simple probabilistic programming language without loop and with a limited form of
conditional statement. The syntax of the language is given by the following grammar, 
where $r$ represents a real number, $z$ and $v_i$ variables storing a real, and $p$ the name of a procedure taking two real-valued parameters and returning a real number:
\[
\begin{array}{@{}r@{}l@{}}
        \textit{Programs}\ C & {} ::=
A \,\mid\,
C_1;C_2 
\\
        \textit{Atomic Commands} \ A & {} ::=
        z  \sim \mathcal{N}(v_1,v_2) \,\mid\, \code{obs}(\mathcal{N}(v_0,v_1),r) \,\mid\, v_0:=\code{if}\ (v_1 > v_2)\ v_3\ \code{else}\ v_4
\\
        & {} \ \,\,\mid\ \,\, v_0:= r \,\mid\, v_0:=v_1 \,\mid\, v_0:=p(v_1,v_2)
\end{array}
\]
%\begin{align*}
%        \textit{Programs}\ C & ::=
%A \,\mid\,
%C_1;C_2 
%\\[-0.3ex]
%        \textit{Atomic Commands} \ A & ::=
%        z  \sim \mathcal{N}(v_1,v_2) \,\mid\, \code{obs}(\mathcal{N}(v_0,v_1),r) \,\mid\, v_0:=\code{if}\ (v_1 > v_2)\ v_3\ \code{else}\ v_4
%\\[-0.3ex]
%        & {} \ \,\,\mid\ \,\, v_0:= r \,\mid\, v_0:=v_1 \,\mid\, v_0:=p(v_1,v_2)
%\end{align*}
%\begin{align*}
%\mbox{\it Real Expressions}\quad  E & ::= z \,\mid\, r
%\\
%\mbox{\it Distribution Constructors}\quad D & ::= \code{normal}
%\\
%\mbox{\it Atomic Commands}\quad  A & ::= 
%\code{skip}
%\\
%& \
%\,\mid\, z:=\code{assign\_vpc}()
%\\
%& \
%\,\mid\, z:=\code{assign\_vpv}()
%\\
%& \
%\,\mid\, z:=\code{assign\_vtc}()
%\\
%& \
%\,\mid\, z:=\code{assign\_vtv}()
%\\
%& \
%\,\mid\, z:=\code{assign\_sel}()
%\\
%& \ 
%\,\mid\, z:=\code{sample\_cc}()
%\\
%& \ 
%\,\mid\, z:=\code{sample\_vc}()
%\\
%& \ 
%\,\mid\, z:=\code{proc}()
%\\
%& \ 
%\,\mid\, z:=\code{observe}()
%\\
%\mbox{\it Commands}\quad  C & ::=  A \,\mid\, C;C \
%\end{align*}
Programs in the language are constructed by sequentially composing atomic commands. The language supports six types of \emph{atomic commands}. The first type is  $z \sim \mathcal{N}(v_1,v_2)$, which
draws a sample from the normal distribution with mean $v_1$ and variance $v_2$, and assigns the sampled
value to $z$. The second command, $\code{obs}(\mathcal{N}(v_0,v_1),r)$, states that a random variable is
drawn from $\mathcal{N}(v_0,v_1)$ and its value is observed to be $r$. The next 
is a restricted form of a conditional statement that selects one of $v_3$ and $v_4$ depending on the result
of the comparison $v_1 > v_2$. The following two commands are different kinds of assignments, one for assigning a constant and
the other for copying a value from one variable to another.
The last atomic command $v_0:=p(v_1,v_2)$ is a call to one of the known deterministic procedures, which may be standard binary operations such as addition and multiplication, or complex non-trivial functions that are used to build advanced, non-conventional models. When $p$ is a standard binary operation, we use the usual infix notation and write, for example, $v_1+v_2$, instead of $+(v_1,v_2)$.

%\begin{figure}[t]
%\begin{align*}
%& \mathit{m} := 0; \ \mathit{v} := 2; \ \mathit{w} := 1; \ \mathit{tr} = 0.7; \\
%& \mathit{z_1} \sim \mathcal{N}(m,v); \ \code{obs}(\mathcal{N}(z_1,w),0.5); \\
%& \mathit{y_2} := \mathit{tr}\times\mathit{z_1}; \ \mathit{z_2} \sim \mathcal{N}(y_2,v); \ \code{obs}(\mathcal{N}(z_2,w), 0.8); \\
%& \mathit{y_3} := \mathit{tr}\times\mathit{z_2}; \ \mathit{z_3} \sim \mathcal{N}(y_3,v); \ \code{obs}(\mathcal{N}(z_3,w), 1.3); \\
%& \mathit{y_4} := \mathit{tr}\times\mathit{z_3}; \ \mathit{z_4} \sim \mathcal{N}(y_4,v); \ \code{obs}(\mathcal{N}(z_4,w), 1.8)
%\end{align*}
%%\begin{verbatim}
%%m := 0;  v := 2;  w:= 1; c := 0.7;
%%z1 ~ N(m,v); obs(N(z1,w),0.5);
%%y2 := c*z1;  z2 ~ N(y2,v); obs(N(z2,w),0.8);
%%y3 := c*z2;  z3 ~ N(y3,v); obs(N(z3,w),1.3);
%%y4 := c*z3;  z4 ~ N(y4,v); obs(N(z4,w),1.8)
%%\end{verbatim}
%\caption{
%	Linear state-space model in our language. Models with general affine transitions and emmisions can be modelled using our language.
%	\hy{I am adding some example just to fill space. Should be revised.}
%}
%\label{fig:example-lssm}
%\end{figure}

We permit only the programs where a variable does not appear more than once on the left-hand side of the $:=$ and $\sim$ symbols. This means that no variable is updated twice or more, and it corresponds to the so-called static single assignment assumption in the work on compilers. This restriction lets us regard variables updated by $\sim$ as latent random variables. We denote those variables by $z_1,\ldots,z_n$. 

%Note that the language restricts the syntactic forms of atomic commands. For instance, it requires that the arguments to a normal distribution or to a procedure $p$ should be variables, not constants or addition of two variables.
We use this simple language for two reasons. First, the restriction imposed on our language enables the simple definition of our white-box inference algorithm. %For instance, 
The language supports only a limited form of conditional statements and restricts the syntactic forms of atomic commands; the arguments to a normal distribution or to a procedure $p$ should be variables, not general expression forms such as addition of two variables.
As we will show soon, this restriction makes it easy to exploit information about the type of each atomic command in our  inference algorithm; we use different neural networks for different types of atomic commands in the algorithm.
Second, the language is intended to serve as an intermediate language of a compiler for a high-level PPL, not the one to be used directly by the end user. The compilation scheme in, for instance, \S3 of \citep{van2018introduction}
%\hy{Add the section number of the cited book, as we did in our NeurIPS response}
from high-level probabilistic programs with general conditional statements and for loops to graphical models can be adopted to compile such programs into our language. See Appendix~\ref{appendix:ppl-translation} for further discussion.

Fig.~\ref{fig:example-mixture} shows a simple model for clustering four data points $\{-1.9,-2.2,2.4,2.2\}$ into two clusters, where the cluster assignment of each data point is decided by thresholding a sample from the standard normal distribution. The variables $z_1$ and $z_2$ store the centers of the two clusters, and $z_3,\ldots,z_6$ hold the random draws that decide cluster assignments for the data points. 
See Appendix~\ref{appendix:galaxy-compiled} for the Milky Way example in Fig.~\ref{fig:example-galaxy} compiled to a program
in our language.
%Fig.~\ref{fig:example-galaxy-compiled} in the appendix shows the Milky Way example in the language.
%Fig.~$8$ in the supplementary material shows the Milky Way example in the language.
%The Milky Way example in Fig.~\ref{fig:example-galaxy} becomes another example program of the language if complex expressions in it get compiled to sequences of atomic commands.
%The outcome of this compilation appears in Fig.~\ref{fig:example-galaxy-compiled} of the appendix.

Probabilistic programs in the language denote unnormalised probability densities over $\R^n$ for some $n$. Specifically, for a program $C$, if $z_1,\ldots,z_n$ are all the variables assigned by the sampling statements $z_i \sim \mathcal{N}(\ldots)$ in $C$ in that order and  $C$ contains $m$ observe statements with observations $r_1,\ldots,r_m$, then $C$ denotes an unnormalised density $p_C$ over the real-valued random variables $z_1,\ldots,z_n$:
$p_C(z_{1:n}) = p_C(x_{1:m}\,{=}\, r_{1:m} | z_{1:n}) \times \prod_{i = 1}^n p_C(z_i | z_{1:i-1})$,
%\begin{align*}
%       p_C(z_{1:n}) =
%        p_C(x_{1:m}\,{=}\, r_{1:m} | z_{1:n}) \times
%        \prod_{i = 1}^n p_C(z_i | z_{1:i-1}),
%\end{align*}
where $x_1,\ldots,x_m$ are variables not appearing in $C$ and are used to denote observed variables.
This density is defined inductively over the structure of $C$. See 
%the supplementary material for details.
Appendix~\ref{appendix:semantics} for details.
The goal of our white-box inference algorithm is to compute efficiently accurate approximate posterior and marginal likelihood estimate for a given $C$ (that is, for the normalised version of $p_C$ and the normalising constant of $p_C$), when $p_C$ has a finite non-zero marginal likelihood and, as a result, a well-defined posterior density. We next describe how the algorithm attempts to achieve this goal.

\section{White-box inference algorithm}
\label{sec:algorithm}

Given a program $C=(A_1;\ldots;A_k)$, our white-box inference algorithm views $C$ as a sequence of its constituent atomic commands $(A_1,A_2,\ldots,A_k)$, and computes an approximate posterior and a marginal likelihood estimate for $C$ by sequentially processing the $A_i$'s. Concretely, the algorithm starts by initialising its internal state to $h_0 = \vec{0} \in \R^s$ and the current marginal-likelihood estimate to $Z_0=1$. Then, it updates these two components based on the first atomic command $A_1$ of $C$. It picks a neural network appropriate for the type of $A_1$, applies it to $h_0$ and gets a new state $h_1 \in \R^s$. Also, it updates the marginal likelihood estimate to $Z_1$ by analysing the semantics of $A_1$. This process is repeated for the remaining atomic commands $A_2,A_3,\ldots,A_k$ of $C$, and eventually produces the last state $h_k$ and estimate $Z_k$. Finally, the state $h_k$ gets decoded to a probability density on the latent variables of $C$ by a neural network, which together with $Z_k$ becomes the result of the algorithm. 
%%%A good way to understand our algorithm is to view our algorithm as a version of the message-passing algorithm, which receives a message $h_{i-1}$ and sends a transformed message $h_i$ at each $A_i$; or to understand the algorithm as a variant of the forward filtering algorithm for state-space models, where $h_i$ encodes information about the approximate filtering distribution.

Formally, our inference algorithm 
%Our white-box inference algorithm 
is built on top of three kinds of neural networks: the ones for transforming the internal state $h \in \R^s$ of the algorithm; a neural network for decoding the internal states $h$ to probability densities; and the last neural network for approximately solving integration questions that arise from the marginal likelihood computation in observe statements. We present these neural networks for the programs that sample $n$-many latent variables $z_1,\ldots,z_n$, and use at most $m$-many variables (so $m \geq n$). Let $\V$ be $[0,1]^m$, the space of the one-hot encodings of those $m$ variables, and $\PP$ the set of procedure names. Our algorithm uses the following neural networks:
\[
        \begin{array}{r@{}lr@{}lr@{}l}
        \nn_{\sample,\phi_1} & \,{:}\, \V^3 \,{\times}\, \R^s \,{\to}\, \R^s,
        &
        \nn_{\observe,\phi_2} & \,{:}\, \V^2 \,{\times}\, \R \,{\times}\, \R^s \,{\to}\, \R^s,
        &
        \nn_{\cond,\phi_3} & \,{:}\, \V^5 \,{\times}\, \R^s \,{\to}\, \R^s,
        \\
        \nn^\mathrm{c}_{\assign,\phi_4} & \,{:}\, \V \,{\times}\, \R \,{\times}\, \R^s \,{\to}\, \R^s,
        &
        \nn^\mathrm{v}_{\assign,\phi_5} & \,{:}\, \V^2 \,{\times}\, \R^s \,{\to}\, \R^s,
        &
        \nn_{p,\phi_p} & \,{:}\, \V^3 \,{\times}\, \R^s \,{\to}\, \R^s \ \text{for $p \in \PP$},
        \\
        \nn_{\decode,\phi_6} & \,{:}\, \R^s \,{\to}\, (\R \,{\times}\, \R)^n,
        &
        \nn_{\integral,\phi_7} & \,{:}\, \V^2 \,{\times}\, \R \,{\times}\, \R^s \,{\to}\, \R,
        \end{array}
\]
%\begin{align*}
%        \nn_{\sample,\phi_1} & \,{:}\, \V^3 \,{\times}\, \R^s \,{\to}\, \R^s,
%        &
%        \nn_{\observe,\phi_2} & \,{:}\, \V^2 \,{\times}\, \R \,{\times}\, \R^s \,{\to}\, \R^s,
%        &
%        \nn_{\cond,\phi_3} & \,{:}\, \V^5 \,{\times}\, \R^s \,{\to}\, \R^s,
%        \\[-0.3ex]
%        \nn^\mathrm{c}_{\assign,\phi_4} & \,{:}\, \V \,{\times}\, \R \,{\times}\, \R^s \,{\to}\, \R^s,
%        &
%        \nn^\mathrm{v}_{\assign,\phi_5} & \,{:}\, \V^2 \,{\times}\, \R^s \,{\to}\, \R^s,
%        &
%        \nn_{p,\phi_p} & \,{:}\, \V^3 \,{\times}\, \R^s \,{\to}\, \R^s \ \text{for $p \in \PP$},
%        \\[-0.3ex]
%        \nn_{\decode,\phi_6} & \,{:}\, \R^s \,{\to}\, (\R \,{\times}\, \R)^n,
%        &
%        \nn_{\integral,\phi_7} & \,{:}\, \V^2 \,{\times}\, \R \,{\times}\, \R^s \,{\to}\, \R,
%\end{align*}
where $\phi_{1:7}$ and $\phi_p$ for $p \in \PP$ are network parameters. The top six networks are for the six types of atomic commands in our language. For instance, when an atomic command to analyse next is a sample statement $z \sim \mathcal{N}(v_1,v_2)$, the algorithm runs the first network $\nn_\sample$ on the current internal state $h$, and obtains a new state $h' = \nn_{\sample,\phi_1}(\overline{z}, \overline{v_{1:2}},h)$, where $\overline{z}$ and $\overline{v_{1:2}}$ mean the one-hot encoded variables $z$, $v_1$ and $v_2$. The next $\nn_{\decode,\phi_6}$ is a decoder of the states $h$ to probability densities over the latent variables $z_1,\ldots,z_n$, which are the product of $n$ independent normal distributions. The network maps $h$ to the means and variances of these distributions. The last $\nn_{\integral,\phi_7}$ is used when our algorithm updates the marginal likelihood estimate based on an observe statement $\code{obs}(\mathcal{N}(v_0,v_1),r)$. When we write the meaning of this observe statement as the likelihood $\mathcal{N}(r ; v_0,v_1)$, and the filtering distribution for $v_0$ and $v_1$ under (the decoded density of) the current state $h$ as $p_h(v_0,v_1)$,\footnote{The $p_h(v_0,v_1)$ is a filtering distribution, not prior.} 
%Also, to convey the intuition, we assume here that the marginal distribution has a density, but in practice, this assumption does not necessarily hold, and our algorithm does not rely on the assumption.} 
the last neural network computes the following approximation:
$\nn_{\integral,\phi_7}(\overline{v_{0:1}},r,h) \approx \int \mathcal{N}(r ; v_0,v_1) p_h(v_0,v_1) dv_0 dv_1$.
%\[
%\nn_{\integral,\phi}(\overline{v_{0:1}},r,h) \approx \int \mathcal{N}(r ; v_0,v_1) p_h(v_0,v_1) dv_0 dv_1.
%\]
See Appendix~\ref{appendix:marginal-likelihood} for the full derivation of the marginal likelihood.

Given a program $C = (A_1;\ldots;A_k)$ that draws $n$ samples (and so uses latent variables $z_1,\ldots,z_n$), 
the algorithm approximates the posterior and marginal likelihood of $C$ as follows:
\[
\begin{array}{r@{}l}
 \postinfer(C) \;=\; {} & \textbf{let}\ (h_0,Z_0) = (\vec{0},1)\ \textbf{and}\ 
(h_k,Z_k) = (\postinfer(A_k) \circ \ldots \circ \postinfer(A_1))(h_0,Z_0)\ \textbf{in}
        \\
& \textbf{let}\ ((\mu_1,\sigma^2_1), \ldots, (\mu_n,\sigma^2_n)) = \nn_{\decode,\phi_6}(h_k)\ \textbf{in}\
    %\left(
    \textbf{return}\ \Big(\prod_{i = 1}^n \mathcal{N}(z_i \mid \mu_i,\sigma^2_i),\; Z_k\Big),
    %\right),
\end{array}
\]
%\begin{align*}
% \postinfer(C) \;=\; {} & \textbf{let}\ (h_0,Z_0) = (\vec{0},1)\ \textbf{and}\ 
%(h_k,Z_k) = (\postinfer(A_k) \circ \ldots \circ \postinfer(A_1))(h_0,Z_0)\ \textbf{in}
%        \\[-0.3ex]
%& \textbf{let}\ ((\mu_1,\sigma^2_1), \ldots, (\mu_n,\sigma^2_n)) = \nn_{\decode,\phi_9}(h_k)\ \textbf{in}\
%    %\left(
%    \textbf{return}\ \Big(\prod_{i = 1}^n \mathcal{N}(z_i \mid \mu_i,\sigma^2_i),\; Z_k\Big),
%    %\right),
%\end{align*}
where $\postinfer(A_i):  \R^s \times \R \to \R^s \times \R$ picks an appropriate neural network based on the type of $A_i$, and uses it to transform $h$ and $Z$:
\begin{align*}
&
\postinfer(\code{obs}(\mathcal{N}(v_0,v_1),r))(h,Z)
= 
(\nn_{\observe}(\overline{v_{0:1}},r,h),
Z\,{\times}\, \nn_{\integral}(\overline{v_{0:1}},r,h)),
\\[-0.2ex]
&
\postinfer(v_0 \,{:=}\, \code{if}\  (v_1 \,{>}\, v_2)\ v_3\ \code{else}\ v_4)(h,Z) 
 =
(\nn_{\cond}(\overline{v_{0:4}},h),Z),
\\[-0.2ex]
&
\postinfer(v_0 \,{:=}\, r)(h,Z) 
= 
(\nn^\mathrm{c}_{\assign}(\overline{v_0},r,h),Z),
\ 
\postinfer(z \,{\sim}\, \mathcal{N}(v_1,v_2))(h,Z)
=
(\nn_{\sample}(\overline{z},\overline{v_{1:2}},h),Z),
\\[-0.2ex]
& \postinfer(v_0 \,{:=}\, v_1)(h,Z) 
=
(\nn^\mathrm{v}_{\assign}(\overline{v_{0:1}},h),Z),
\ 
\postinfer(v_0 \,{:=}\, p(v_1,v_2))(h,Z) 
=
(\nn_p(\overline{v_{0:2}},h),Z).
\end{align*}
We remind the reader that $\overline{v_{0:k}}$ refers to the sequence of the one-hot encodings of variables $v_0,\ldots,v_k$. For the update of the state $h$, the subroutine $\postinfer(A)$ relies on neural networks. But for the computation of the marginal likelihood estimate, it exploits prior knowledge that non-observe commands do not change the marginal likelihood (except only indirectly by changing the filtering distribution), and keeps the input $Z$ for those atomic commands. 
% Hongseok: The next sentence describes what we hoped but couldn't justify experimentally
%Our experimental evaluation shows the benefit of approximating the posterior and the marginal likelihood simultaneously and exploiting prior knowledge just mentioned, when we train the neural networks used by our algorithm. \hy{Make sure that the last sentence tells the truth.}

\section{Meta-learning parameters}
\label{sec:learning}

The parameters of our white-box inference algorithm are learnt from a collection of probabilistic programs in our language. Assume that we are given a training set of programs $\cD = \{C_1,\ldots,C_N\}$ such that each $C_i$ samples $n$ latent variables $z_1,\ldots,z_n$ and uses at most $m$ variables. Let $\phi = (\phi_{1:7}, (\phi_p)_{p \in \PP})$ be the parameters of all the neural networks used in the algorithm. We learn these parameters by solving the following optimisation problem:\footnote{Strictly speaking, we assume that the marginal likelihood of any $C \in \cD$ is non-zero and finite.}
\begin{equation*}
\argmin_\phi \sum_{C \in \cD} \KL[\pi_C(z_{1:n}) {||} q_C(z_{1:n})] + \frac{\lambda}{2} (N_C - Z_C)^2 
\end{equation*}
where $\lambda > 0$ is a hyper-parameter, $N_C$ is the marginal likelihood (or the normalising constant) 
$\int p_C(z_{1:n})dz_{1:n}$ for $p_C$,
the next $\pi_C(z_{1:n})$ is the normalised posterior $p_C(z_{1:n})/N_C$ for $C$,
and the last $q_C$ and $Z_C$ are the approximate posterior and marginal likelihood estimate computed by the inference algorithm
(that is, $(q_C(z_{1:n}),Z_C) = \postinfer(C)$).
%\begin{align*}
%        N_C & = \int p_C(z_{1:n})dz_{1:n},
%       &
%        \pi_C(z_{1:n}) & = \frac{p_C(z_{1:n})}{N_C}, 
%       &
%        (q_C(z_{1:n}),Z_C) & = \postinfer(C).
%\end{align*}
%That is, 
%$N_C$ is the marginal likelihood (or the normalising constant) for $p_C$,
%the next
%$\pi_C$ is the normalised posterior for $C$, and the last $q_C$ and $Z_C$ are the approximate posterior and marginal likelihood estimate computed by the interference algorithm. 
Note that $q_C$ and $Z_C$ both depend on $\phi$, since $\postinfer$ uses the $\phi$-parameterised neural networks.

We optimise the objective by stochastic %(or black-box) 
gradient descent. The key component of the optimisation is a gradient estimator
%, which is 
derived as follows:
$(\nabla_\phi \sum_{C \in \cD} \KL[\pi_C {||} q_C] + \frac{\lambda}{2} (N_C - Z_C)^2) 
 =  (\sum_{C \in \cD} \EE_{z_{1:n} \sim \pi_C}[-\nabla_\phi \log q_C(z_{1:n})] - \lambda (N_C - Z_C) \nabla_\phi Z_C) 
\approx
\sum_{C \in \cD}
- \widehat{L_{C,\phi}} - 
\lambda (\widehat{N_{C}} - Z_{C}) \nabla_\phi Z_{C}$.
%\begin{align*}
% \nabla_\phi \sum_{C \in \cD} \KL[\pi_C {||} q_C] + \frac{\lambda}{2} (N_C - Z_C)^2 
%%& {} =  \sum_{C \in \cD} \nabla_\phi \KL[\pi_C {||} q_C] - \lambda (N_C - Z_C) \nabla_\phi Z_C 
%%       \\[-0.3ex]
%& {} =  \sum_{C \in \cD} \EE_{z_{1:n} \sim \pi_C}[-\nabla_\phi \log q_C(z_{1:n})] - \lambda (N_C - Z_C) \nabla_\phi Z_C 
%	\\[-0.3ex]
%& {} \approx
%\sum_{C \in \cD}
%- \widehat{L_{C,\phi}} - 
%\lambda (\widehat{N_{C}} - Z_{C}) \nabla_\phi Z_{C}.
%\end{align*}
Here
$\widehat{L_{C,\phi}}$ and $\widehat{N_{C}}$
are sample estimates of $\EE_{z_{1:n} \sim \pi_C}[\nabla_\phi \log q_C(z_{1:n})]$ and the marginal likelihood, respectively. Both estimates can be computed using standard Monte-Carlo algorithms. For instance, we can run the self-normalising importance sampler with prior as proposal, and generate weighted samples $\{(w^{(j)}, z^{(j)}_{1:n})\}_{1 \leq j \leq M}$ for the unnormalised posterior $p_C$. Then, we can use these samples to compute the required estimates: $\widehat{N_{C}} = \frac{1}{M}\sum_{j = 1}^M w^{(j)}$ and $\widehat{L_{C,\phi}} = \frac{1}{M}\sum_{j = 1}^M (w^{(j)} \nabla_\phi \log q_C(z^{(j)}_{1:n}))/\widehat{N_C}$. 
%\begin{align*}
%	\widehat{N_{C}} &= \sum_{j = 1}^M w^{(j)},
%	&
%	\widehat{L_{C,\phi}} = \sum_{j = 1}^M \frac{w^{(j)}}{\widehat{N_{C}}} \nabla_\phi \log q_C(z^{(j)}_{1:n}). 
%\end{align*}
Alternatively, we may run Hamiltonian Monte Carlo (HMC)~\citep{duane1987hybrid} to generate posterior samples, and use those samples to draw weighted importance samples using, for instance, the layered adaptive importance sampler~\citep{martino2017layered}. Then, we compute $\widehat{L_{C,\phi}}$ using posterior samples, and $\widehat{N_{C}}$ using weighted importance samples.
Note that neither $\pi_C$ in  $\EE_{z_{1:n} \sim \pi_C}[-\nabla_\phi \log q_C(z_{1:n})]$ 
nor $N_{C}$ 
depends on the parameters $\phi$. Thus, for each $C \in \cD$, $N_C$ needs to be estimated only once throughout the entire optimisation process, and the posterior samples from $\pi_C$ need to be generated only once as well. We use this fact to speed up the computation of each gradient-update step. 
%Another optimisation is to use subsampling over $\cD$ to avoid the cost of summing over all $C$'s in $\cD$ in a single gradient-update step. Both optimisations are used in our experiments.

\section{Empirical evaluation}
\label{sec:empirical}

An effective meta-algorithm should generalise well: the learnt inference algorithm should accurately predict the posteriors of 
programs unseen during training which have different parameters (\S\ref{sec:interpol}) and model structures (\S\ref{sec:extrapol}),
as long as the programs are similar to those in the training set. We empirically show that our 
meta-algorithm learns such an inference algorithm, and that in some cases
using the learnt inference algorithm achieves higher test-time efficiency than alternative approaches such as
HMC~\citep{duane1987hybrid} (\S\ref{sec:mulmod}).
We implemented our inference algorithm and meta-algorithm using ocaml-torch \citep{ocaml-torch}, an OCaml binding for PyTorch.
For HMC, we used the Python interface for Stan~\citep{carpenter2017stan}.
%For all the evaluations, 
We used a Ubuntu server with Intel(R) Xeon(R) Gold 6234 CPU @ 3.30GHz with
$16$ cores, $32$ threads, and 263G memory.
%We provide the full description of our model classes and the detailed experimental setup in the supplementary material.
See Appendix~\ref{appendix:models}
for the full list of our model classes and their details, and
Appendix~\ref{appendix:empirical-setup}
for the detailed experimental setup.

\begin{figure*}[t]
	%\vspace{-9.5mm}
	\centering
	\begin{subfigure}{0.329\textwidth}
		\includegraphics[width=\textwidth]
		{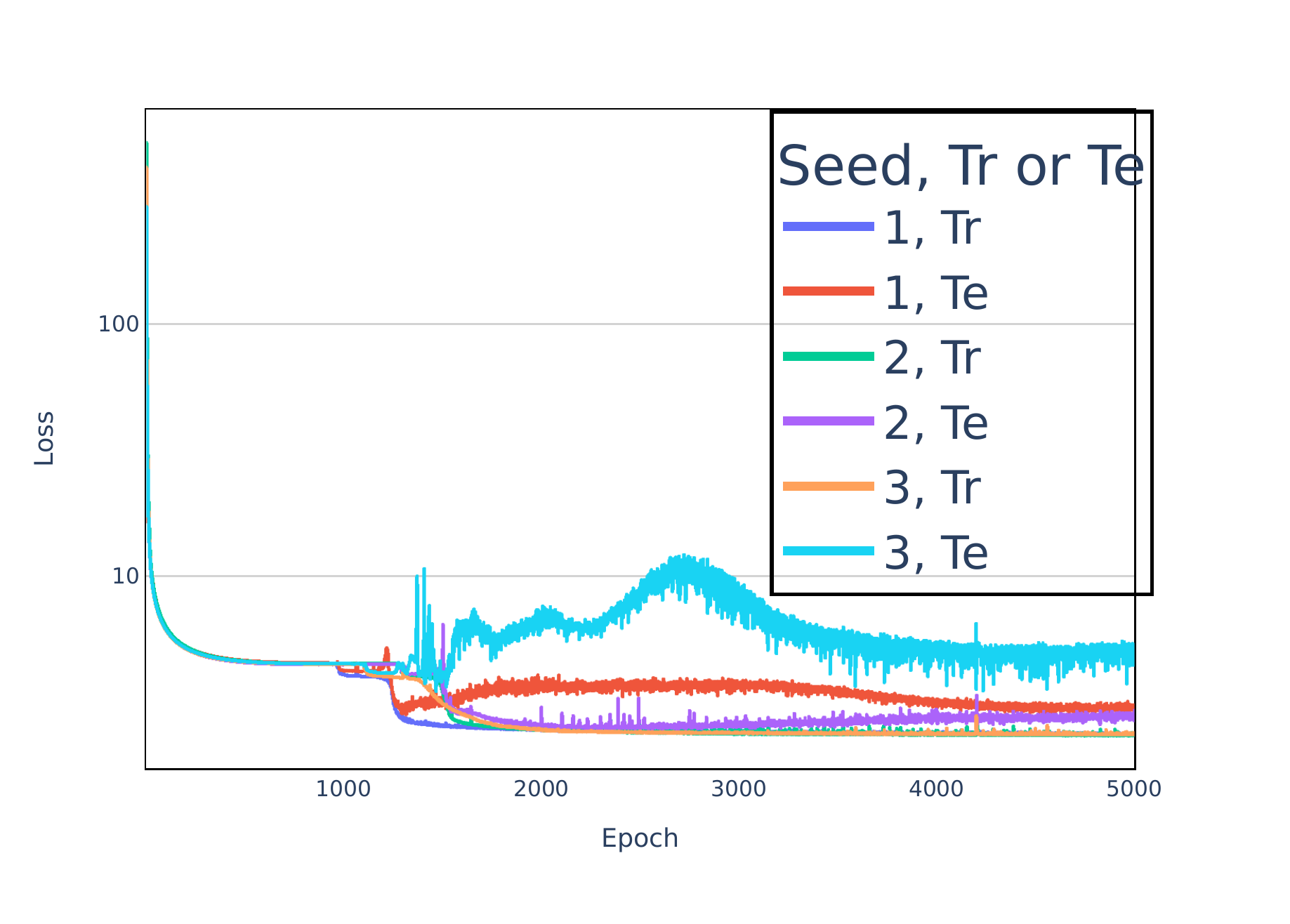}
		\caption{$\gauss$}
		\label{fig:gauss-loss}
	\end{subfigure}
	\begin{subfigure}{0.329\textwidth}
		\includegraphics[width=\textwidth]
		{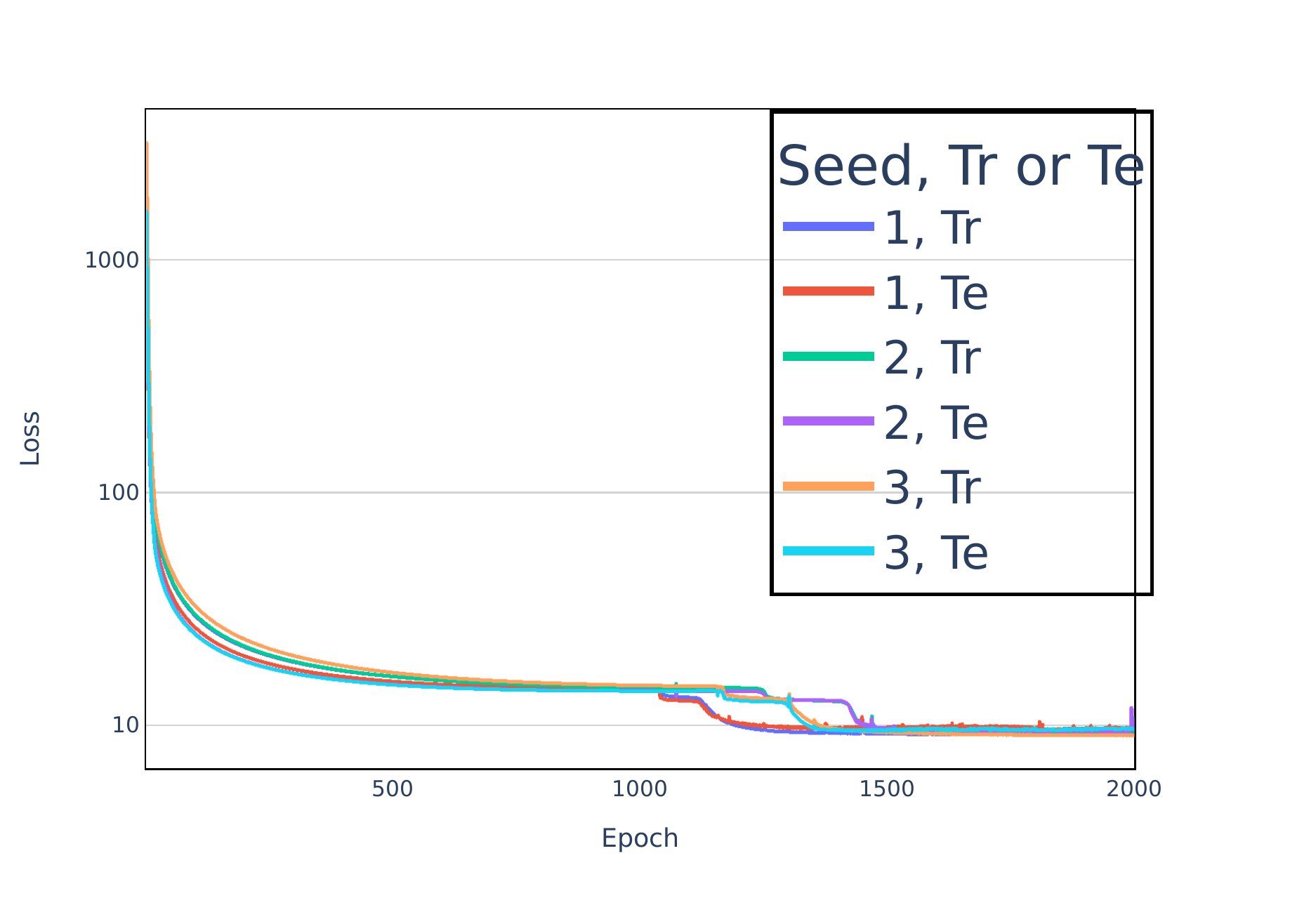}
		\caption{$\hierl$}
		\label{fig:hierl-loss}
	\end{subfigure}
	\begin{subfigure}{0.329\textwidth}
		\includegraphics[width=\textwidth]
		{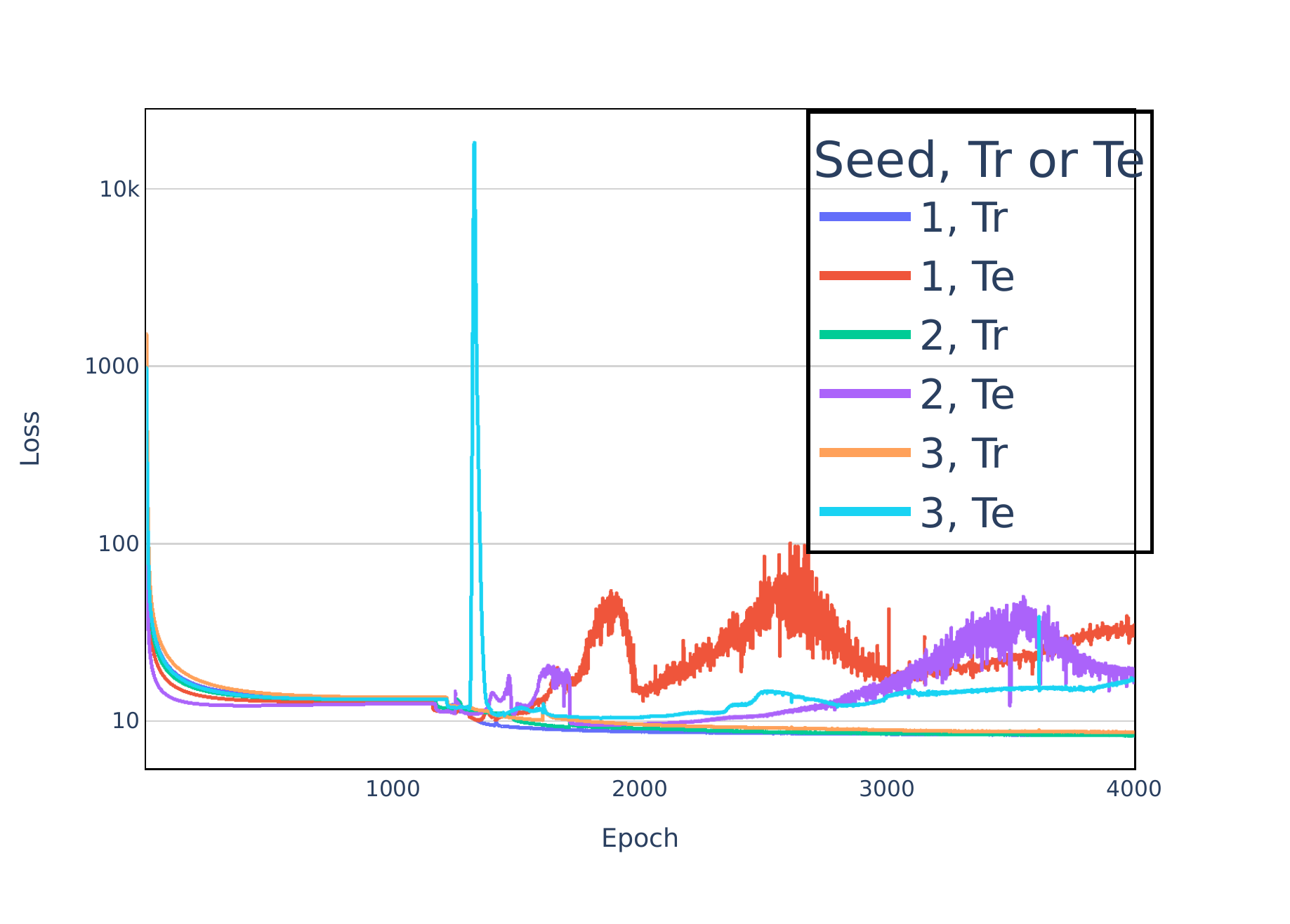}
		\caption{$\milky$}
		\label{fig:milky-loss}
	\end{subfigure}
	\caption{
		Average training and test losses under three random seeds. The $y$-axes are log-scaled.
                The increases in later epochs of Fig.~\ref{fig:milky-loss} were due to only one or
                a few test programs out of $50$.}
	\label{fig:loss}
        %\vspace{-5mm}
\end{figure*}

\begin{figure*}[t]
	\centering
	\begin{subfigure}{0.32\textwidth}
		\includegraphics[width=\textwidth]
		{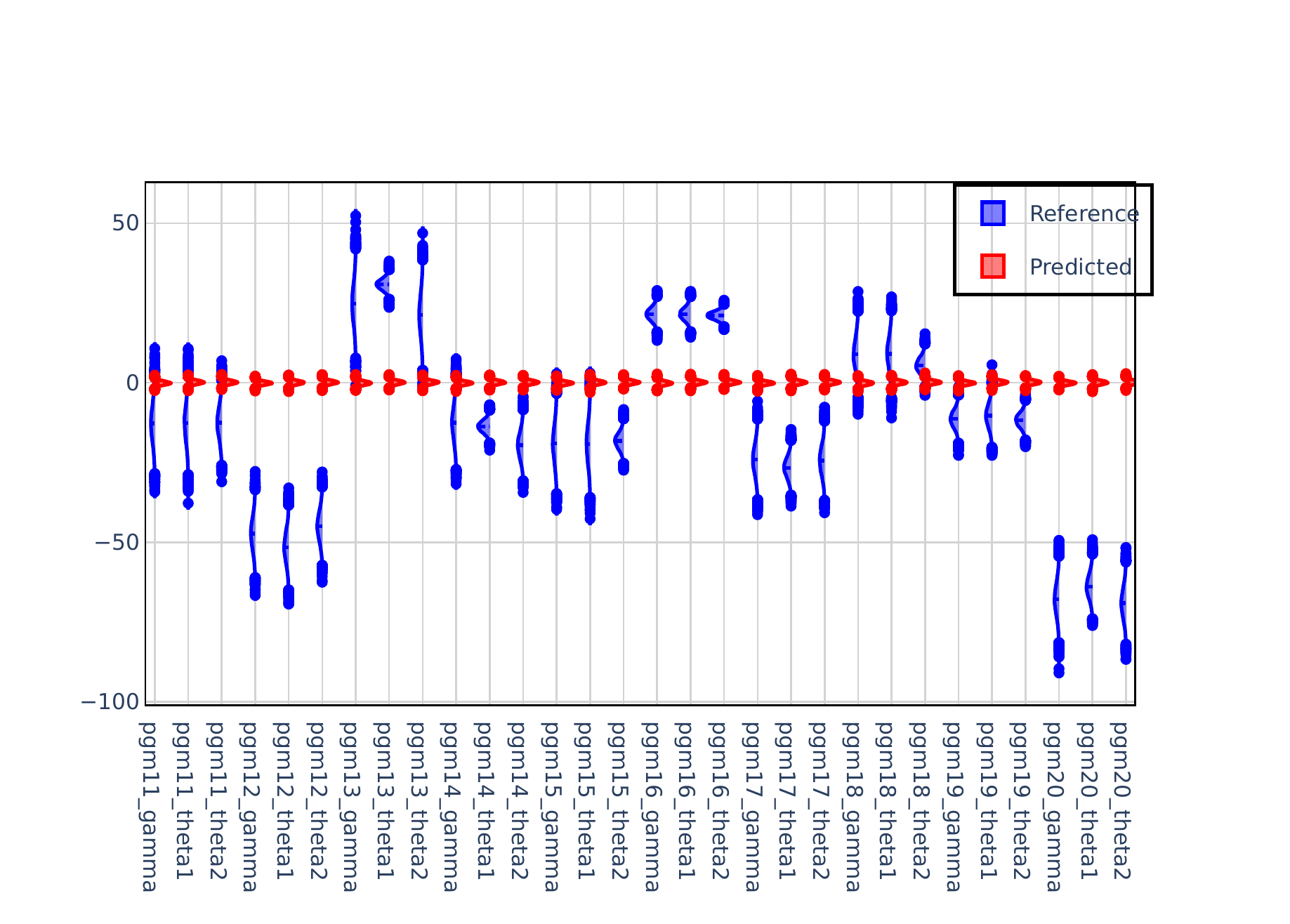}
		\caption{Before training}
		\label{fig:hierl-init}
	\end{subfigure}
	\begin{subfigure}{0.32\textwidth}
		\includegraphics[width=\textwidth]
		{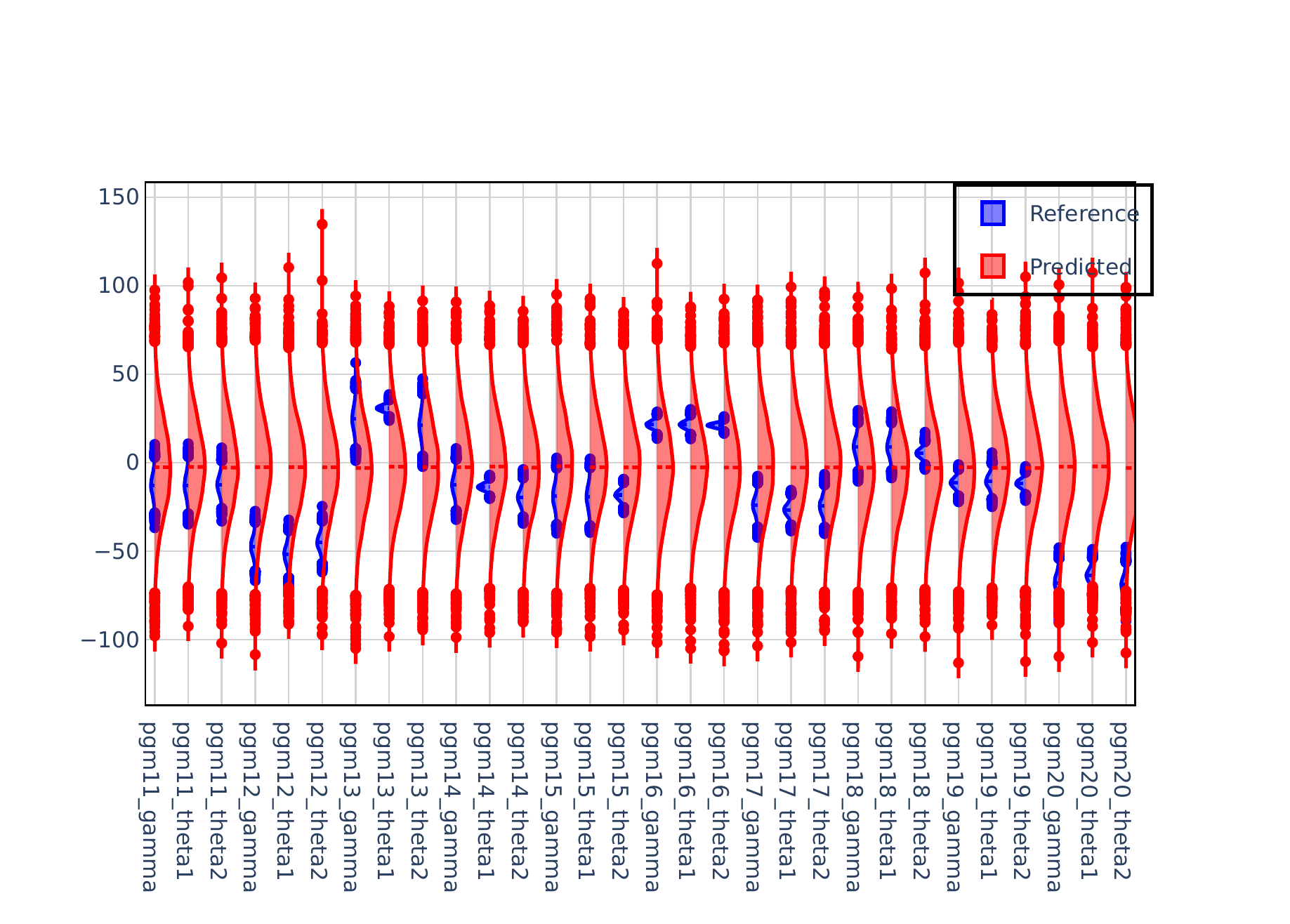}
		\caption{After 1K epochs}
		\label{fig:hierl-1K-epoch}
	\end{subfigure}
	\begin{subfigure}{0.32\textwidth}
		\includegraphics[width=\textwidth]
		{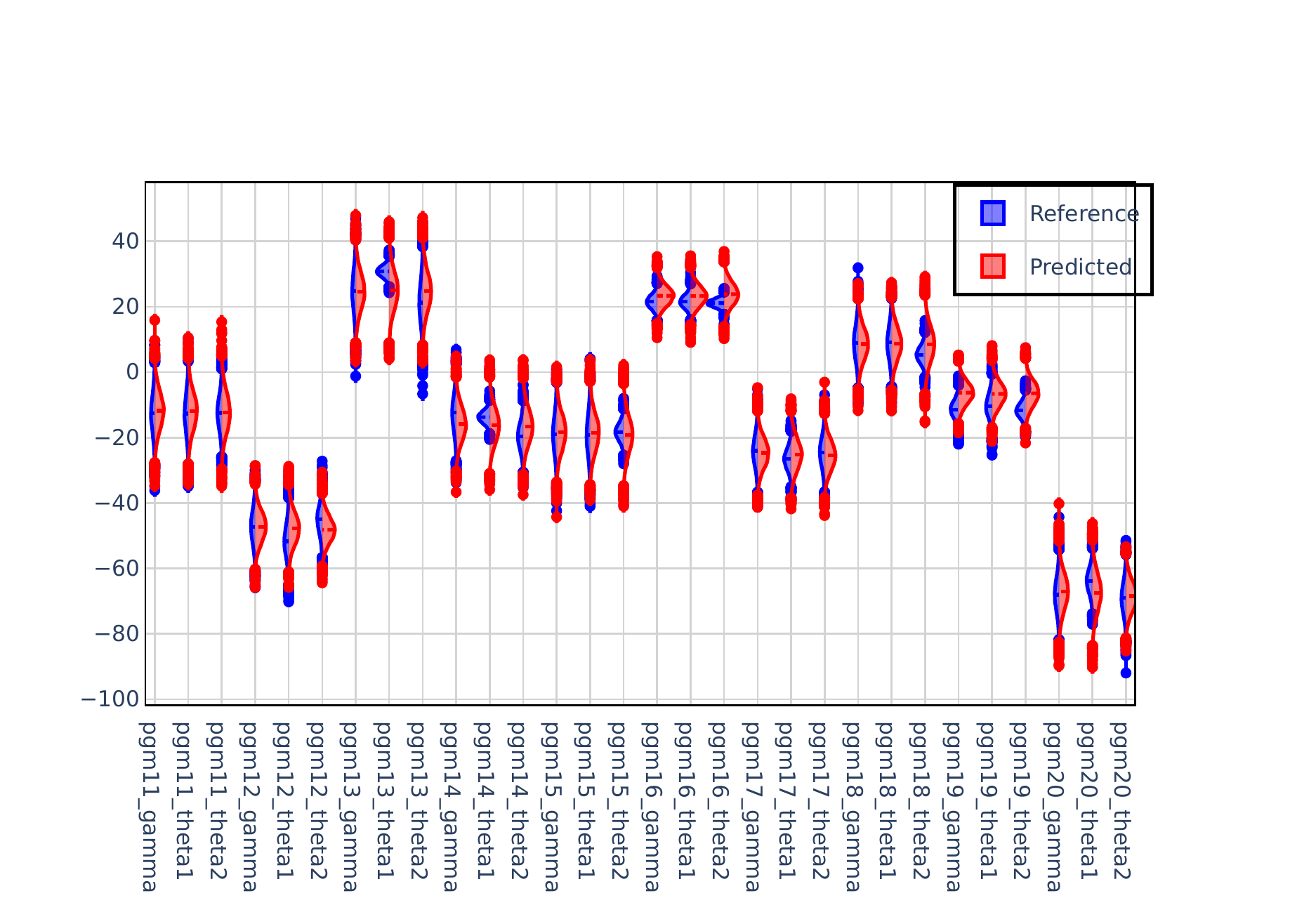}
		\caption{After 2K epochs}
		\label{fig:hierl-2K-epoch}
	\end{subfigure}
	\caption{Comparisons of predicted and reference marginal posteriors recorded at different training steps: at the initial step,
		after 1K epochs, and after 2K epochs.}
	\label{fig:hierl-evolution}
        %\vspace{-3mm}
\end{figure*}

\subsection{Generalisation to new model parameters and observations}
\label{sec:interpol}

We evaluated our approach with six model classes: (1) Gaussian models ($\gauss$) with a latent variable and an observation where the mean of the Gaussian likelihood is an affine transformation of the latent; (2) hierarchical models with three hierarchically structured latent variables ($\hierl$);
%where one of them, called a hyperprior, controls the other two; 
(3) hierarchical or multi-level models with both latent variables and data structured 
hierarchically ($\hierd$) where data are modelled as a regression of latent variables at different levels; (4) clustering models ($\cluster$) where five
observations are clustered into two groups;
%modelled by two latent variables and the cluster assignment is modelled by arithmetic comparison between $0$ and
%another latent variable that is normally distributed and centered at the origin; 
(5) Milky Way models ($\milky$), and their multiple-observations extension
($\milkyo$) where five observations are made for each satellite galaxy; 
%and an extended observe command takes a list of five observations at once;
and (6) models ($\rbk$) using the Rosenbrock function,\footnote{The function is often used to evaluate learning and inference algorithms~\citep{goodman2010ensemble,wang2018adaptive,pagani2019n}} which is expressed as an external procedure, to show that our approach can in principle 
handle models with non-trivial computation blocks.

The purpose of our evaluation is to show the feasibility of our approach, not to develop the state-of-the-art inference algorithm automatically, and also to identify the challenges of the approach. These models are chosen for this purpose. For instance, an inference algorithm should be able to reason about
affine transformations and Gaussian distributions (for $\gauss$), and dependency relationships among variables (for $\hierl$ and $\hierd$) to compute a
posterior accurately. Successful outcomes in the classes indicate that our approach learns an inference algorithm with such capacity in some
cases.

\noindent{\bf Setup}\ \
For each model class, we used $400$ 
programs to meta-learn an inference algorithm, and then applied the learnt algorithm to $50$ unseen test programs. 
%Specifically, 
We measured
the average test loss over the $50$ test programs, and checked if the loss also decreases when the training loss decreases.
We also compared the marginal 
posteriors predicted by our learnt inference algorithm with the reference marginal posteriors that were computed analytically, or approximately by HMC.
%When we relied on HMC to approximate the reference marginals.
When we relied on HMC,
we computed the marginal sample means and standard deviations using one of the
$10$ Markov chains generated by independent HMC runs. Each chain consisted of 500K samples after 50K warmups. We ensured the convergence of the 
chains using diagnostics such as $\hat{R}$ \citep{gelman1992inference}. All training and test programs were automatically generated by a random program 
generator. This generator takes a program class and hyperparameters %for the class 
(e.g., boundaries of the quantities that are used to specify
the models), and returns programs from the class randomly
%(see the supplementary material).
(see Appendix~\ref{appendix:models}).

For each training program, our meta-algorithm used $2^{15}$ samples from the analytic (for $\gauss$) or approximate (for the rest, by HMC) posterior 
distribution for the program.\footnote{Except for $\rbk$; see the discussion on Rosenbrock models in
	%the supplementary material.
	Appendix~\ref{appendix:multimodality}.
}
Similarly, our meta-algorithm computed the marginal likelihood analytically (for $\gauss$) or approximately (for the rest) using layered adaptive importance
sampling \citep{martino2017layered} where the proposals were defined by an HMC chain. We performed mini-batch training; a single gradient update was done with a training program and
a mini batch of size $2^{12}$ (out of $2^{15}$ samples for the program).
%, and so a single epoch consisted of $8$ gradient updates. 
We used Adam \citep{kingma2015adam} with its hyperparameters 
$\{\beta_1=0.9,\,\beta_2=0.999,\,\textrm{weight\_decay}=0\}$, and the initial learning rate was set to $0.001$.
When the average training loss converged enough, the training stopped.
We repeated the same experiments three times using different random seeds.
% to show that our meta-algorithm is resilient to the random seed effects.
%\hy{My changes are coloured red. I tried to clarify that a gradient update is done per a single training program with a mini-batch of the samples for the program. I commented out ``to show that our meta-algorithm is resilient to the random seed effects''. Either the reader will recorgnise this for himself or herself immediately after learning that we used three random seeds, or she or he won't care about multiple runs with different random seeds. The graphs are still small. Think about a way to make them large, in particular, the ones in Figure 3.}

\noindent{\bf Results}\ \
Fig.~\ref{fig:loss} shows the training and test losses for $\gauss$, $\hierl$, and $\milky$ under three random seeds. The losses for the other
model classes are in
Appendix~\ref{appendix:interpol-losses}.
%the supplementary material.
%The x-axis is the training epochs, and the y-axis the loss values.
The training loss was averaged over the training set and $8$ batch updates, and the test loss over the test set.
%The plots for training and test losses are drawn as solid and dotted lines, respectively, and the results with different random seeds are coloured differently.
The training losses in all three experiments decreased rapidly, and more importantly, these decreases were accompanied by the downturns
of the test losses, which shows that the learnt parameters generalised to the test programs well. The later part of
Fig.~\ref{fig:milky-loss} shows cases where the test loss increases. This was because the loss of only
a few programs in the test set (of $50$ programs) became large. Even in this situation, the losses of the rest remained small.

Fig.~\ref{fig:hierl-evolution} compares, for $10$ test programs in $\hierl$, the reference marginal posteriors (blue) and their predicted counterparts (red) by the learnt inference algorithm instantiated at three different training epochs. The predicted marginals were initially around zero 
(Fig.~\ref{fig:hierl-init}), evolved to cover the reference marginals (Fig.~\ref{fig:hierl-1K-epoch}), and finally captured them precisely in terms of 
both mean and standard deviation for most of the variables (Fig.~\ref{fig:hierl-2K-epoch}). The results show that our meta-algorithm improves the parameters of our inference 
algorithm, and eventually finds optimal ones that generalise well. We observed similar patterns for the other model classes and random seeds, except for 
$\cluster$ and $\rbk$; programs from these classes often have multimodal posteriors, and we provide an analysis for them in
%the supplementary material.
Appendix~\ref{appendix:multimodality}.

\begin{figure*}[t]
	%\vspace{-9.5mm}
	\centering
	\begin{subfigure}{0.329\textwidth}
		\includegraphics[width=\textwidth]
		{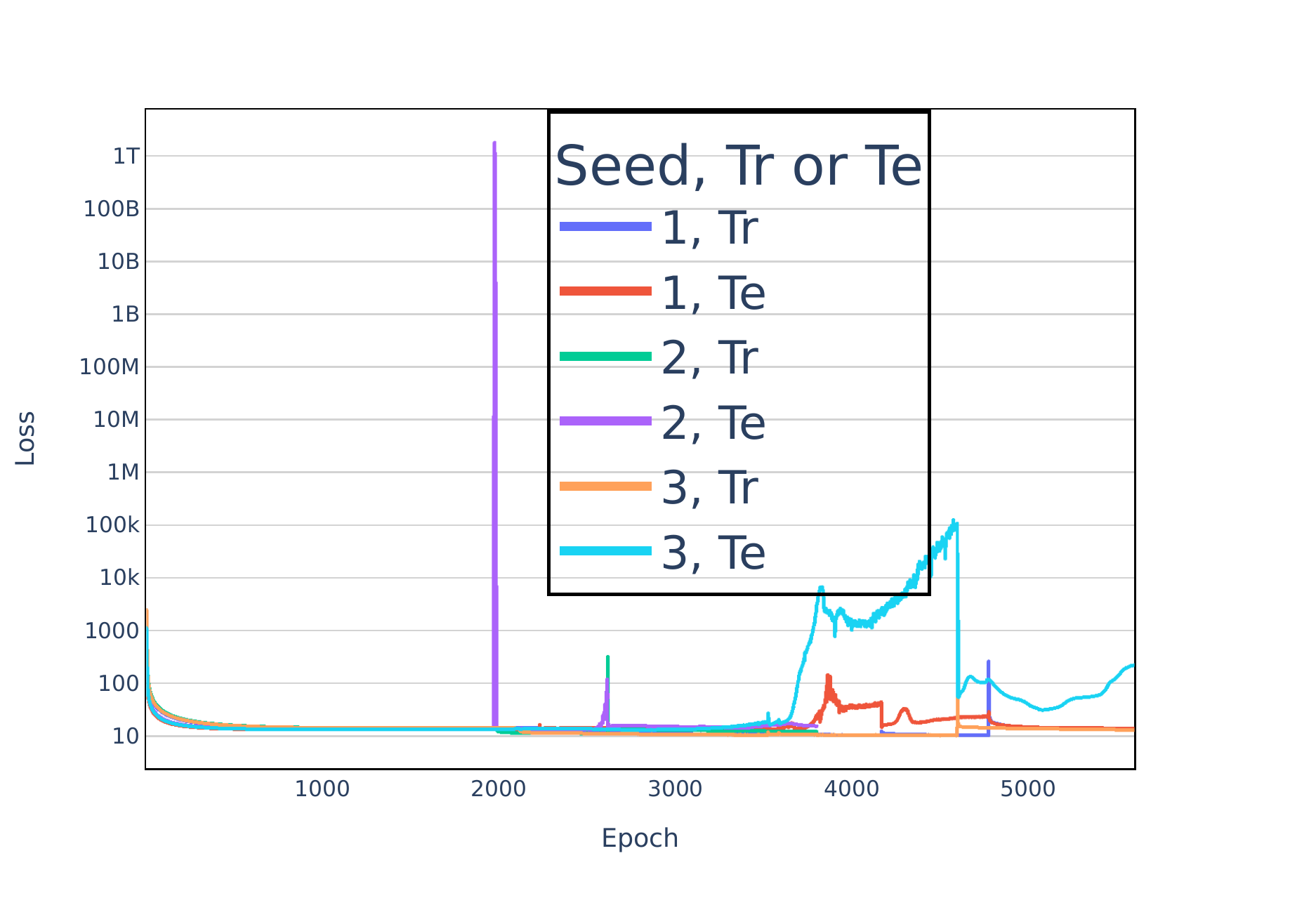}
		\caption{To 1st dep. graph.}
		%\label{}
	\end{subfigure}
	\begin{subfigure}{0.329\textwidth}
		\includegraphics[width=\textwidth]
		{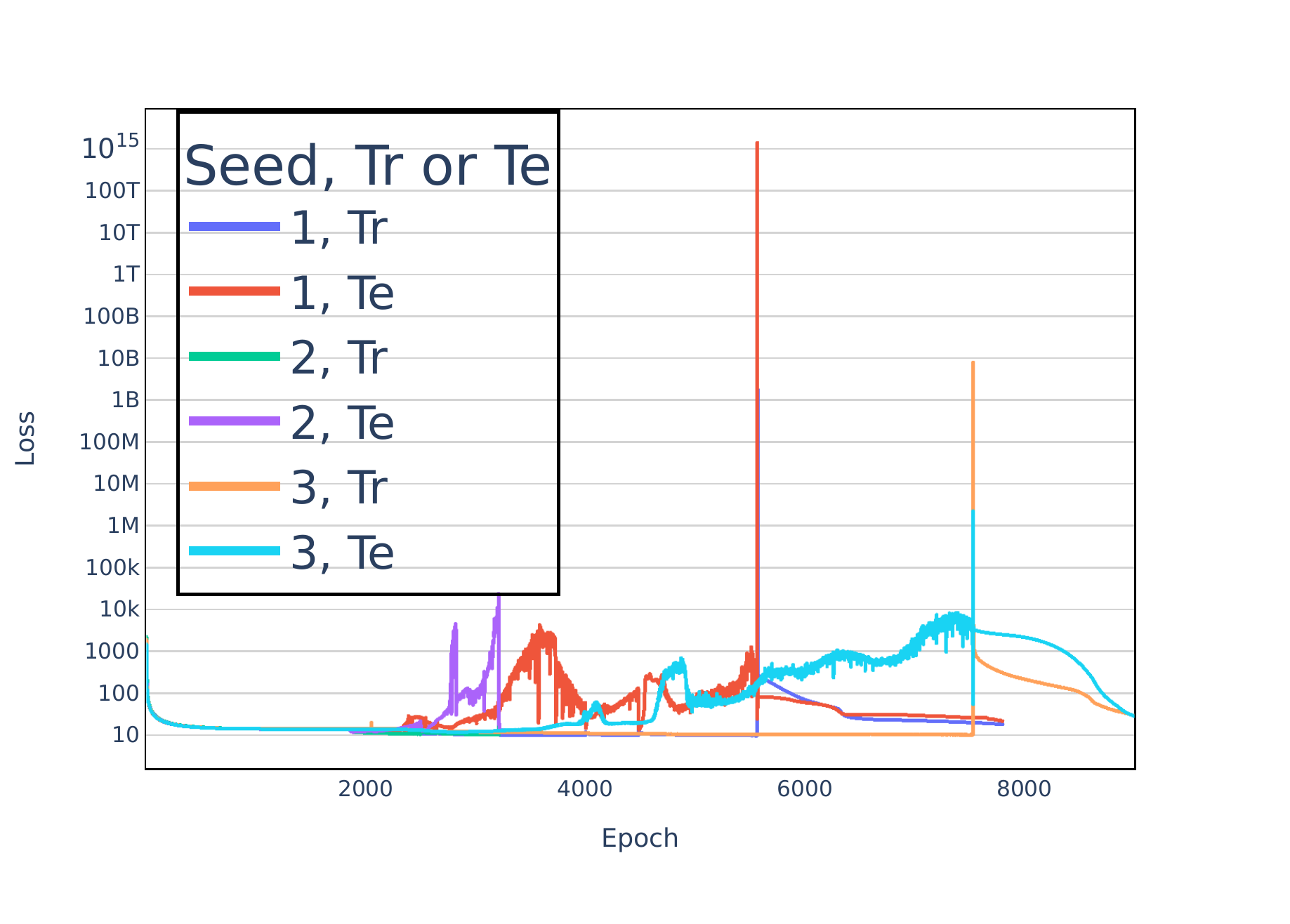}
		\caption{To 2nd dep. graph.}
		%\label{}
	\end{subfigure}
	\begin{subfigure}{0.329\textwidth}
		\includegraphics[width=\textwidth]
		{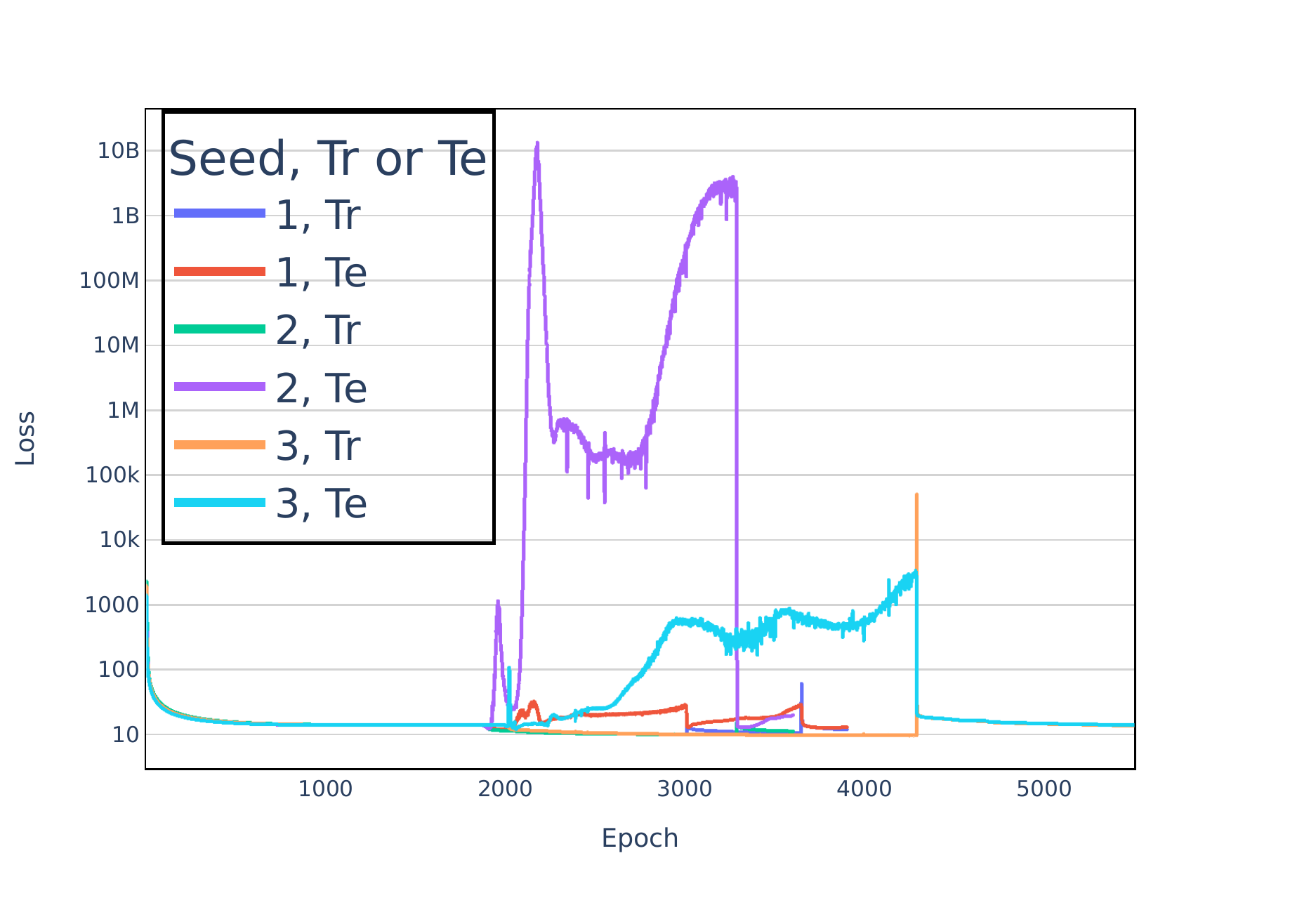}
		\caption{To 3rd dep. graph.}
		%\label{}
	\end{subfigure}
	\caption{
		Average losses for generalisation to dependency graphs in $\ext1$. The y-axes are log-scaled.
	}
	\label{fig:ext1}
        %\vspace{-3mm}
\end{figure*}

\begin{figure*}[t]
	\centering
	\begin{subfigure}{0.329\textwidth}
		\includegraphics[width=\textwidth]
		{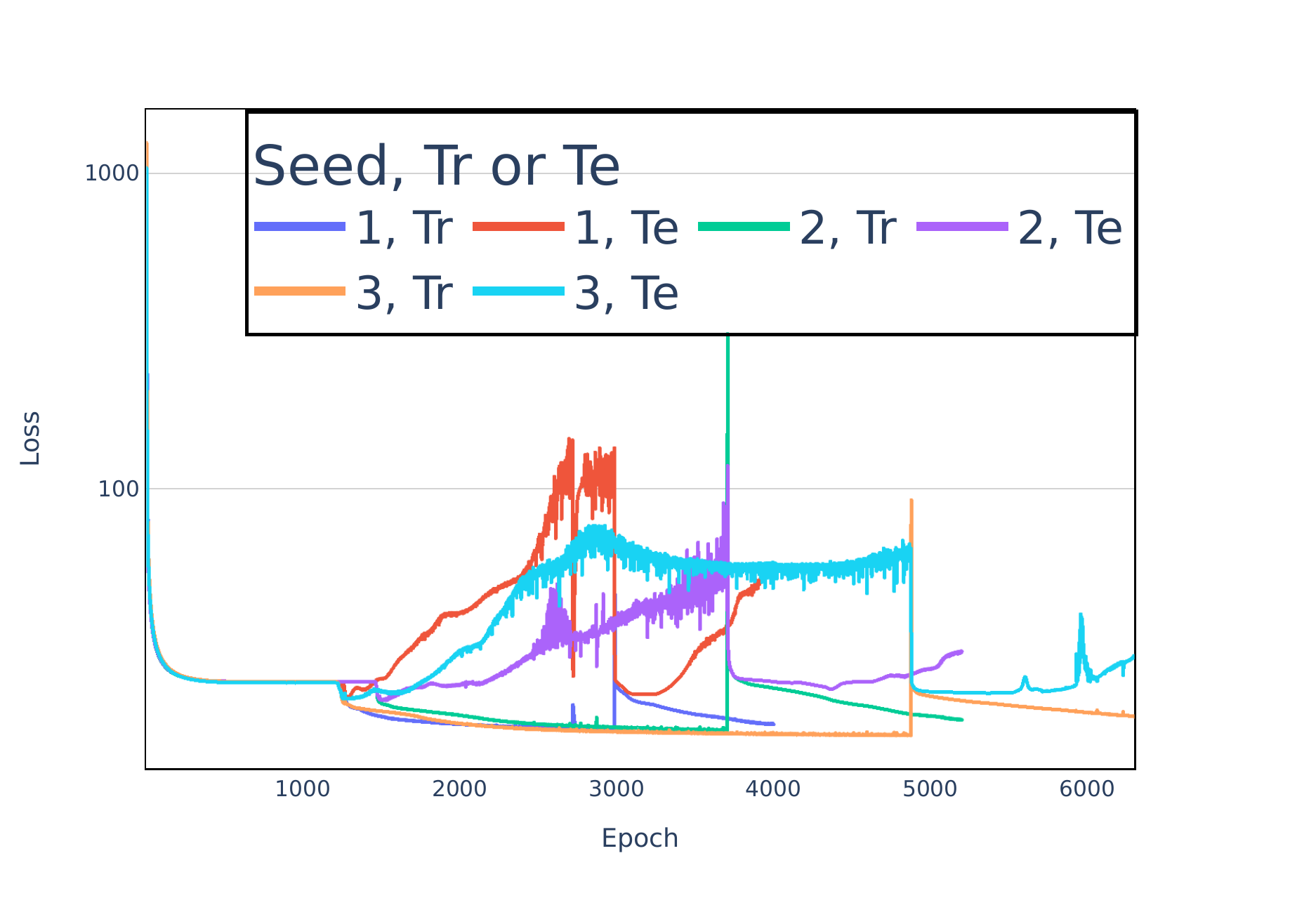}
		\caption{To 1st dep. graph.}
		%\label{}
	\end{subfigure}
	\begin{subfigure}{0.329\textwidth}
		\includegraphics[width=\textwidth]
		{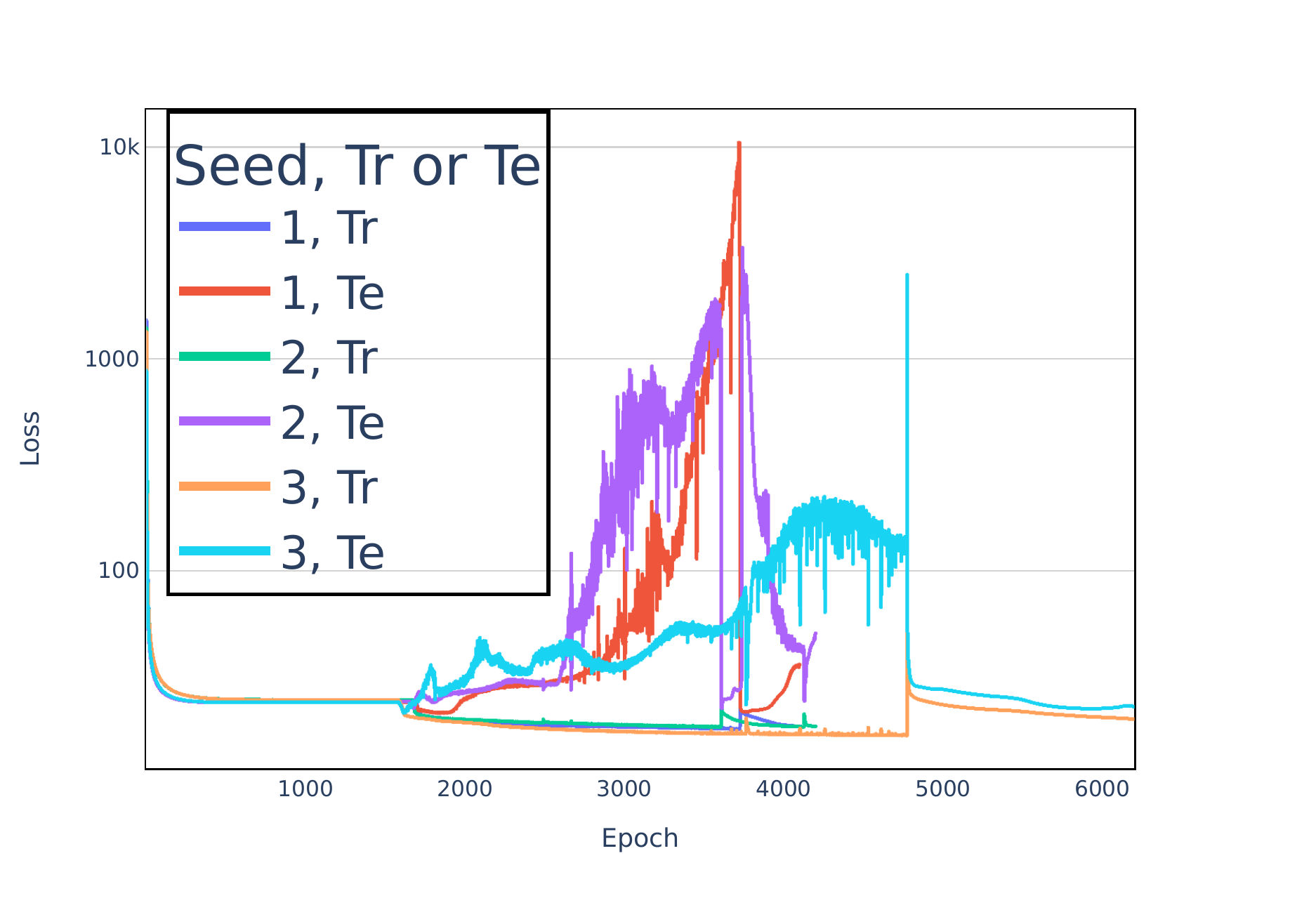}
		\caption{To 2nd dep. graph.}
		%\label{}
	\end{subfigure}
	\begin{subfigure}{0.329\textwidth}
		\includegraphics[width=\textwidth]
		{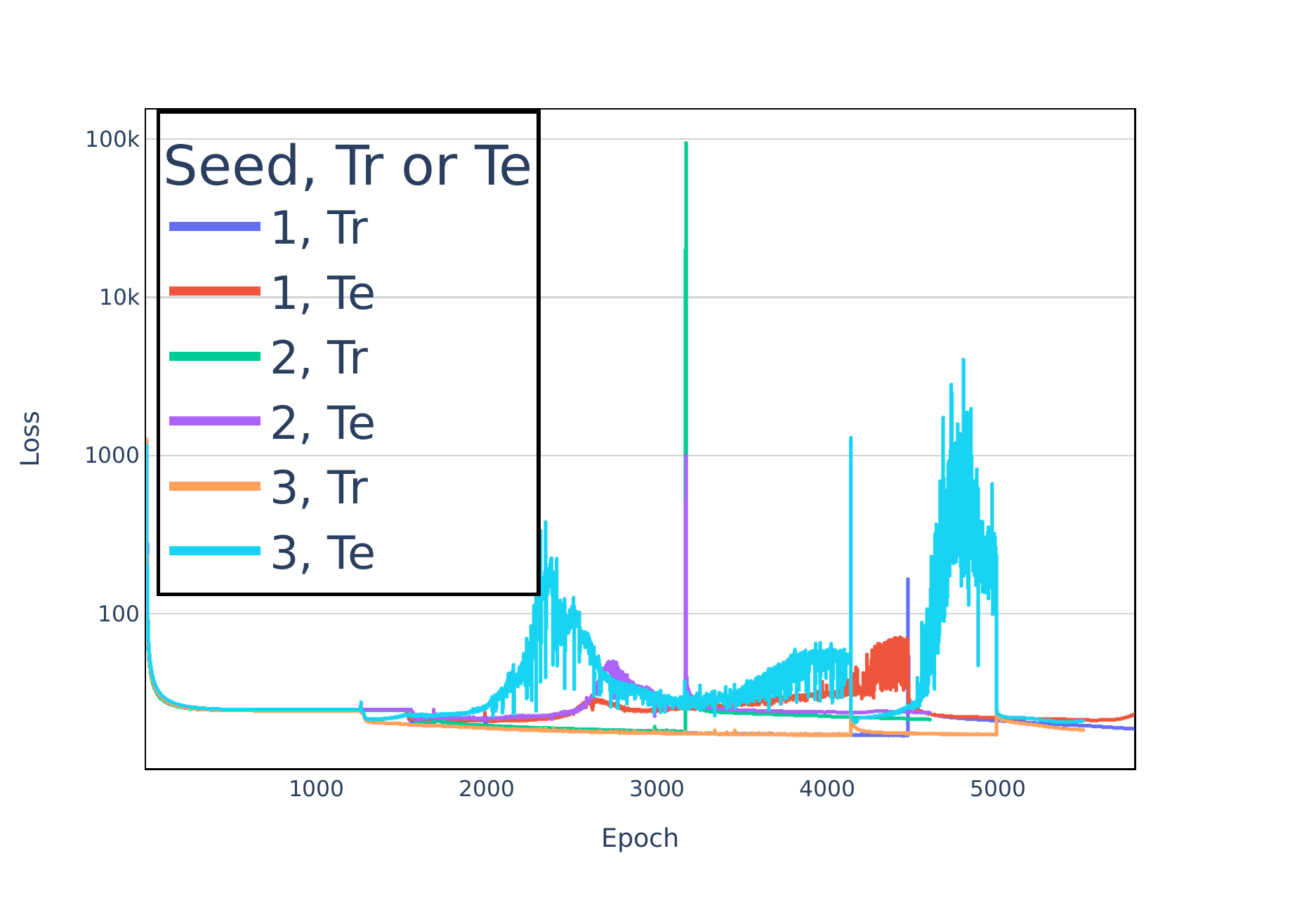}
		\caption{To 3rd dep. graph.}
		%\label{}
	\end{subfigure}
	\caption{
		Average losses for generalisation to dependency graphs in $\ext2$. The y-axes are log-scaled.
	}
	\label{fig:ext2}
        %\vspace{-3mm}
\end{figure*}

%%%%%%%%%%%%%%%
% TODO: Move the results below to appendix. (These experiments are not mentioned in the main text.)
%%%%%%%%%%%%%%%

%\begin{figure*}[t]
%	\centering
%	\begin{subfigure}{0.24\textwidth}
%		\includegraphics[width=\textwidth]
%		{extrapol/NEURIPS21_losses_nld3_te_subtype1.pdf}
%		\caption{$\extstype{1}{1}$}
%		\label{}
%	\end{subfigure}
%	\begin{subfigure}{0.24\textwidth}
%		\includegraphics[width=\textwidth]
%		{extrapol/NEURIPS21_losses_nld3_te_subtype2.pdf}
%		\caption{$\extstype{1}{2}$}
%		\label{}
%	\end{subfigure}
%	\begin{subfigure}{0.24\textwidth}
%		\includegraphics[width=\textwidth]
%		{extrapol/NEURIPS21_losses_nld3_te_subtype3.pdf}
%		\caption{$\extstype{1}{3}$}
%		\label{}
%	\end{subfigure}
%	\begin{subfigure}{0.24\textwidth}
%		\includegraphics[width=\textwidth]
%		{extrapol/NEURIPS21_losses_nld3_te_subtype4.pdf}
%		\caption{$\extstype{1}{4}$}
%		\label{}
%	\end{subfigure}
%	\caption{
%		First extrapolation experiments.
%	}
%	\label{fig:ext1-loss}
%\end{figure*}

\subsection{Generalisation to new model structures}
\label{sec:extrapol}
We let two kinds of model structure vary across programs: the dependency (or data-flow) graph for the variables of a program and the position of a nonlinear
%, deterministic 
function in the program. 
%We also allowed the variable names (indices) to be permuted arbitrarily. 
Specifically, we considered two 
model classes: (1) models ($\ext1$) with three Gaussian variables and one deterministic variable storing the value of the function $\nl(x) = 50 / \pi \times \arctan(x/10)$, where the models have $12$ different types --- four different dependency graphs of the variables, and three different positions of the deterministic $\nl$ variable for each of these graphs; and (2) models ($\ext2$) with six Gaussian variables and one $\nl$ variable, which are grouped into five types based on their dependency graphs.
The evaluation was done for $\ext1$ and $\ext2$ independently as in \S\ref{sec:interpol}, but here each of $\ext1$ and $\ext2$ itself has programs of 
multiple ($12$ for $\ext1$ and $5$ for $\ext2$) model types.
See Fig.~\ref{fig:ext1-types} and \ref{fig:ext2-types} in the appendix for visualisation of the different model types in $\ext1$ and $\ext2$, respectively.

%Before running our inference algorithm, we canonicalise the names of variables in a given program based on its dependency (i.e., data-flow) graph.  Although not perfect, this preprocessing %step 
%removes a superficial difference between programs caused by different variable names, and enables us to avoid unnecessary complexity caused by variable-renaming symmetries at training and inference times.

\noindent{\bf Setup}\ \
For $\ext1$, we ran seven different experiments. Three of them evaluated generalisation to unseen positions of the $\nl$ variable, and the other four
to unseen dependency graphs. Let $T_{i,j}$ be the type in $\ext1$ that corresponds to the $i$-th position of $\nl$ and the $j$-th dependency graph, and
$T_{-i,*}$ be all the types in $\ext1$ that correspond to any $\nl$ positions except the $i$-th and any of four dependency graphs. For generalisation to the
$i$-th position of $\nl$ ($i=1,2,3$), we used programs from $T_{-i,*}$ for training and those from $T_{i,*}$ for testing. For generalisation to the $j$-th 
dependency graph ($j=1,2,3,4$), we used programs from $T_{*,-j}$ for training and those from $T_{*,j}$ for testing.
For $\ext2$, we ran five different experiments where each of them tested generalisation to an unseen dependency graph after training with the other four 
dependency graphs. All these experiments were repeated three times under different random seeds. So, the total numbers of experiment runs were 
$21\,(=7\times3)$ and $15\,(=5\times3)$ for $\ext1$ and $\ext2$, respectively.
\iffalse
For $\ext1$, we ran seven different experiments. Four of them evaluated generalisation to unseen dependency graphs. Among all the four possible dependency graphs  of programs we considered, these experiments used only three of them during training, and the other for testing: the dependency graph of each training program is one of these three (with the $\nl$ variable positioned in all the three possible places in the graph), and that of each test program is the other dependency graph.  The remaining three experiments evaluated generalisation to unseen positions of the $\nl$ variable where we fixed the position of the $\nl$ variable in the four dependency graphs, and used programs with the variable in one of those fixed positions for testing, while using programs with the variables in all the other positions for training. For $\ext2$, we ran five different experiments where each of them 
tested generalisation to an unseen dependency graph after training with the other four types of dependency graphs.
All of these experiments were repeated three times under different random seeds. So, the total numbers of experiment runs were $21\,(=7\times3)$ and $15\,
(=5\times3)$ for
$\ext1$ and $\ext2$, respectively.
\fi

In each experiment run for $\ext1$, we used $720$ programs for training, and $90$ (when generalising to new graphs) or $100$ (when generalising to new positions of the $\nl$ variable) unseen programs for testing. In each run for $\ext2$, we used $600$ programs for training and tested the learnt inference algorithm on $50$ unseen programs. We ran HMC to estimate posteriors and marginal likelihoods, and used $200$K samples after $10$K warmups to compute reference posteriors. We stopped training after giving enough time for convergence within a limit of computational resources. The rest was the same as in \S\ref{sec:interpol}.

\noindent{\bf Results}\ \
Fig.~\ref{fig:ext1} shows the average training and test losses for generalisation to the first three dependency graphs in $\ext1$. Fig.~\ref{fig:ext2}
shows the losses for generalisation to the first three dependency graphs in $\ext2$. The losses for generalisation to the last dependency graph and to all positions of the $\nl$ variable in $\ext1$,
and those for generalisation to the $4$th and $5$th dependency graphs in $\ext2$ are in
%the supplementary material.
Appendices~\ref{appendix:ext1-losses} and \ref{appendix:ext2-losses}.
%, respectively.
In $17$ runs (out of $21$) for $\ext1$, the decrease in the training losses eventually stabilised or reduced the test losses,
even when the test losses were high and fluctuated in earlier training epochs. In $8$ runs (out of $15$) for $\ext2$, the test losses were 
stabilised as the training losses decreased. In $4$ runs out of the other $7$, the test losses increased only slightly.
In terms of predicted posteriors, we observed highly accurate predictions in $8$ runs for $\ext1$. For $\ext2$, the predicted posteriors were accurate in 
$7$ runs. For quantified accuracy, we refer the reader to Appendix~\ref{appendix:predicted-posteriors-accuracy}.
Overall, the learnt algorithms generalised to unseen types of models well or 
fairly well in many cases.

\begin{table}[t]
	%\vspace{-7mm}
	\small
	\caption{ESS per sec for the $60$ test programs by HMC vs. IS-pred vs. IS-prior.}
	\centering
	\aboverulesep=0.3ex
	\belowrulesep=0.3ex
	\begin{tabular}{lccccccccc}
		\toprule
		&\multicolumn{3}{c}{HMC}&\multicolumn{3}{c}{IS-pred}&\multicolumn{3}{c}{IS-prior}
		\\ \midrule
		&GM&Q1&Q3&GM&Q1&Q3&GM&Q1&Q3
		\\ \midrule
		ESS&$204.8$K&$4.1$K&$4.6$M&$4.2$K&$2.2$K&$13.8$K&$2.8$K&$1.1$K&$9.5$K
		\\ \midrule
		Time&$48.2$s&$27.4$s&$82.3$s&$22.7$ms&$21.4$ms&$23.0$ms&$23.1$ms&$22.1$ms&$24.0$ms
		\\ \midrule[0.08em]
		ESS / sec&$4.3$K&$124$&$127.7$K&$\bm{196.5}$\textbf{K}&$\bm{102.4}$\textbf{K}&$\bm{646.5}$\textbf{K}&$123.8$K&$52.6$K&$436.1$K
		\\ \bottomrule
	\end{tabular}
	\label{tab:mulmod-ess-all-test-pgms}
\end{table}

\begin{figure*}[t]
	%\vspace{-4mm}
	\centering
	\begin{subfigure}{0.329\textwidth}
		\includegraphics[width=\textwidth]
		{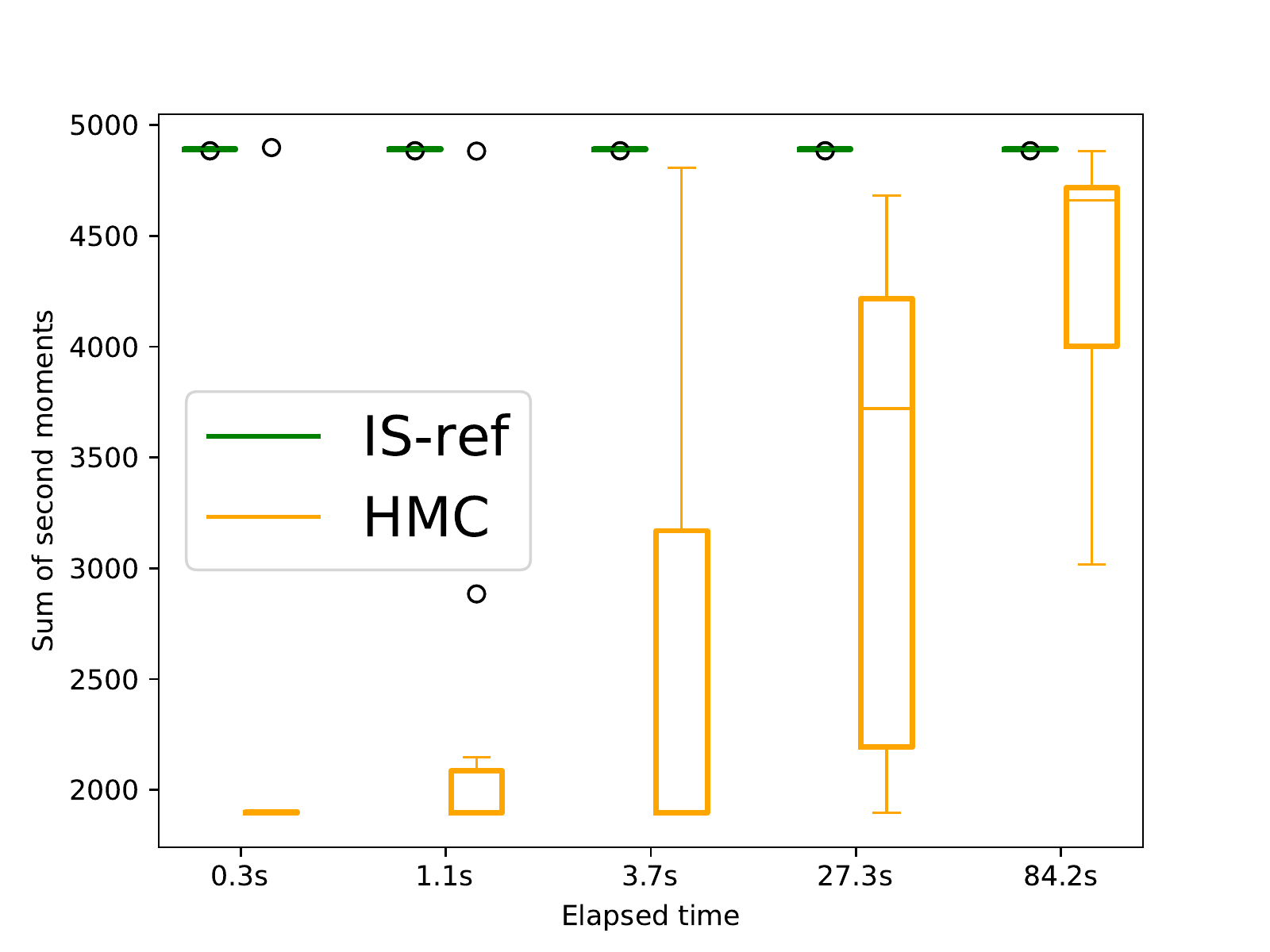}
		\caption{Moments by HMC.}
		\label{fig:mulmod-hmc-moments}
	\end{subfigure}
	\begin{subfigure}{0.329\textwidth}
		\includegraphics[width=\textwidth]
		{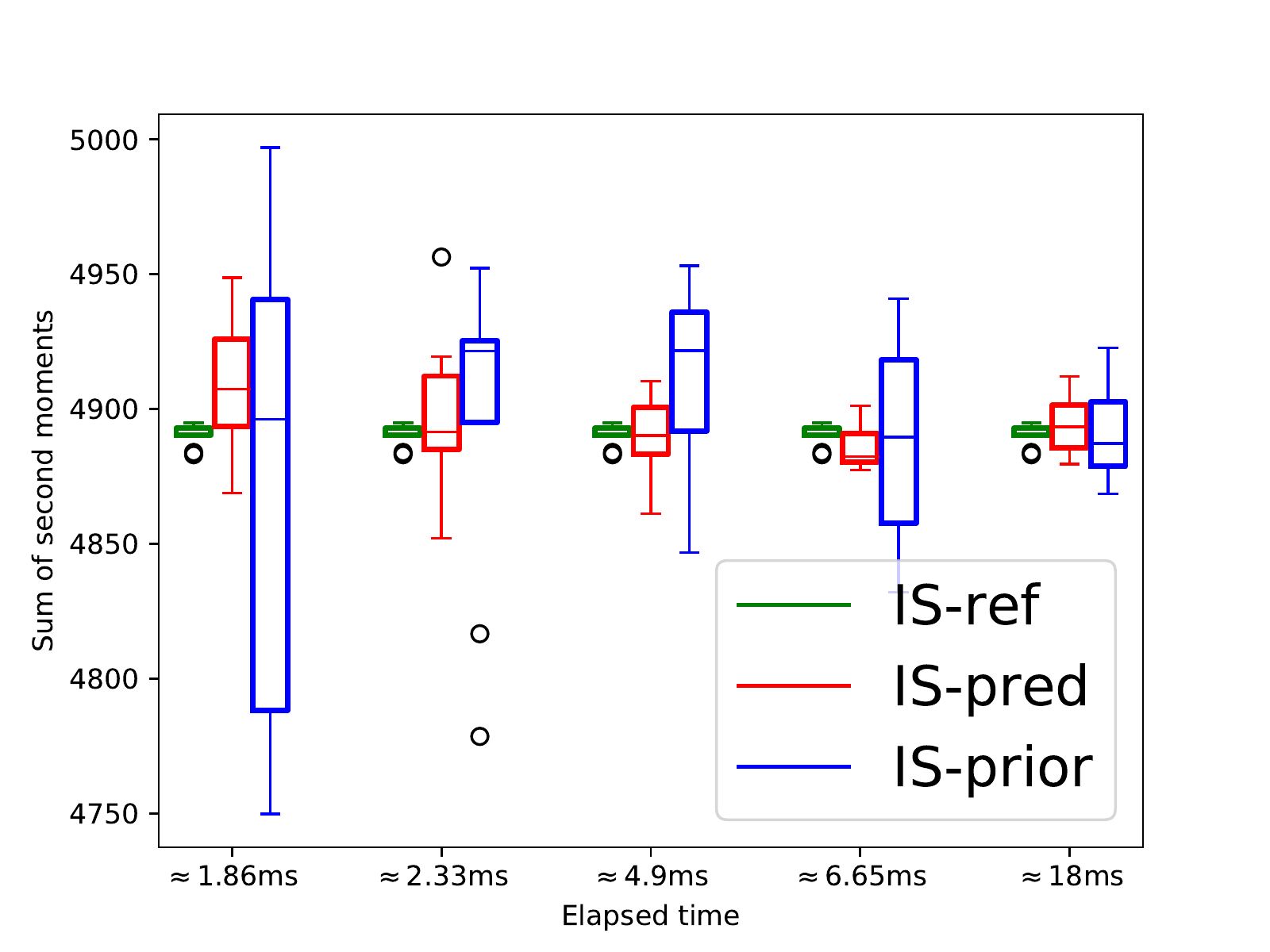}
		\caption{Moments by IS-pred and IS-prior.}
		\label{fig:mulmod-is-moments}
	\end{subfigure}
	\begin{subfigure}{0.329\textwidth}
		\includegraphics[width=\textwidth]
		{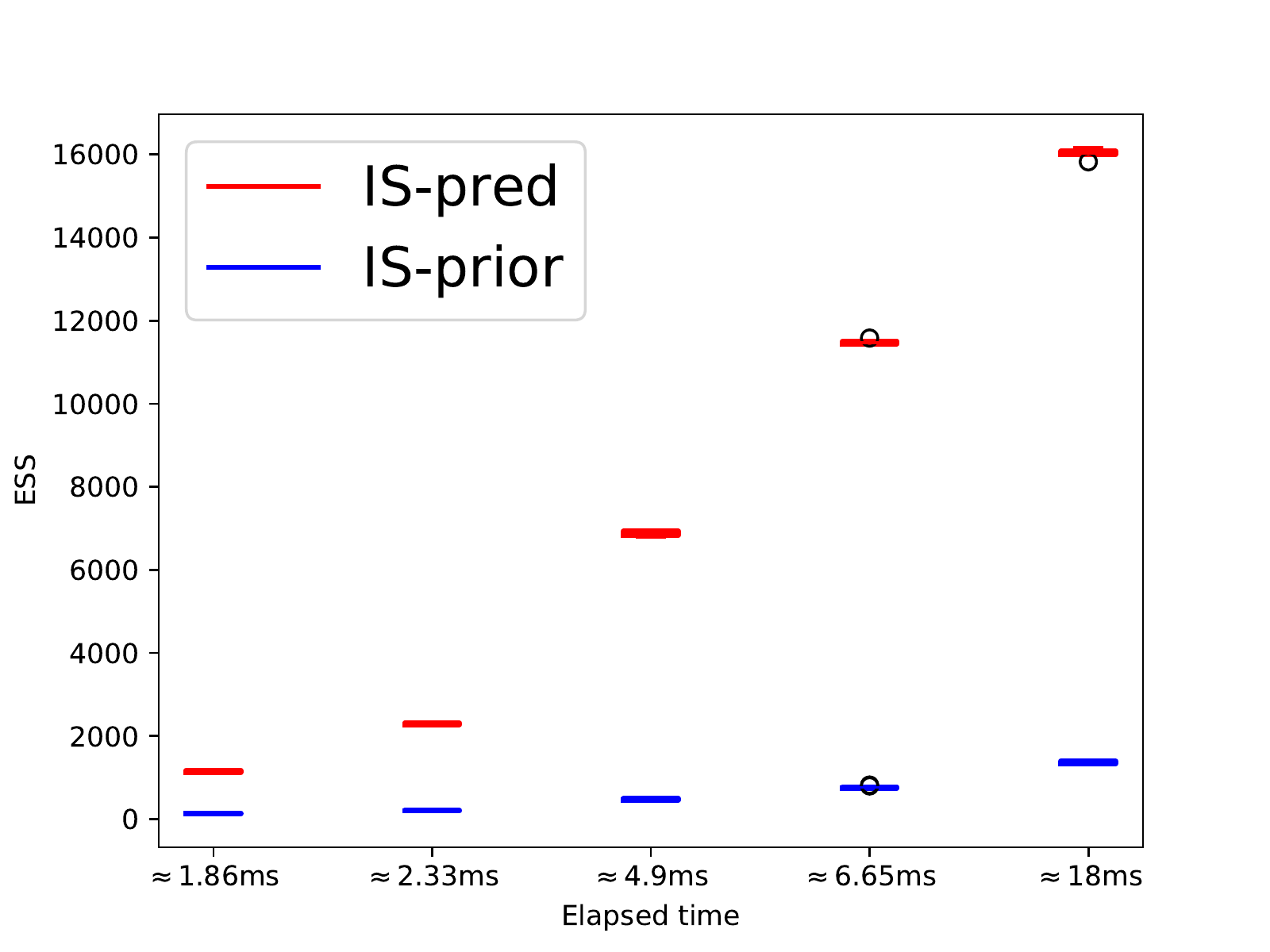}
		\caption{ESS by IS-pred and IS-prior.}
		\label{fig:mulmod-is-ess}
	\end{subfigure}
	\caption{}
	%\label{}
	%\vspace{-6mm}
\end{figure*}

%\begin{table}[t]
%	\caption{
%		ESS by IS-pred, compared with those by HMC and IS-prior. For HMC, two kinds of ESS were computed
%		(bulk and tail~\citep{carpenter2017stan}) for each latent variable, and we report the maximum among them. For IS-\{pred, prior\},
%		ESS was averaged over the $10$ trials. The elapsed time for all the three approaches was averaged over $10$ trials.
%	}
%	\centering
%	\aboverulesep=0.3ex
%	\belowrulesep=0.3ex
%	\begin{tabular}{lccc}
%		\toprule
%		&HMC&IS-pred&IS-prior
%		\\ \midrule
%		ESS&$80.0$&$\bm{16,030.1}$&$1,364.1$
%		\\ \midrule
%		Elapsed time&$84.2$s&$\bm{\approx 18}\textbf{ms}$&$\approx 18$ms
%		\\ \midrule
%		Sample size&$1$M&$70$K&$100$K
%		\\ \bottomrule
%	\end{tabular}
%	\label{tab:mulmod-hmc-ess}
%	%\vspace{-3mm}
%\end{table}

\begin{figure}[t]
	%\vspace{-8mm}
	\centering
	\begin{subfigure}{0.245\textwidth}
		\includegraphics[width=\textwidth]
		{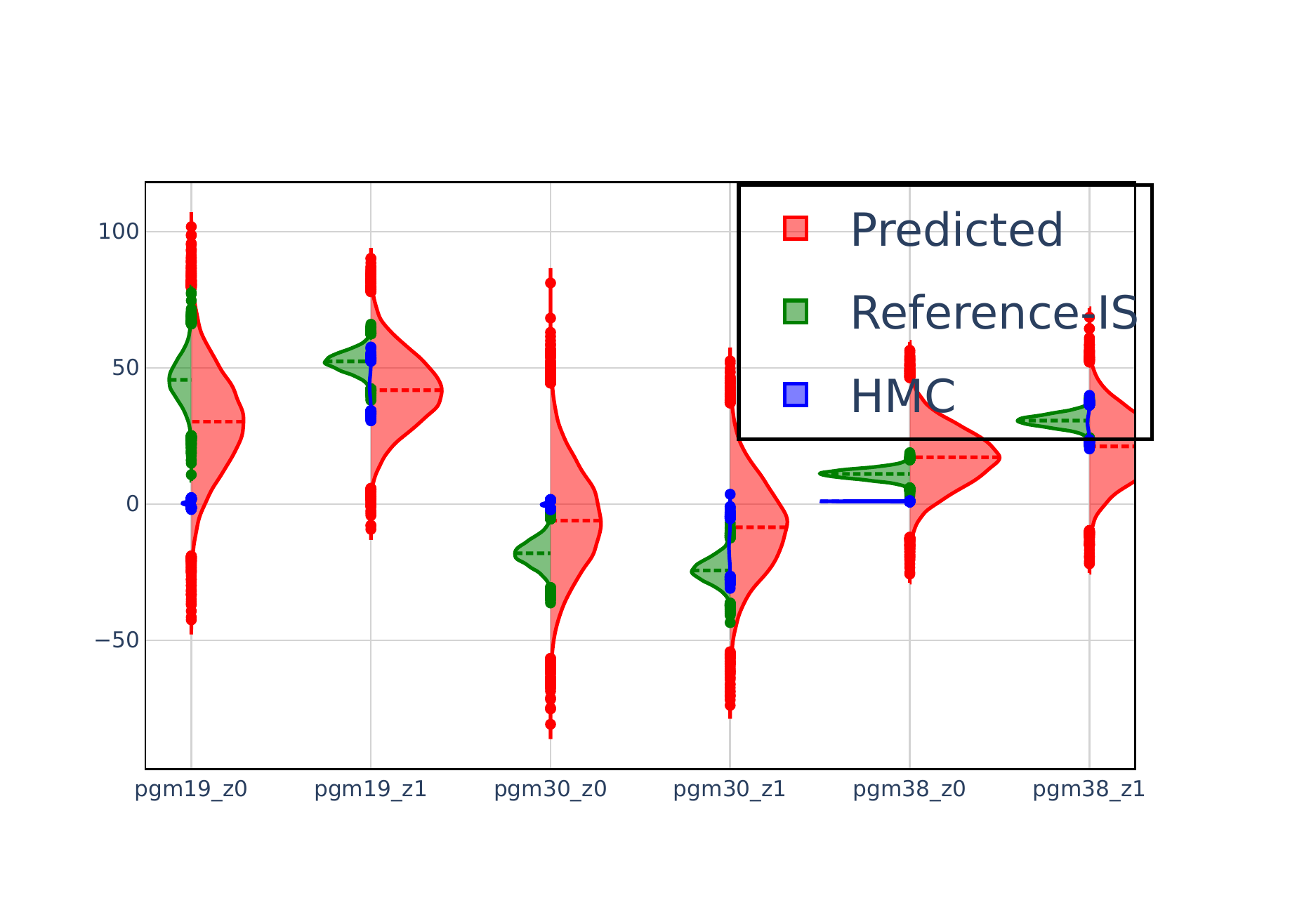}
		\caption{Marginal posteriors}
		\label{fig:mulmod-comparison}
	\end{subfigure}
	\begin{subfigure}{0.245\textwidth}
		\includegraphics[width=\textwidth]
		{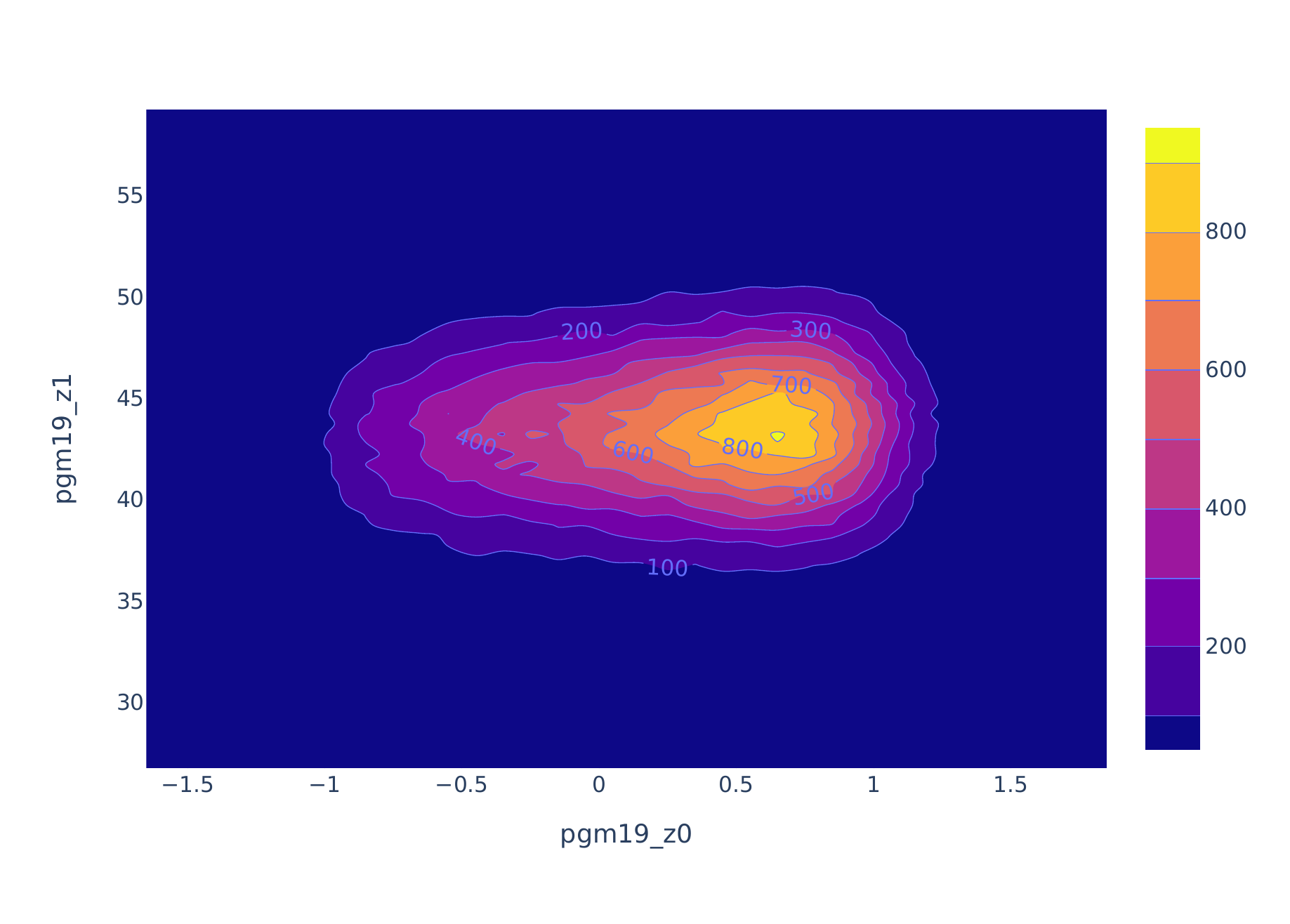}
		\caption{Chain 1}
		\label{fig:mulmod-contour-chain1}
	\end{subfigure}
	\begin{subfigure}{0.245\textwidth}
		\includegraphics[width=\textwidth]
		{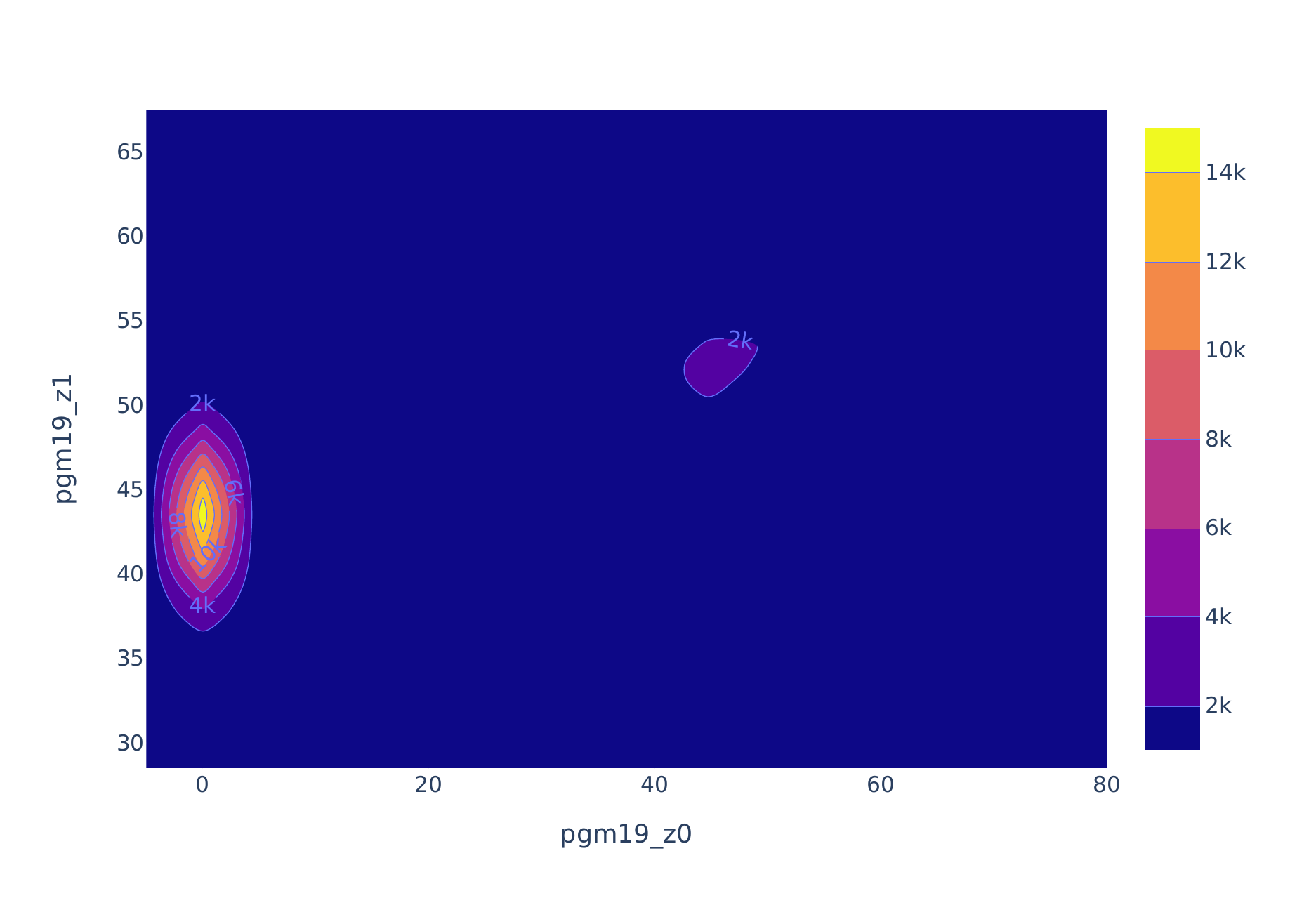}
		\caption{Chain 2}
		\label{fig:mulmod-contour-chain2}
	\end{subfigure}
	\begin{subfigure}{0.245\textwidth}
		\includegraphics[width=\textwidth]
		{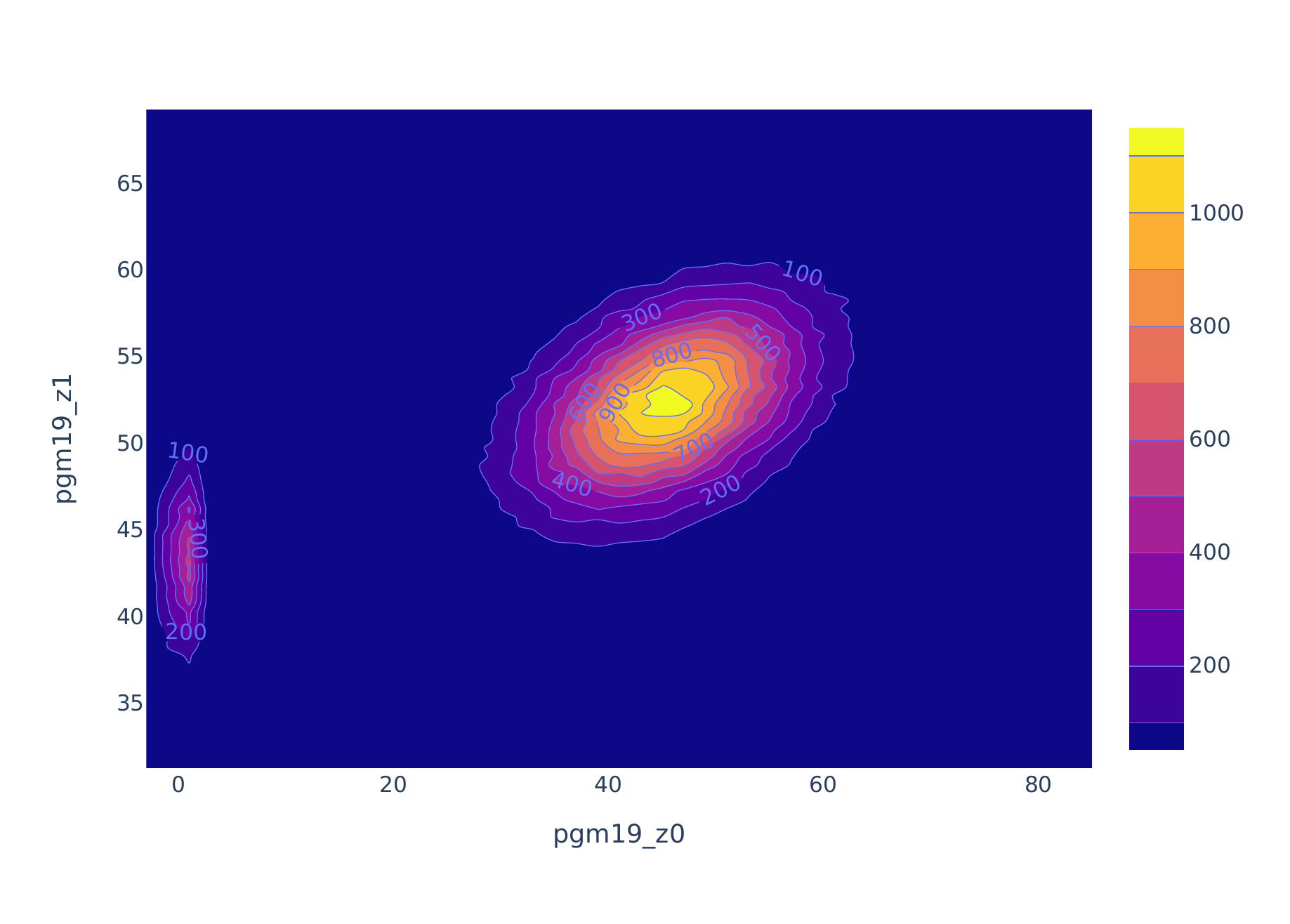}
		\caption{Chain 7}
		\label{fig:mulmod-contour-chain7}
	\end{subfigure}
	\caption{Marginal posteriors for the comparison, and contours of three HMC chains for $\mathsf{pgm\_19}$ where the x-axis is for $\mathsf{z0}$ and the 
		y-axis $\mathsf{z1}$.}
	%\label{fig:mulmod-manual-inspection}
	%\vspace{-3mm}
\end{figure}

%\begin{figure}[t]
%	\vspace{-3mm}
%	\begin{minipage}{0.329\linewidth}
%		\includegraphics[width=\textwidth]{mulmod/res_vis_te_refHMC_pred_refIS_pgmgroup1.pdf}
%		\caption{
%			Comparison with HMC for three test programs from $\mulmod$.
%		}
%		\label{fig:mulmod-comparison}
%	\end{minipage}
%	\hfill
%	\begin{minipage}{0.6\linewidth}
%		\centering
%		\begin{subfigure}{0.32\textwidth}
%			\includegraphics[width=\textwidth]
%			{mulmod/contour_res_nuts_te19_chain1.pdf}
%			\caption{Chain 1}
%			\label{fig:mulmod-contour-chain1}
%		\end{subfigure}
%		\begin{subfigure}{0.32\textwidth}
%			\includegraphics[width=\textwidth]
%			{mulmod/contour_res_nuts_te19_chain2.pdf}
%			\caption{Chain 2}
%			\label{fig:mulmod-contour-chain2}
%		\end{subfigure}
%		\begin{subfigure}{0.42\textwidth}
%			\includegraphics[width=\textwidth]
%			{mulmod/contour_res_nuts_te19_chain4.pdf}
%			\caption{Chain 4}
%			\label{fig:mulmod-contour-chain4}
%		\end{subfigure}
%		\begin{subfigure}{0.32\textwidth}
%			\includegraphics[width=\textwidth]
%			{mulmod/contour_res_nuts_te19_chain7.pdf}
%			\caption{Chain 7}
%			\label{fig:mulmod-contour-chain7}
%		\end{subfigure}
%		\caption{
%			Contours of four HMC chains for $\mathsf{pgm\_19}$. The x-axis is for $\mathsf{z0}$ and the y-axis $\mathsf{z1}$.
%			%The plots reveal the multimodality in the posterior of the program, and so the difficulty that HMC faces.
%		}
%		\label{fig:mulmod-contour}
%	\end{minipage}
%        \vspace{-5mm}
%\end{figure}

\subsection{Test-time efficiency in comparison with alternatives}
\label{sec:mulmod}

We demonstrate the test-time efficiency of our approach using three-variable models ($\mulmod$) where two latent variables follow normal distributions and
the other stores the value of the function $\mm(x) \defeq 100 \times x^3 / (10 + x^4)$. The models are grouped into three types defined by their dependency
graphs and the positions of $\mm$ in the programs (see Fig.\ref{fig:mulmod-types} in the appendix). We ran our meta-algorithm using $600$ programs 
from all three types using importance samples (not HMC samples). Then for $60$ test programs from the last model type, we measured ESS and the sum of second moments
along the wall-clock time using three approaches: importance sampling (IS-pred; ours) with the predicted posteriors as proposal using $70$K samples, importance sampling 
(IS-prior) with prior as proposal using $100$K samples, and HMC with $1$M samples after $500$ warmups. As the reference sampler, we used importance sampling (IS-ref) with prior as proposal using $5$M samples.
All the approaches were repeated $10$ times.

Table~\ref{tab:mulmod-ess-all-test-pgms} shows the average ESS per unit time over the $60$ test programs, by the three approaches. For HMC, ``ESS'' is
the ESS computed using $10$ Markov chains averaged over the $60$ programs, and ``ESS / sec'' is the ESS per unit time, averaged over the programs.
For IS-\{pred, prior\}, ``ESS'' and ``ESS / sec'' are the average ESS and ESS per unit time, respectively, both over the $10$ trials and the $60$ programs.
GM is the geometric mean, and Q1 and Q3 are the first and third quartiles, respectively. We used the geometric mean, since the ESSes had outliers.
The results show that IS-pred achieved the highest ESS per unit time in terms of both mean (GM) and the quartiles (Q1 and Q3).

We provide further analysis for a test program ($\mathsf{pgm19}$; see Appendix~\ref{appendix:mulmod-pgm19}). Fig.~\ref{fig:mulmod-hmc-moments} and 
\ref{fig:mulmod-is-moments} show the moments estimated by HMC and IS-\{pred, prior\}, respectively, in comparison
with the same (across the two figures) reference moments by IS-ref. The estimates by IS-pred (red) quickly converged to the reference (green) within 
$18$ms, while those by HMC (orange)
did not converge even after $84$s. IS-pred (red) and IS-prior (blue) tended to produce better estimates as the elapsed time increased, but each time, IS-pred
estimated the moments more precisely with a smaller variance than IS-prior.
In the same runs of the three approaches as in the last columns of Fig.~\ref{fig:mulmod-hmc-moments} and \ref{fig:mulmod-is-moments},
IS-pred produced over $16$K effective samples in $18$ms, while HMC generated only $80$ effective samples even after $84$s. Similarly,
IS-prior generated fewer than $1.4$K effective samples in the approximately same elapsed time as in IS-pred. In fact, Fig.~\ref{fig:mulmod-is-ess} shows that as the time increases, the gap between the ESSes of IS-pred and IS-prior gets widen, because the former increases at a rate significantly higher than the latter. Note that IS-pred has to scan a program twice at test time, once for computing
the proposal and another for IS with the predicted proposal. See Appendix~\ref{appendix:is-pred-cost} for discussion.

Our manual inspection revealed that the programs in $\mulmod$ often have multimodal posteriors. Fig.~\ref{fig:mulmod-comparison} shows the posteriors
for \{$\mathsf{pgm19}$, $\mathsf{pgm30}$, $\mathsf{pgm38}$\} in the test set, computed by our learnt inference algorithm (without IS), HMC ($200$K samples
after $10$K warmups) , and IS-ref. The variable $\mathsf{z0}$ in the three programs had multimodal posteriors. For $\mathsf{pgm19}$, the learnt inference algorithm 
took only $0.6$ms to compute the posteriors, while HMC took $120$s on average to generate a chain.
The predictions (red) from the learnt inference algorithm for $\mathsf{z0}$
describe  the reference posteriors (green) better than those (blue) by HMC in terms of mean, variance, and mode covering.\footnote{Our inference algorithm in a multimodal-posterior case leads to a good approximation in the following sense: the approximating distribution $q$ covers the regions of the modes well, and also approximates the mean and variance of the target distribution accurately. Note that such a $q$ is useful when it is used as the proposal of an importance sampler.}
The contour plots in Fig.~\labelcref{fig:mulmod-contour-chain1,fig:mulmod-contour-chain2,fig:mulmod-contour-chain7} visualise three
HMC chains for $\mathsf{pgm19}$. Here, HMC failed to converge, and Fig.~\ref{fig:mulmod-contour-chain1} explains the poor estimate (blue) in the first
column of Fig.~\ref{fig:mulmod-comparison}.

\noindent{\bf Limitations and future work}\ \
Currently, a learnt inference algorithm in our work does not generalise to programs with different sizes~\citep{yan20execute}, e.g., from clustering models with
two clusters to those with ten clusters. Each model class assumes a fixed number of variables, and the neural networks crucially exploit the assumption. Also, our meta-algorithm does not scale in practice. When applied to large programs, e.g., state-space models with a few hundred time steps, it cannot learn an
optimal inference algorithm within a reasonable amount of time. Overcoming these limitations is a future work. Another direction that we are considering is to remove the strong independence assumption (via mean field Gaussian) on the 
approximating distribution in our inference algorithm, and to equip the algorithm with the capability of generating an appropriate form of the approximation distribution with rich dependency structure, by, e.g., incorporating the ideas from
\citet{ambrogioni2021automatic}. This direction is closely related to automatic guide generation in Pyro~\citep{bingham2018pyro}.

\noindent{\bf Conclusion}\ \
In this paper, we presented a white-box inference algorithm that computes an approximate posterior and a marginal likelihood estimate by analysing
the given program sequentially using neural networks, and a meta-algorithm that
learns the network parameters over a training set of probabilistic programs. In our experiments, the meta-algorithm learnt an inference
algorithm that generalises well to similar but unseen programs, and the learnt inference algorithm sometimes had
test-time advantages over alternatives. A moral of this work is that the description of a probabilistic model itself has useful information, and learning to extract and exploit the information may lead to an efficient inference. We hope that our work encourages further exploration of this research direction.

%\noindent{\bf Reproducibility statement}\ \
%Our paper provides detailed information that is needed for reproducing the results. In each of \S\ref{sec:interpol}, \S\ref{sec:extrapol}, and \S\ref{sec:mulmod} of the main text, we explain the key experimental design and setup clearly. More detailed experimental setup is in
%Appendix~\ref{appendix:empirical-setup}, where we specify the hyperparameter and the design of the neural networks.
%In Appendix~\ref{appendix:models}, we provide full details of the model classes and how programs from the classes were automatically generated
%in our evaluation.

\section*{Acknowledgments}
This work was supported by the Engineering Research Center Program through the National Research Foundation of Korea (NRF) funded by the Korean Government MSIT (NRF2018R1A5A1059921).

\bibliography{refs}
\bibliographystyle{iclr2022_conference}

\newpage
\appendix
\section{Further discussion about the translation of an expressive PPL into our intermediate language}
\label{appendix:ppl-translation}

Programs with recursion or while loops cannot generally be translated into our intermediate language, since such programs may go into infinite loops while
the programs in our language always terminate. Programs with for loops and general branches can in theory be translated into a less expressive language
such as ours. For example, \citet{van2018introduction} explain a language called FOPPL (Section 2), which has for loops and branches, and the translation
of FOPPL into graphical models (Section 3). We think that these graphical models can be translated into programs in our language. Of course, this does not 
mean that the learnt inference algorithm would interact well with the compilation; the interaction between compilation and inference in the context of
meta-learning is something to be explored in future work.

\begin{figure}[!ht]
	\hrule
	\begin{align*}
	& \mathit{one}:=1;\ t := 2;\ f := 5;\ \mathit{ten} := 10; \\
	& z_1 \sim \mathcal{N}(f,\mathit{ten}); \ \texttt{/\!\!/}\ \text{log of the mass of Milky Way}\\
	& \mathit{mass}_1 := z_1 \times t; \\
	& z_2 \sim \mathcal{N}(\mathit{mass}_1 ,f); \ \texttt{/\!\!/}\ \text{for the first satellite galaxy}\\
	&\code{obs}(\mathcal{N}(z_2,\mathit{one}),\mathit{ten}); \  \texttt{/\!\!/}\ \text{$x_1=10$ for $x_1 \sim \mathcal{N}(z_2,\mathit{one})$} \\
	& \mathit{mass}_2 := z_1 + f; \\
	& z_3 \sim \mathcal{N}(\mathit{mass}_2,t); \ \texttt{/\!\!/}\ \text{for the second satellite galaxy}\\
	& \code{obs}(\mathcal{N}(z_3,\mathit{one}), 3) \ \texttt{/\!\!/}\ \text{$x_2=3$ for $x_2  \sim \mathcal{N}(z_3,\mathit{one})$}
	\end{align*}
	\hrule
	$\,$
	\caption{Milky Way example compiled to the probabilistic programming language used in the paper.}
	\label{fig:example-galaxy-compiled}
\end{figure}

\section{Milky Way example in the probabilistic programming language}
\label{appendix:galaxy-compiled}

Fig.~\ref{fig:example-galaxy-compiled} shows the compiled version of the Milky way example to the probabilistic programming language of the paper.

\section{Formal semantics of the probabilistic programming language}
\label{appendix:semantics}

In \S\ref{sec:setup}, we stated that a program $C$ in our language denotes an unnormalised density $p_C$ that is factorised as follows:
\begin{align*}
& p_C(z_{1:n}) =
        p_C(x_{1:m} = r_{1:m} | z_{1:n})
        \times
        \prod_{i = 1}^n p_C(z_i | z_{1:i-1}).
\end{align*}
Here $z_1,\ldots,z_n$ are all the variables assigned by the sampling statements $z_i \sim \mathcal{N}(\ldots)$ in $C$ in that order, the program $C$ contains $m$ observe statements with observations $r_1,\ldots,r_m$, and these observed random variables are denoted by $x_1,\ldots,x_m$. The goal of this section is to provide the details of our statement. That is, we describe the formal semantics of our probabilistic programming language, and from it, we derive a map from programs $C$ to unnormalised densities $p_C$.  

To define the formal semantics of programs in our language, we need a type system that tracks information about updated variables and observations, and also formalises the syntactic conditions that we imposed informally in \S\ref{sec:setup}. The type system lets us derive the following judgements for programs $C$ and atomic commands $A$:
\[
(S,V,\alpha) \vdash_1 C : (T,W,\beta),
\quad
(S,V,\alpha) \vdash_2 A : (T,W,\beta),
\]
where $S$ and $T$ are sequences of distinct variables, $V$ and $W$ are sets of variables that do not appear in $S$ and $T$, respectively, and $\alpha$ and $\beta$ are sequences of reals. The first judgement says that if before running the program $C$, the latent variables in $S$ are sampled in that order, the program variables in $V$ are updated by non-sample statements, and the real values in the sequence $\alpha$ are observed in that order, then running $C$ changes these three data to $T$, $W$, and $\beta$. The second judgement means the same thing except that we consider the execution of $A$, instead of $C$. The triples $(S,V,\alpha)$ and $(T,W,\beta)$ serve as types in this type system.

The rules for deriving the judgements for $C$ and $A$ follow from the intended meaning just explained. We show these rules below, using  the notation $@$ for the concatenation operator for two sequences and also $\mathrm{set}(S)$ for the set of elements in the sequence $S$:
\begin{align*}
&
\infer{
(R,U,\alpha) \vdash_1 (C_1;C_2) : (T,W,\gamma)
}{
(R,U,\alpha) \vdash_1 C_1 : (S,V,\beta)
&
(S,V,\beta) \vdash_1 C_2 : (T,W,\gamma)
}
\qquad
\infer{
(S,V,\alpha) \vdash_1 A : (T,W,\beta)
}{
(S,V,\alpha) \vdash_2 A : (T,W,\beta)
}
\\[2.5ex]
& 
\infer{ 
(S,V,\alpha) \vdash_2 (z \sim \mathcal{N}(v_1,v_2)) : (S@[z], V, \alpha) 
}{
z \not\in \mathrm{set}(S) \cup V
&
v_1,v_2 \in \mathrm{set}(S) \cup V}
\qquad
\infer{ 
(S,V,\alpha) \vdash_2 \code{obs}(\mathcal{N}(v_0,v_1),r) : (S, V,\alpha @ [r]) 
}{
v_0,v_1 \in \mathrm{set}(S) \cup V}
\\[2.5ex]
& \qquad\qquad\qquad
\infer{ 
(S,V,\alpha) \vdash_2 (v_0 := \code{if}\ (v_1 > v_2)\ v_3\ \code{else}\ v_4)
: (S, V \cup \{v_0\},\alpha) 
}{
v_0 \not\in \mathrm{set}(S) \cup V
&
v_1,v_2,v_3,v_4 \in \mathrm{set}(S) \cup V}
\\[2.5ex]
& 
\infer{ 
(S,V,\alpha) \vdash_2 (v_0 := r) : (S,V \cup \{v_0\},\alpha)
}{
v_0 \not\in \mathrm{set}(S) \cup V}
 \qquad
\infer{ 
(S,V,\alpha) \vdash_2 (v_0 := v_1) : (S,V \cup \{v_0\},\alpha) 
}{
v_0 \not\in \mathrm{set}(S) \cup V
&
v_1 \in \mathrm{set}(S) \cup V}
\\[2.5ex]
& \qquad\qquad\qquad\qquad\qquad
\infer{ 
(S,V,\alpha) \vdash_2 (v_0 := p(v_1,v_2)) : (S,V \cup \{v_0\},\alpha)
}{
v_0 \not\in \mathrm{set}(S) \cup V
&
v_1,v_2 \in \mathrm{set}(S) \cup V}
\end{align*}

We now define our semantics, which specifies mappings from judgements for $C$ and $A$ to mathematical entities. First, we interpret each type $(S,V,\alpha)$ as a set, and it is denoted by 
$\db{(S,V,\alpha)}$:
\begin{align*}
\db{(S,V,\alpha)} \defeq 
\{(p,f,l) \,\mid\, {}
&
p \text{ is a (normalised) density on $\R^{|S|}$, } 
\
f = (f_v)_{v \in \mathrm{set}(S) \cup V},
\\
&
\text{each $f_v$ is a measurable map from $\R^{|S|}$ to $\R$,} 
\\
&
l \text{ is a measurable function from $\R^{|S|} \times \R^{|\alpha|}$ to $\R_+$}\},
\end{align*}
where $|S|$ and $|\alpha|$ are the lengths of the sequences $S$ and $\alpha$, and $\R_+$ means the set of positive reals. Next, we define the semantics of the judgements $(S,V,\alpha) \vdash_1 C : (T,W,\beta)$ and
$(S,V,\alpha) \vdash_2 A : (T,W,\beta)$ that can be derived by the rules from above. The formal semantics of these
judgements, denoted by the $\db{-}$ notation, are maps of the following type:
\begin{align*}
\db{(S,V,\alpha) \vdash_1 C : (T,W,\beta)} & : \db{(S,V,\alpha)} \to \db{(T,W,\beta)},
\\
\db{(S,V,\alpha) \vdash_2 A : (T,W,\beta)} & : \db{(S,V,\alpha)} \to \db{(T,W,\beta)}.
\end{align*}
The semantics is given by induction on the size of the derivation of each judgement, under the assumption that for each procedure name $p \in \mathbb{P}$, we have its interpretation as a measurable map from $\R^2$ to $\R$:
\[
\db{p} : \R^2 \to \R.
\]
We spell out the semantics below,
first the one for programs and next that for atomic commands.
\begin{align*}
\db{(S,V,\alpha) \vdash_1 A : (T,W,\beta)}(p,f,l)
& {} \defeq 
\db{(S,V,\alpha) \vdash_2 A : (T,W,\beta)}(p,f,l),
\\
\db{(R,U,\alpha) \vdash_1 (C_1;C_2) : (T,W,\gamma)}(p,f,l)
& {} \defeq 
(\db{(S,V,\beta) \vdash_2 C_2 : (T,W,\gamma)}
\\
& \qquad\qquad
{} \circ
\db{(R,U,\alpha) \vdash_2 C_1 : (S,V,\beta)})(p,f,l).
\end{align*}
Let $\mathcal{N}(a;b,c)$ be the density of the normal distribution with mean $b$
and variance $c$ when $c > 0$ and $1$ when $c \leq 0$. For a family of functions $f = (f_v)_{v \in V}$,
a variable $w \not\in V$, and a function $f'_w$, we write $f\oplus f'_w$ for the extension of $f$ with a 
new $w$-indexed member $f'_w$. 
\begin{align*}
& 
\db{(S,V,\alpha) \vdash_2 z \sim \mathcal{N}(v_1,v_2) : (S@[z], V, \alpha)}(p,f,l) 
\defeq (p',f',l') 
\\
& \qquad
(\text{where}\ p'(a_{1:|S|+1}) \defeq p(a_{1:|S|})
\times  \mathcal{N}(a_{|S|+1};f_{v_1}(a_{1:|S|}),f_{v_2}(a_{1:|S|})), 
\\
& \phantom{\qquad \text{where}\ }
f'_v(a_{1:|S|+1}) \defeq f_v(a_{1:|S|})\ \text{for all $v \in V$},
\ \,
f'_{z}(a_{1:|S|+1}) \defeq a_{|S|+1},
\\
& \phantom{\qquad \text{where}\ }
l'(a_{1:|S|+1},b_{1:|\alpha|}) \defeq l(a_{1:|S|},b_{1:|\alpha|})),
\\[2ex]
& 
\db{(S,V,\alpha) \vdash_2 \code{obs}(\mathcal{N}(v_0,v_1),r) : (S, V,\alpha @ [r])}(p,f,l) 
\defeq (p,f,l') 
\\
& \qquad
(\text{where}\ l'(a_{1:|S|},b_{1:|\alpha|+1}) \defeq l(a_{1:|S|},b_{1:|\alpha|})
\times  \mathcal{N}(b_{|\alpha|+1};f_{v_1}(a_{1:|S|}),f_{v_2}(a_{1:|S|})), 
\\[2ex]
&
\llbracket (S,V,\alpha) \vdash_2 (v_0 := \code{if}\ (v_1 > v_2)\ v_3\ 
 \code{else}\ v_4) : (S, V \cup \{v_0\},\alpha)\rrbracket(p,f,l)
 \defeq (p,f \oplus f'_{v_0},l) 
\\
& \qquad
(\text{where}\ f'_{v_0}(a_{1:|S|}) \defeq \mathrm{if}\ (f_{v_1}(a_{1:|S|}) > f_{v_2}(a_{1:|S|}))\
\mathrm{then}\ f_{v_3}(a_{1:|S|})\ \mathrm{else}\ f_{v_4}(a_{1:|S|})),
\\[2ex]
& 
\db{(S,V,\alpha) \vdash_2 (v_0 := r) : (S, V \cup \{v_0\},\alpha)}(p,f,l) {}
\defeq (p,f \oplus f'_{v_0},l) 
\\
&\qquad (\text{where}\ f'_{v_0}(a_{1:|S|}) \defeq r),
\\[2ex]
& 
\db{(S,V,\alpha) \vdash_2 (v_0 := v_1) : (S, V \cup \{v_0\},\alpha)}(p,f,l)
 \defeq (p,f \oplus f'_{v_0},l) 
\\
& \qquad
(\text{where}\ f'_{v_0}(a_{1:|S|}) \defeq f_{v_1}(a_{1:|S|})),
\\[2ex]
& 
\db{(S,V,\alpha) \vdash_2 (v_0 := p'(v_0,v_1)) : (S, V \cup \{v_0\},\alpha)}(p,f,l) 
\defeq (p,f \oplus f'_{v_0},l) 
\\
& \qquad
(\text{where}\ f'_{v_0}(a_{1:|S|}) \defeq \db{p'}(f_{v_0}(a_{1:|S|}), f_{v_1}(a_{1:|S|}))).
\end{align*}
Finally, we define $p_C$ for the well-initialised well-typed programs $C$, i.e., programs $C$ for which we can derived
\[
([],\emptyset,[]) \vdash_1 C : (S,V,\alpha).
\]
For such a $C$, the definition of $p_C$ is given below:
\[
p_C(z_{1:|S|}) = p(z_{1:|S|}) \times l(z_{1:|S|},\alpha)
\]
where $(p,\_,l) = \db{([],\emptyset,[]) \vdash_1 C : (S,V,\alpha)}(p_0,f_0,l_0)$ for the constant-$1$ functions $p_0$ and $l_0$ of appropriate types and the empty family $f_0$ of functions.

\section{Marginal likelihood computation: derivation and correctness}
\label{appendix:marginal-likelihood}

Let $x_n$ be the random variable (RV) that is observed by the command $\code{obs}(\mathcal{N}(v_0, v_1), r)$ and $x_{1:(n-1)}$ be the $(n-1)$ RVs that are
observed before the command. When our algorithm is about to analyse this observe command, we have (an estimate of) $p(x_{1:(n-1)})$ by induction. Then,
the marginal likelihood of $x_{1:n}$ can be computed as follows:
\begin{align*}
& p(x_{1:(n-1)}, x_n) \\
&= \iint p(x_{1:(n-1)}, x_n, v_0, v_1)\, dv_0\, dv_1 \\
&= \iint p(x_{1:(n-1)})\, p(v_0, v_1 | x_{1:(n-1)})\, p(x_n | x_{1:(n-1)}, v_0, v_1)\, dv_0\, dv_1 \\
&\approx p(x_{1:(n-1)}) \iint p_h(v_0, v_1)\, p(x_n | x_{1:(n-1)}, v_0, v_1)\, dv_0\, dv_1 \,\, \\
&\qquad \qquad \text{// The filtering distribution $p(v_0, v_1 | x_{1:(n-1)})$ is approximated by $p_h$.} \\
&= p(x_{1:(n-1)}) \iint p_h(v_0, v_1)\, p(x_n | v_0, v_1)\, dv_0\, dv_1 \,\, \\
&\qquad \qquad \text{// The RV $x_n$ is conditionally independent of $x_{1:(n-1)}$ given $v_0, v_1$.} \\
&= p(x_{1:(n-1)}) \iint p_h(v_0, v_1)\, \mathcal{N}(r; v_0, v_1)\, dv_0\, dv_1 \\
&\qquad \qquad \text{// $p(x_{1:(n-1)})$ is $Z$ in the description of $\postinfer(A_i)$ in Section~\ref{sec:learning}, and the neural network} \\
&\qquad \qquad \text{// $\nn_{\integral,\phi_7}$ aims at approximating the integral term accurately.}
\end{align*}
This derivation leads to the equation in the main text.

In a setting of probabilistic programming where observations are allowed to be different in true and false branches, the marginal likelihood may fail to
be defined, and such a setting is beyond the scope of our language. Using variables multiple times or having observe commands spread out in the program 
does not make differences in the derivation above.

\begin{table*}
\caption{Full list of the model classes in the empirical evaluation.}
\label{tab:full-list}
%\vspace{}
%\small
\centering
\begin{adjustbox}{max width=\textwidth}
\begin{tabular}{|c|l|p{6.5cm}|p{2.1cm}|}
\hline
Section & \multicolumn{1}{c|}{Model class} & \multicolumn{1}{c|}{Description} & \multicolumn{1}{c|}{Detail} \\ \hline
\multirow{6}{*}{\S\ref{sec:interpol}}& $\gauss$ & Gaussian models with a latent variable and an observation where the mean of the Gaussian likelihood is an affine transformation of the latent.& Appendix~\ref{appendix:gauss} \\ \cline{2-4}
& $\hierl$ & Hierarchical models with three hierarchically structured latent variables. & Appendix~\ref{appendix:hierl} \\ \cline{2-4}
& $\hierd$ & Hierarchical or multi-level models with both latent variables and data structured hierarchically where data are modelled as a regression of latent variables of different levels. & Appendix~\ref{appendix:hierd} \\ \cline{2-4}
& $\cluster$ & Clustering models where five observations are clustered into two groups. & Appendix~\ref{appendix:cluster} \\ \cline{2-4}
& $\milky$ and $\milkyo$& Milky Way models, and their multiple-observations extension where five observations are made for each satellite galaxy. & Appendix~\ref{appendix:milky} \\ \cline{2-4}
& $\rbk$ & Models with the Rosenbrock function, which is expressed as an external procedure. & Appendix~\ref{appendix:rb} \\ \hline
\multirow{2}{*}{\S\ref{sec:extrapol}}& $\ext1$ & Models with three Gaussian variables and one deterministic variable storing the value of the function $\nl(x) = 50 / \pi \times \arctan(x/10)$, where the models have $12$ different types --- four different dependency graphs of the variables, and three different positions of the deterministic $\nl$ variable for each of these graphs. & Appendix~\ref{appendix:ext1} (and Fig.~\ref{fig:ext1-types}) \\ \cline{2-4}
& $\ext2$ & Models with six Gaussian variables and one $\nl$ variable, which are grouped into five model types based on their dependency graphs. & Appendix~\ref{appendix:ext2} (and Fig.~\ref{fig:ext2-types}) \\ \hline
\multirow{1}{*}{\S\ref{sec:mulmod}}& $\mulmod$ & Three-variable models where two latent variables follow normal distributions and the other stores the value of the function $\mm(x) \defeq 100 \times x^3 / (10 + x^4)$. The models in this class are grouped into three types defined by their dependency graphs and
the positions of $\mm$ in the programs. & Appendix~\ref{appendix:mulmod} (and Fig.~\ref{fig:mulmod-types}) \\ \hline
\end{tabular}
\end{adjustbox}
\end{table*}

\section{Detailed descriptions for probabilistic models used in the empirical evaluation}
\label{appendix:models}

Table~\ref{tab:full-list} shows the full list of the model classes that we considered in our empirical evaluation (\S\ref{sec:empirical}).
We detail the program specifications for the classes using the
probabilistic programming language in \S\ref{sec:setup}, and then describe how our program generator generated programs from those classes randomly.

In the program specifications to follow, randomly-generated constants are written in the Greek alphabets ($\theta$), and latent and other program
variables in the English alphabets. Also, we often use more intuitive variable names instead of using $z_i$ for latent variables and
$v_i$ for the other program variables, to improve readability.
When describing random generation of the parameter values, we let $\mathrm{U}(a,b)$ denote the uniform distribution whose domain is $(a,b)
\subset \R$; we use this only for describing the random program generation process itself, not the generated programs (only normal 
distributions are used in our programs, with the notation $\mathcal{N}$).

\subsection{Generalisation to new model parameters and observations}
This section details the model classes in \S\ref{sec:interpol}.

\subsubsection{$\gauss$}
\label{appendix:gauss}
The model class is described as follows:
\begin{align*}
& m_z := \theta_1;\ v_z := \theta_2';\ c_1 := \theta_3;\ c_2 := \theta_4;\ v_x := \theta_5'; \\
& z_1 \sim \mathcal{N}(m_z,v_z);\ z_2 := z_1 \times c_1;\ z_3 := z_2 + c_2; \\
& \code{obs}(\mathcal{N}(z_3,v_x),o) \\
\end{align*}
For each program of the class, our random program generator generated the parameter values as follows:
\begin{align*}
& \theta_1 \sim \mathrm{U}(-5,5),\ \theta_2 \sim \mathrm{U}(0,20),\ \theta_2' = (\theta_2)^2,\ \theta_3 \sim \mathrm{U}(-3,3) \\
& \theta_4 \sim \mathrm{U}(-10,10),\ \theta_5 \sim \mathrm{U}(0.5, 10),\ \theta_5' = (\theta_5)^2
\end{align*}
and then generated the observation $o$ by running the program forward where the value for $z_1$ was sampled from
$z_1 \sim \mathrm{U}(m_z - 2 \times \sqrt{v_z}, m_z + 2 \times \sqrt{v_z})$.

\subsubsection{$\hierl$}
\label{appendix:hierl}
The model class is described as follows:
\begin{align*}
& m_g := \theta_1;\ v_g := \theta_2';\ v_{t_1} := \theta_3';\ v_{t_2} := \theta_4';\ v_{x_1} := \theta_5'; \\
& v_{x_2} := \theta_6';\ g \sim \mathcal{N}(m_g,v_g);\  t_1 \sim \mathcal{N}(g,v_{t_1});\ t_2 \sim \mathcal{N}(g,v_{t_2}); \\
& \code{obs}(\mathcal{N}(t_1,v_{x_1}),o_1);\ \code{obs}(\mathcal{N}(t_2,v_{x_2}),o_2)
\end{align*}
For each program of the class, our generator generated the parameter values as follows:
\begin{align*}
& \theta_1 \sim \mathrm{U}(-5,5),\ \theta_2 \sim \mathrm{U}(0,50),\ \theta_2' = (\theta_2)^2,\ \theta_3 \sim \mathrm{U}(0,10) \\
& \theta_3' = (\theta_3)^2,\ \theta_4 \sim \mathrm{U}(0,10),\ \theta_4' = (\theta_4)^2,\ \theta_5 \sim \mathrm{U}(0.5,10) \\
& \theta_5' = (\theta_5)^2,\ \theta_6 \sim \mathrm{U}(0.5,10),\ \theta_6' = (\theta_6)^2
\end{align*}
and then generated the observations $o_1$ and $o_2$ by running the program (i.e., simulating the model) forward.

\subsubsection{$\hierd$}
\label{appendix:hierd}
The model class is described as follows:
\begin{align*}
& m_{a_0} := \theta_1;\ v_{a_0} := \theta_2';\ v_{a_1} := \theta_3';\ v_{a_2} := \theta_4';\ m_b := \theta_5; \\
& v_b := \theta_6';\ d_1 = \theta_7;\ d_2 = \theta_8;\ v_{x_1} := \theta_9';\ v_{x_2} := \theta_{10}'; \\
& a_0 \sim \mathcal{N}(m_{a_0}, v_{a_0});\ a_1 \sim \mathcal{N}(a_0, v_{a_1});\ a_2 \sim \mathcal{N}(a_0, v_{a_2}); \\
& b \sim \mathcal{N}(m_b, v_b); \\
& t_1 := b \times d_1;\ t_2 := a_1 + t_1;\ \code{obs}(\mathcal{N}(t_2,v_{x_1}), o_1); \\
& t_3 := b \times d_2;\ t_4 := a_2 + t_3;\ \code{obs}(\mathcal{N}(t_4,v_{x_2}), o_2)
\end{align*}
For each program of the class, our generator generated the parameter values as follows:
\begin{align*}
& \theta_1 \sim \mathrm{U}(-10,10),\ \theta_2 \sim \mathrm{U}(0,100),\ \theta_2' = (\theta_2)^2,\ \theta_3 \sim \mathrm{U}(0,10) \\
& \theta_3' = (\theta_3)^2,\ \theta_4 \sim \mathrm{U}(0,10),\ \theta_4' = (\theta_4)^2,\ \theta_5 \sim \mathrm{U}(-5,5) \\
& \theta_6 \sim \mathrm{U}(0,10),\ \theta_6' = (\theta_6)^2,\ \theta_7 \sim \mathrm{U}(-5,5),\ \theta_8 \sim \mathrm{U}(-5,5) \\
& \theta_9 \sim \mathrm{U}(0.5,10),\ \theta_9' = (\theta_9)^2,\ \theta_{10} \sim \mathrm{U}(0.5,10),\ \theta_{10}' = (\theta_{10})^2
\end{align*}
and then generated the observations $o_1$ and $o_2$ by running the program forward where the values for $a_0$, $a_1$, $a_2$, and $b$
in this specific simulation were sampled as follows:
\begin{align*}
& a_0 \sim \mathrm{U}(m_{a_0} - 2 \times \sqrt{v_{a_0}},\ m_{a_0} + 2 \times \sqrt{v_{a_0}}) \\
& a_1 \sim \mathrm{U}(a_0 - 2 \times \sqrt{v_{a_1}},\ a_0 + 2 \times \sqrt{v_{a_1}}) \\
& a_2 \sim \mathrm{U}(a_0 - 2 \times \sqrt{v_{a_2}},\ a_0 + 2 \times \sqrt{v_{a_2}}) \\
& b \sim \mathrm{U}(m_b - 2 \times \sqrt{v_b},\ m_b + 2 \times \sqrt{v_b})
\end{align*}

\subsubsection{$\cluster$}
\label{appendix:cluster}
The model class is described as follows:
\begin{align*}
& m_{g_1} := \theta_1;\ v_{g_1} := \theta_2';\ m_{g_2} := \theta_3;\ v_{g_2} := \theta_4';\ v_x := \theta_5'; \\
& g_1 \sim \mathcal{N}(m_{g_1}, v_{g_1});\ g_2 \sim \mathcal{N}(m_{g_2}, v_{g_2}); \\
& \mathit{zero} := 0;\ \mathit{hund} := 100; \\
& t_1 \sim \mathcal{N}(\mathit{zero},\mathit{hund});\ m_1 := \code{if}\ (t_1 > \mathit{zero})\ g_1\ \code{else}\ g_2; \\
& \code{obs}(\mathcal{N}(m_1, v_x), o_1); \\
& t_2 \sim \mathcal{N}(\mathit{zero},\mathit{hund});\ m_2 := \code{if}\ (t_2 > \mathit{zero})\ g_1\ \code{else}\ g_2; \\
& \code{obs}(\mathcal{N}(m_2, v_x), o_2); \\
& t_3 \sim \mathcal{N}(\mathit{zero},\mathit{hund});\ m_3 := \code{if}\ (t_3 > \mathit{zero})\ g_1\ \code{else}\ g_2; \\
& \code{obs}(\mathcal{N}(m_3, v_x), o_3); \\
& t_4 \sim \mathcal{N}(\mathit{zero},\mathit{hund});\ m_4 := \code{if}\ (t_4 > \mathit{zero})\ g_1\ \code{else}\ g_2; \\
& \code{obs}(\mathcal{N}(m_4, v_x), o_4); \\
& t_5 \sim \mathcal{N}(\mathit{zero},\mathit{hund});\ m_5 := \code{if}\ (t_5 > \mathit{zero})\ g_1\ \code{else}\ g_2; \\
& \code{obs}(\mathcal{N}(m_5, v_x), o_5)
\end{align*}
For each program of the class, our generator generated the parameter values as follows:
\begin{align*}
& \theta_1 \sim \mathrm{U}(-15,15),\ \theta_2 \sim \mathrm{U}(0.5,50),\ \theta_2' = (\theta_2)^2 \\
& \theta_3 \sim \mathrm{U}(-15,15),\ \theta_4 \sim \mathrm{U}(0.5,50),\ \theta_4' = (\theta_4)^2 \\
& \theta_5 \sim \mathrm{U}(0.5,10),\ \theta_5' = (\theta_5)^2
\end{align*}
and then generated the observations $o_{1:5}$ by running the program forward.

\subsubsection{$\milky$ and $\milkyo$}
\label{appendix:milky}
The model class $\milky$ is described as follows:
\begin{align*}
& m_{\mathit{mass}} := \theta_1;\ v_{\mathit{mass}} := \theta_2';\ c_1 := \theta_3;\ v_{g_1} := \theta_4';\ c_2 := \theta_5; \\
& v_{g_2} := \theta_6';\ v_{x_1} := \theta_7';\ v_{x_2} := \theta_8'; \\
& \mathit{mass} \sim \mathcal{N}(m_{\mathit{mass}}, v_{\mathit{mass}}); \\
& \mathit{mass}_1 := \mathit{mass} \times c_1;\ g_1 \sim \mathcal{N}(\mathit{mass}_1,v_{g_1}); \\
& \mathit{mass}_2 := \mathit{mass} + c_2;\ g_2 \sim \mathcal{N}(\mathit{mass}_2,v_{g_2}); \\
& \code{obs}(\mathcal{N}(g_1,v_{x_1}), o_1);\ \code{obs}(\mathcal{N}(g_2,v_{x_2}), o_2)
\end{align*}
For each program of $\milky$, our generator generated the parameter values as follows:
\begin{align*}
& \theta_1 \sim \mathrm{U}(-10,10),\ \theta_2 \sim \mathrm{U}(0,30),\ \theta_2' = (\theta_2)^2,\ \theta_3 \sim \mathrm{U}(-2,2) \\
& \theta_4 \sim \mathrm{U}(0,10),\ \theta_4' = (\theta_4)^2,\ \theta_5 \sim \mathrm{U}(-5,5),\ \theta_6 \sim \mathrm{U}(0,10) \\
& \theta_6' = (\theta_6)^2,\ \theta_7 \sim \mathrm{U}(0.5,10),\ \theta_7' = (\theta_7)^2,\ 
\theta_8 \sim \mathrm{U}(0.5,10) \\
& \theta_8' = (\theta_8)^2
\end{align*}
and then generated the observations $o_1$ and $o_2$ by running the program forward.

Everything remained the same for the $\milkyo$ class, except that the two $\code{obs}$ commands were extended to
$\code{obs}(\mathcal{N}(g_1, v_{x_1}), [o_1,o_2,o_3,o_4,o_5])$ and
$\code{obs}(\mathcal{N}(g_2, v_{x_2}), [o_6,o_7,o_8,o_9,o_{10}])$, respectively,
and all the observations were generated similarly by running the extended model forward.

\subsubsection{$\rbk$}
\label{appendix:rb}
The model class $\rbk$ is described as follows:
\begin{align*}
& m_{z_1} := \theta_1;\ v_{z_1} := \theta_2';\ m_{z_2} := \theta_3;\ v_{z_2} := \theta_4';\ v_x := \theta_5'; \\
& z_1 \sim \mathcal{N}(m_{z_1}, v_{z_1});\ z_2 \sim \mathcal{N}(m_{z_2}, v_{z_2});\ r := \mathrm{Rosenbrock}(z_1,z_2); \\
& \code{obs}(\mathcal{N}(r,v_x), o)
\end{align*}
where $\mathrm{Rosenbrock}(z_1,z_2) = 0.05 \times (z_1 - 1)^2 + 0.005 \times (z_2 - {z_1}^2)^2$.
For each program of the class, our generator generated the parameter values as follows:
\begin{align*}
& \theta_1 \sim \mathrm{U}(-8,8),\ \theta_2 \sim \mathrm{U}(0,5),\ \theta_2' = (\theta_2)^2,\ \theta_3 \sim \mathrm{U}(-8,8) \\
& \theta_4 \sim \mathrm{U}(0,5),\ \theta_4' = (\theta_4)^2,\ \theta_5 \sim \mathrm{U}(0.5,10),\ \theta_5' = (\theta_5)^2
\end{align*}
and then generated the observation $o$ by running the program forward where the values
for $z_1$ and $z_2$ in this specific simulation were sampled as follows:
\begin{align*}
& z_1 \sim \mathrm{U}(m_{z_1} - 1.5 \times \sqrt{v_{z_1}},\ m_{z_1} + 1.5 \times \sqrt{v_{z_1}}) \\
& z_2 \sim \mathrm{U}(m_{z_2} - 1.5 \times \sqrt{v_{z_2}},\ m_{z_2} + 1.5 \times \sqrt{v_{z_2}})
\end{align*}

\subsection{Generalisation to new model structures}
This section details the model classes, and different types in each model class in \S\ref{sec:extrapol}. For readability, we present canonicalised 
dependency graphs where variables are named in the breadth-first order. In the experiments reported in this section, we used a minor extension of our probabilistic programming language with procedures taking one parameter.

\begin{figure}[t]
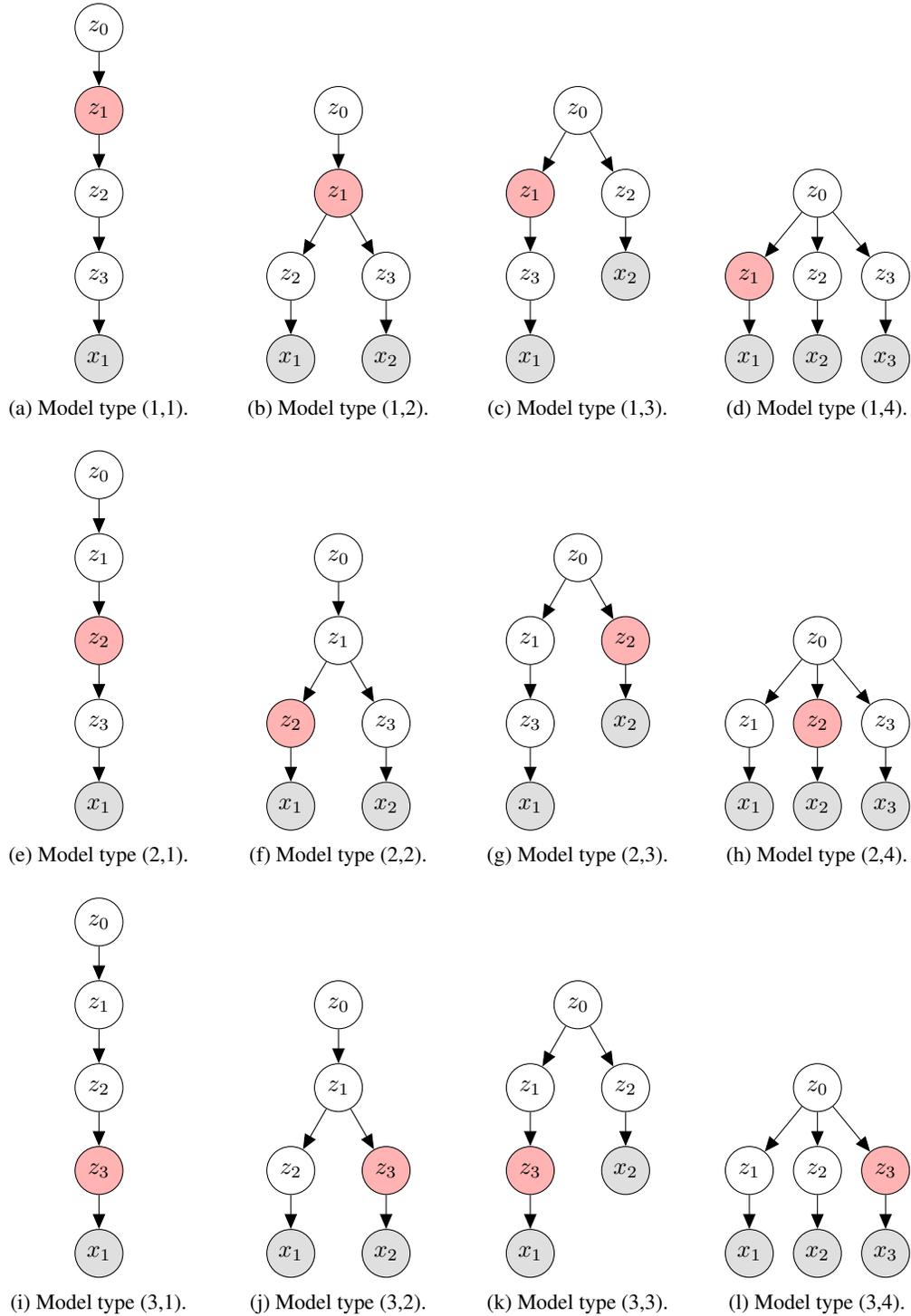

% ext1; 4 dependency graphs for 1st position of the nl variable.
\begin{subfigure}[b]{0.245\textwidth}
\centering
\tikz{
	% nodes
	\node[latent](z0){$z_0$};
	\node[latent, below=.5cm of z0, fill=red!30](z1){$z_1$};
	\node[latent, below=.5cm of z1](z2){$z_2$};
	\node[latent, below=.5cm of z2](z3){$z_3$};
	\node[obs, below=.5cm of z3](x1){$x_1$};
	% edges
	\edge{z0}{z1};
	\edge{z1}{z2};
	\edge{z2}{z3};
	\edge{z3}{x1};
}
\caption{Model type (1,1).}
\end{subfigure}
\begin{subfigure}[b]{0.245\textwidth}
\centering
\tikz{
	% nodes
	\node[latent](z0){$z_0$};
	\node[latent, below=.5cm of z0, fill=red!30](z1){$z_1$};
	\node[latent, below=.5cm of z1, xshift=-.7cm](z2){$z_2$};
	\node[latent, below=.5cm of z1, xshift=.7cm](z3){$z_3$};
	\node[obs, below=.5cm of z2](x1){$x_1$};
	\node[obs, below=.5cm of z3](x2){$x_2$};
	% edges
	\edge{z0}{z1};
	\edge{z1}{z2};
	\edge{z1}{z3};
	\edge{z2}{x1};
	\edge{z3}{x2};
}
\caption{Model type (1,2).}
\end{subfigure}
\begin{subfigure}[b]{0.245\textwidth}
\centering
\tikz{
	% nodes
	\node[latent](z0){$z_0$};
	\node[latent, below=.5cm of z0, xshift=-.7cm, fill=red!30](z1){$z_1$};
	\node[latent, below=.5cm of z0, xshift=.7cm](z2){$z_2$};
	\node[latent, below=.5cm of z1](z3){$z_3$};
	\node[obs, below=.5cm of z3](x1){$x_1$};
	\node[obs, below=.5cm of z2](x2){$x_2$};
	% edges
	\edge{z0}{z1};
	\edge{z0}{z2};
	\edge{z1}{z3};
	\edge{z3}{x1};
	\edge{z2}{x2};
}
\caption{Model type (1,3).}
\end{subfigure}
\begin{subfigure}[b]{0.245\textwidth}
\centering
\tikz{
	% nodes
	\node[latent](z0){$z_0$};
	\node[latent, below=.5cm of z0, xshift=-1cm, fill=red!30](z1){$z_1$};
	\node[latent, below=.5cm of z0](z2){$z_2$};
	\node[latent, below=.5cm of z0, xshift=1cm](z3){$z_3$};
	\node[obs, below=.5cm of z1](x1){$x_1$};
	\node[obs, below=.5cm of z2](x2){$x_2$};
	\node[obs, below=.5cm of z3](x3){$x_3$};
	% edges
	\edge{z0}{z1};
	\edge{z0}{z2};
	\edge{z0}{z3};
	\edge{z1}{x1};
	\edge{z2}{x2};
	\edge{z3}{x3};
}
\caption{Model type (1,4).}
\end{subfigure}
\vspace{0mm}

% ext1; 4 dependency graphs for 2nd position of the nl variable.
\begin{subfigure}[b]{0.245\textwidth}
\centering
\tikz{
	% nodes
	\node[latent](z0){$z_0$};
	\node[latent, below=.5cm of z0](z1){$z_1$};
	\node[latent, below=.5cm of z1, fill=red!30](z2){$z_2$};
	\node[latent, below=.5cm of z2](z3){$z_3$};
	\node[obs, below=.5cm of z3](x1){$x_1$};
	% edges
	\edge{z0}{z1};
	\edge{z1}{z2};
	\edge{z2}{z3};
	\edge{z3}{x1};
}
\caption{Model type (2,1).}
\end{subfigure}
\begin{subfigure}[b]{0.245\textwidth}
\centering
\tikz{
	% nodes
	\node[latent](z0){$z_0$};
	\node[latent, below=.5cm of z0](z1){$z_1$};
	\node[latent, below=.5cm of z1, xshift=-.7cm, fill=red!30](z2){$z_2$};
	\node[latent, below=.5cm of z1, xshift=.7cm](z3){$z_3$};
	\node[obs, below=.5cm of z2](x1){$x_1$};
	\node[obs, below=.5cm of z3](x2){$x_2$};
	% edges
	\edge{z0}{z1};
	\edge{z1}{z2};
	\edge{z1}{z3};
	\edge{z2}{x1};
	\edge{z3}{x2};
}
\caption{Model type (2,2).}
\end{subfigure}
\begin{subfigure}[b]{0.245\textwidth}
\centering
\tikz{
	% nodes
	\node[latent](z0){$z_0$};
	\node[latent, below=.5cm of z0, xshift=-.7cm](z1){$z_1$};
	\node[latent, below=.5cm of z0, xshift=.7cm, fill=red!30](z2){$z_2$};
	\node[latent, below=.5cm of z1](z3){$z_3$};
	\node[obs, below=.5cm of z3](x1){$x_1$};
	\node[obs, below=.5cm of z2](x2){$x_2$};
	% edges
	\edge{z0}{z1};
	\edge{z0}{z2};
	\edge{z1}{z3};
	\edge{z3}{x1};
	\edge{z2}{x2};
}
\caption{Model type (2,3).}
\end{subfigure}
\begin{subfigure}[b]{0.245\textwidth}
\centering
\tikz{
	% nodes
	\node[latent](z0){$z_0$};
	\node[latent, below=.5cm of z0, xshift=-1cm](z1){$z_1$};
	\node[latent, below=.5cm of z0, fill=red!30](z2){$z_2$};
	\node[latent, below=.5cm of z0, xshift=1cm](z3){$z_3$};
	\node[obs, below=.5cm of z1](x1){$x_1$};
	\node[obs, below=.5cm of z2](x2){$x_2$};
	\node[obs, below=.5cm of z3](x3){$x_3$};
	% edges
	\edge{z0}{z1};
	\edge{z0}{z2};
	\edge{z0}{z3};
	\edge{z1}{x1};
	\edge{z2}{x2};
	\edge{z3}{x3};
}
\caption{Model type (2,4).}
\end{subfigure}
\vspace{0mm}

% ext1; 4 dependency graphs for 3rd position of the nl variable.
\begin{subfigure}[b]{0.245\textwidth}
\centering
\tikz{
	% nodes
	\node[latent](z0){$z_0$};
	\node[latent, below=.5cm of z0](z1){$z_1$};
	\node[latent, below=.5cm of z1](z2){$z_2$};
	\node[latent, below=.5cm of z2, fill=red!30](z3){$z_3$};
	\node[obs, below=.5cm of z3](x1){$x_1$};
	% edges
	\edge{z0}{z1};
	\edge{z1}{z2};
	\edge{z2}{z3};
	\edge{z3}{x1};
}
\caption{Model type (3,1).}
\end{subfigure}
\begin{subfigure}[b]{0.245\textwidth}
\centering
\tikz{
	% nodes
	\node[latent](z0){$z_0$};
	\node[latent, below=.5cm of z0](z1){$z_1$};
	\node[latent, below=.5cm of z1, xshift=-.7cm](z2){$z_2$};
	\node[latent, below=.5cm of z1, xshift=.7cm, fill=red!30](z3){$z_3$};
	\node[obs, below=.5cm of z2](x1){$x_1$};
	\node[obs, below=.5cm of z3](x2){$x_2$};
	% edges
	\edge{z0}{z1};
	\edge{z1}{z2};
	\edge{z1}{z3};
	\edge{z2}{x1};
	\edge{z3}{x2};
}
\caption{Model type (3,2).}
\end{subfigure}
\begin{subfigure}[b]{0.245\textwidth}
\centering
\tikz{
	% nodes
	\node[latent](z0){$z_0$};
	\node[latent, below=.5cm of z0, xshift=-.7cm](z1){$z_1$};
	\node[latent, below=.5cm of z0, xshift=.7cm](z2){$z_2$};
	\node[latent, below=.5cm of z1, fill=red!30](z3){$z_3$};
	\node[obs, below=.5cm of z3](x1){$x_1$};
	\node[obs, below=.5cm of z2](x2){$x_2$};
	% edges
	\edge{z0}{z1};
	\edge{z0}{z2};
	\edge{z1}{z3};
	\edge{z3}{x1};
	\edge{z2}{x2};
}
\caption{Model type (3,3).}
\end{subfigure}
\begin{subfigure}[b]{0.245\textwidth}
\centering
\tikz{
	% nodes
	\node[latent](z0){$z_0$};
	\node[latent, below=.5cm of z0, xshift=-1cm](z1){$z_1$};
	\node[latent, below=.5cm of z0](z2){$z_2$};
	\node[latent, below=.5cm of z0, xshift=1cm, fill=red!30](z3){$z_3$};
	\node[obs, below=.5cm of z1](x1){$x_1$};
	\node[obs, below=.5cm of z2](x2){$x_2$};
	\node[obs, below=.5cm of z3](x3){$x_3$};
	% edges
	\edge{z0}{z1};
	\edge{z0}{z2};
	\edge{z0}{z3};
	\edge{z1}{x1};
	\edge{z2}{x2};
	\edge{z3}{x3};
}
\caption{Model type (3,4).}
\end{subfigure}
\vspace{0mm}

\caption{
	Canonicalised dependency graphs for all $12$ model types in $\ext1$. The rows are for different positions of the $\nl$ variable, and the columns are
	for different dependency graphs: the model type ($i$,$j$) means one of the $12$ model types in $\ext1$ that corresponds to the $i$-th position of the
	$\nl$ variable and $j$-th dependency graph.
}
\label{fig:ext1-types}
\end{figure}

\subsubsection{$\ext1$}
\label{appendix:ext1}
Fig.~\ref{fig:ext1-types} shows the dependency graphs for all model types in $\ext1$. The variables $z_0,z_1,\ldots$ and $x_1,x_2,\ldots$ represent latent 
and observed variables, respectively, and observed variables are colored in gray. The red node in each graph represents the position of the $\nl$ variable. 

%\gc{Seems minor: in our language, $p$ takes two variables, but $\nl$ is a single-variable function. \hy{I wrote a sentence for this.}}

Our program generator in this case generates programs from the whole model class $\ext1$; it generates programs of all twelve different types in $\ext1$.
%Our program generator in this case generates programs of all twelve different types in $\ext1$.
We explain this generation process for
the model type (1,1) in Fig.~\ref{fig:ext1-types}, while pointing out that the similar process is applied to the other eleven
types. To generate programs of the model type (1,1), we use the following program template:
\begin{align*}
& m_{z_0} := \theta_1;\ v_{z_0} := \theta_2';\ v_{z_2} := \theta_3';\ v_{z_3} := \theta_4';\ v_{x_1} := \theta_5'; \\
& z_0 \sim \mathcal{N}(m_{z_0},v_{z_0});\ z_1 := \nl(z_0);\ z_2 \sim \mathcal{N}(z_1,v_{z_2});\ z_3 \sim \mathcal{N}(z_2,v_{z_3}); \\
& \code{obs}(\mathcal{N}(z_3,v_{x_1}),o_1)
\end{align*}
where $\nl(z) = 50 / \pi \times \arctan(z/10)$. The generation involves randomly sampling the parameters of this template, converting the template into a program in our language, and creating synthetic observations. Specifically, our generator generates the parameter values as follows:
\begin{align*}
& \theta_1 \sim \mathrm{U}(-5,5),\ \theta_2 \sim \mathrm{U}(0,20),\ \theta_2' = (\theta_2)^2,\ \theta_3 \sim \mathrm{U}(0,20),\ \theta_3' = (\theta_3)^2 \\
& \theta_4 \sim \mathrm{U}(0,20),\ \theta_4' = (\theta_4)^2,\ \theta_5 \sim \mathrm{U}(0.5,10),\ \theta_5' = (\theta_5)^2
\end{align*}
and generates the observation $o_1$ by running the program forward where the values for $z_{0:3}$ in this specific simulation were sampled (and fixed
to specific values) as follows:
\begin{align*}
& z_0 \sim \mathrm{U}(m_{z_0} - 2 \times \sqrt{v_{z_0}},\ m_{z_0} + 2 \times \sqrt{v_{z_0}})\\
& z_1 = \nl(z_0)\\
& z_2 \sim \mathrm{U}(z_1 - 2 \times \sqrt{v_{z_2}},\ z_1 + 2 \times \sqrt{v_{z_2}})\\
& z_3 \sim \mathrm{U}(z_2 - 2 \times \sqrt{v_{z_3}},\ z_2 + 2 \times \sqrt{v_{z_3}}).
\end{align*}
The generator uses different templates for the other eleven model types in $\ext1$, while
sharing the similar process for generation of the parameters and observations.
%The model descriptions and the random program generations are specified similarly for the other model types in $\ext1$, with different dependency graphs
%and positions of the $\nl$ variable.

\begin{figure}[t]
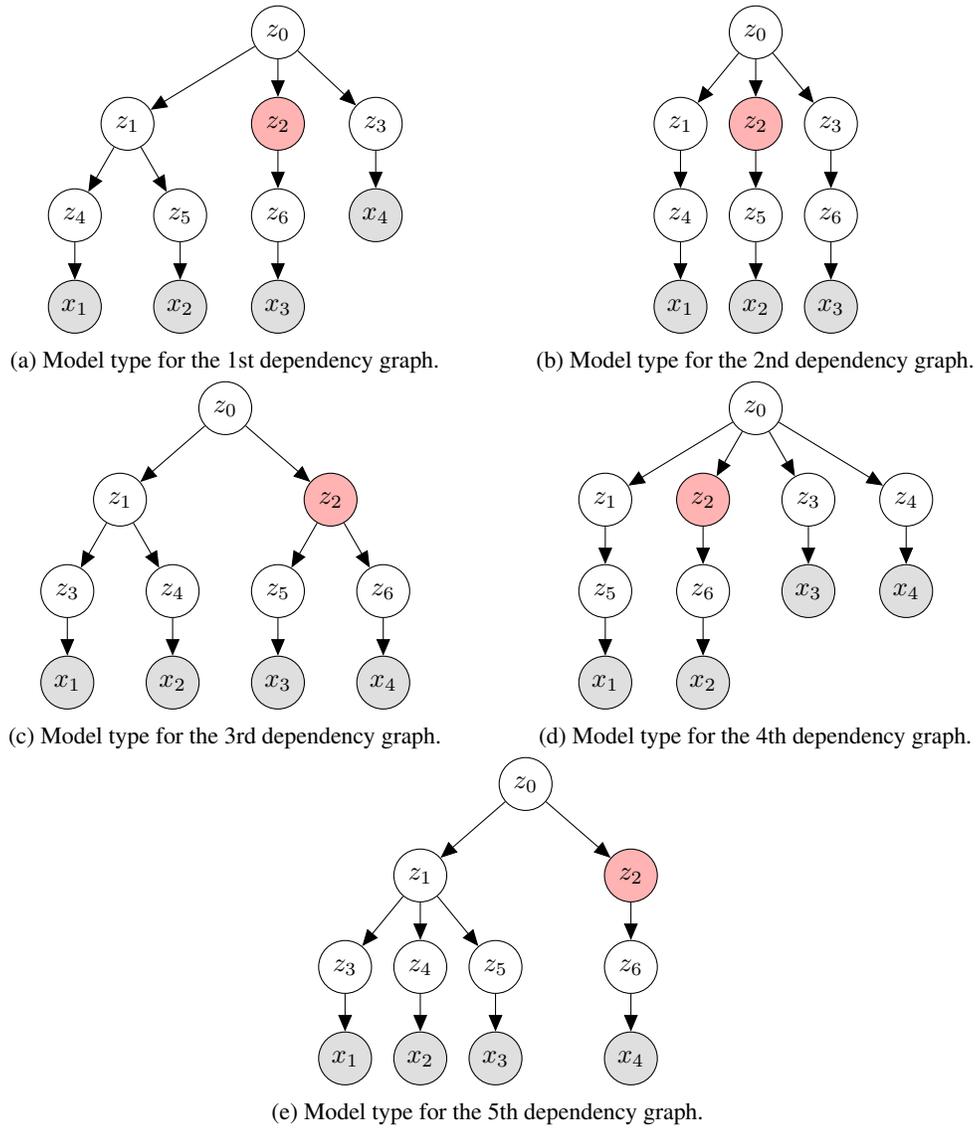

% ext2; 1st dependency graph
\begin{subfigure}{0.5\textwidth}
\centering
\tikz{
	% nodes
	\node[latent](z0){$z_0$};
	\node[latent, below=.5cm of z0, xshift=-2cm](z1){$z_1$};
	\node[latent, below=.5cm of z0, fill=red!30](z2){$z_2$};
	\node[latent, below=.5cm of z0, xshift=1.3cm](z3){$z_3$};
	\node[latent, below=.5cm of z1, xshift=-.7cm](z4){$z_4$};
	\node[latent, below=.5cm of z1, xshift=.7cm](z5){$z_5$};
	\node[latent, below=.5cm of z2](z6){$z_6$};
	\node[obs, below=.5cm of z4](x1){$x_1$};
	\node[obs, below=.5cm of z5](x2){$x_2$};
	\node[obs, below=.5cm of z6](x3){$x_3$};
	\node[obs, below=.5cm of z3](x4){$x_4$};
	% edges	
	\edge{z0}{z1};
	\edge{z0}{z2};
	\edge{z0}{z3};
	\edge{z1}{z4};
	\edge{z1}{z5};
	\edge{z2}{z6};
	\edge{z4}{x1};
	\edge{z5}{x2};
	\edge{z6}{x3};
	\edge{z3}{x4};
}
\caption{Model type for the 1st dependency graph.}
\end{subfigure}
% ext2; 2nd dependency graph
\begin{subfigure}{0.5\textwidth}
\centering
\tikz{
	% nodes
	\node[latent](z0){$z_0$};
	\node[latent, below=.5cm of z0, xshift=-1cm](z1){$z_1$};
	\node[latent, below=.5cm of z0, fill=red!30](z2){$z_2$};
	\node[latent, below=.5cm of z0, xshift=1cm](z3){$z_3$};
	\node[latent, below=.5cm of z1](z4){$z_4$};
	\node[latent, below=.5cm of z2](z5){$z_5$};
	\node[latent, below=.5cm of z3](z6){$z_6$};
	\node[obs, below=.5cm of z4](x1){$x_1$};
	\node[obs, below=.5cm of z5](x2){$x_2$};
	\node[obs, below=.5cm of z6](x3){$x_3$};
	% edges	
	\edge{z0}{z1};
	\edge{z0}{z2};
	\edge{z0}{z3};
	\edge{z1}{z4};
	\edge{z2}{z5};
	\edge{z3}{z6};
	\edge{z4}{x1};
	\edge{z5}{x2};
	\edge{z6}{x3};
}
\caption{Model type for the 2nd dependency graph.}
\end{subfigure}
\vspace{0mm}

% ext2; 3rd dependency graph
\begin{subfigure}{0.5\textwidth}
\centering
\tikz{
	% nodes
	\node[latent](z0){$z_0$};
	\node[latent, below=.5cm of z0, xshift=-1.4cm](z1){$z_1$};
	\node[latent, below=.5cm of z0, xshift=1.4cm, fill=red!30](z2){$z_2$};
	\node[latent, below=.5cm of z1, xshift=-.7cm](z3){$z_3$};
	\node[latent, below=.5cm of z1, xshift=.7cm](z4){$z_4$};
	\node[latent, below=.5cm of z2, xshift=-.7cm](z5){$z_5$};
	\node[latent, below=.5cm of z2, xshift=.7cm](z6){$z_6$};
	\node[obs, below=.5cm of z3](x1){$x_1$};
	\node[obs, below=.5cm of z4](x2){$x_2$};
	\node[obs, below=.5cm of z5](x3){$x_3$};
	\node[obs, below=.5cm of z6](x4){$x_4$};
	% edges	
	\edge{z0}{z1};
	\edge{z0}{z2};
	\edge{z1}{z3};
	\edge{z1}{z4};
	\edge{z2}{z5};
	\edge{z2}{z6};
	\edge{z3}{x1};
	\edge{z4}{x2};
	\edge{z5}{x3};
	\edge{z6}{x4};
}
\caption{Model type for the 3rd dependency graph.}
\end{subfigure}
% ext2; 4th dependency graph
\begin{subfigure}{0.5\textwidth}
\centering
\tikz{
	% nodes
	\node[latent](z0){$z_0$};
	\node[latent, below=.5cm of z0, xshift=-2cm](z1){$z_1$};
	\node[latent, below=.5cm of z0, xshift=-.7cm, fill=red!30](z2){$z_2$};
	\node[latent, below=.5cm of z0, xshift=.7cm](z3){$z_3$};
	\node[latent, below=.5cm of z0, xshift=2cm](z4){$z_4$}; //
	\node[latent, below=.5cm of z1](z5){$z_5$};
	\node[latent, below=.5cm of z2](z6){$z_6$};
	\node[obs, below=.5cm of z5](x1){$x_1$};
	\node[obs, below=.5cm of z6](x2){$x_2$};
	\node[obs, below=.5cm of z3](x3){$x_3$};
	\node[obs, below=.5cm of z4](x4){$x_4$};
	% edges
	\edge{z0}{z1};
	\edge{z0}{z2};
	\edge{z0}{z3};
	\edge{z0}{z4};
	\edge{z1}{z5};
	\edge{z2}{z6};
	\edge{z5}{x1};
	\edge{z6}{x2};
	\edge{z3}{x3};
	\edge{z4}{x4};
}
\caption{Model type for the 4th dependency graph.}
\end{subfigure}
\vspace{0mm}

\centering
% ext2; 5th dependency graph
\begin{subfigure}{0.5\textwidth}
\centering
\tikz{
	% nodes
	\node[latent](z0){$z_0$};
	\node[latent, below=.5cm of z0, xshift=-1.4cm](z1){$z_1$};
	\node[latent, below=.5cm of z0, xshift=1.4cm, fill=red!30](z2){$z_2$};
	\node[latent, below=.5cm of z1, xshift=-1cm](z3){$z_3$};
	\node[latent, below=.5cm of z1](z4){$z_4$}; //
	\node[latent, below=.5cm of z1, xshift=1cm](z5){$z_5$};
	\node[latent, below=.5cm of z2](z6){$z_6$};
	\node[obs, below=.5cm of z3](x1){$x_1$};
	\node[obs, below=.5cm of z4](x2){$x_2$};
	\node[obs, below=.5cm of z5](x3){$x_3$};
	\node[obs, below=.5cm of z6](x4){$x_4$};
	% edges
	\edge{z0}{z1};
	\edge{z0}{z2};
	\edge{z1}{z3};
	\edge{z1}{z4};
	\edge{z1}{z5};
	\edge{z2}{z6};
	\edge{z3}{x1};
	\edge{z4}{x2};
	\edge{z5}{x3};
	\edge{z6}{x4};
}
\caption{Model type for the 5th dependency graph.}
\end{subfigure}
\caption{
	Canonicalised dependency graphs for all five model types in $\ext2$.
}
\label{fig:ext2-types}
\end{figure}

\subsubsection{$\ext2$}
\label{appendix:ext2}
Fig.~\ref{fig:ext2-types} shows the dependency graphs for all five model types in $\ext2$. Programs of these five types are randomly generated by our program generator. As in the $\ext1$ case, we explain the generator only for one model type, which corresponds to the first dependency graph in Fig.~\ref{fig:ext2-types}. To generate programs of this type, we use the following program template:
\begin{align*}
& m_{z_0} := \theta_1;\ v_{z_0} := \theta_2';\ v_{z_1} := \theta_3';\ v_{z_3} := \theta_4';\ v_{z_4} := \theta_5';\ v_{z_5} := \theta_6';\
v_{z_6} := \theta_7'; \\
& v_{x_1} := \theta_8';\ v_{x_2} := \theta_9';\ v_{x_3} := \theta_{10}';\ v_{x_4} := \theta_{11}'; \\
& z_0 \sim \mathcal{N}(m_{z_0},v_{z_0});\ z_1 \sim \mathcal{N}(z_0,v_{z_1});\ z_2 := \nl(z_0);\ z_3 \sim \mathcal{N}(z_0,v_{z_3}); \\
& z_4 \sim \mathcal{N}(z_1,v_{z_4});\ z_5 \sim \mathcal{N}(z_1,v_{z_5});\ z_6 \sim \mathcal{N}(z_2,v_{z_6}); \\
& \code{obs}(\mathcal{N}(z_4,v_{x_1}),o_1);\ \code{obs}(\mathcal{N}(z_5,v_{x_2}),o_2);\ \code{obs}(\mathcal{N}(z_6,v_{x_3}),o_3);\
\code{obs}(\mathcal{N}(z_3,v_{x_4}),o_4)
\end{align*}
In order to generate a program of this model type and observations, our generator instantiates the parameters of the template as follows:
\begin{align*}
& \theta_1 \sim \mathrm{U}(-5,5),\ \theta_2 \sim \mathrm{U}(0,10),\ \theta_2' = (\theta_2)^2,\ \theta_3 \sim \mathrm{U}(0,10),\ \theta_3' = (\theta_3)^2,\
\theta_4 \sim \mathrm{U}(0,10),\ \theta_4' = (\theta_4)^2 \\
& \theta_5 \sim \mathrm{U}(0,10),\ \theta_5' = (\theta_5)^2,\ \theta_6 \sim \mathrm{U}(0,10),\ \theta_6' = (\theta_6)^2,\ \theta_7 \sim \mathrm{U}(0,10),\ 
\theta_7' = (\theta_7)^2 \\
& \theta_8 \sim \mathrm{U}(0,10),\ \theta_8' = (\theta_8)^2,\ \theta_9 \sim \mathrm{U}(0,10),\ \theta_9' = (\theta_9)^2,\
\theta_{10} \sim \mathrm{U}(0,10),\ \theta_{10}' = (\theta_{10})^2 \\
& \theta_{11} \sim \mathrm{U}(0,10),\ \theta_{11}' = (\theta_{11})^2.
\end{align*}
Then, it generates the observations $o_{1:4}$ by running the program forward where the values for $z_{0:6}$ in this specific simulation were sampled (and fixed to specific values) as follows:
\begin{align*}
& z_0 \sim \mathrm{U}(m_{z_0} - 2 \times \sqrt{v_{z_0}},\ m_{z_0} + 2 \times \sqrt{v_{z_0}}) \\
& z_1 \sim \mathrm{U}(z_0 - 2 \times \sqrt{v_{z_1}},\ z_0 + 2 \times \sqrt{v_{z_1}}) \\
& z_2 = \nl(z_0) \\
& z_3 \sim \mathrm{U}(z_0 - 2 \times \sqrt{v_{z_3}},\ z_0 + 2 \times \sqrt{v_{z_3}}) \\
& z_4 \sim \mathrm{U}(z_1 - 2 \times \sqrt{v_{z_4}},\ z_1 + 2 \times \sqrt{v_{z_4}}) \\
& z_5 \sim \mathrm{U}(z_1 - 2 \times \sqrt{v_{z_5}},\ z_1 + 2 \times \sqrt{v_{z_5}}) \\
& z_6 \sim \mathrm{U}(z_2 - 2 \times \sqrt{v_{z_6}},\ z_2 + 2 \times \sqrt{v_{z_6}}).
\end{align*}
The generator uses different templates for the other four model types in $\ext2$, while
sharing the similar process for generation of the parameters and observations.
%The model descriptions and the random program generations are specified similarly for the other model types in $\ext2$, except that they have different
%dependency graphs.

\begin{figure}[t]
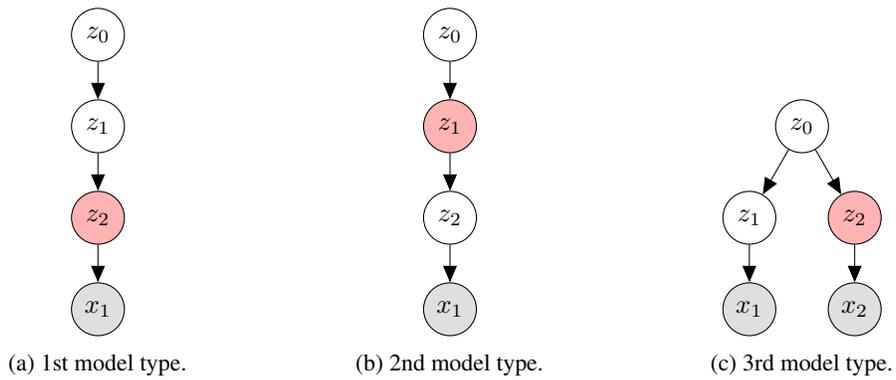

\begin{subfigure}[b]{0.33\textwidth}
\centering
\tikz{
	% nodes
	\node[latent](z0){$z_0$};
	\node[latent, below=.5cm of z0](z1){$z_1$};
	\node[latent, below=.5cm of z1, fill=red!30](z2){$z_2$};
	\node[obs, below=.5cm of z2](x1){$x_1$};
	% edges
	\edge{z0}{z1};
	\edge{z1}{z2};
	\edge{z2}{x1};
}
\caption{1st model type.}
\end{subfigure}
\begin{subfigure}[b]{0.33\textwidth}
\centering
\tikz{
	% nodes
	\node[latent](z0){$z_0$};
	\node[latent, below=.5cm of z0, fill=red!30](z1){$z_1$};
	\node[latent, below=.5cm of z1](z2){$z_2$};
	\node[obs, below=.5cm of z2](x1){$x_1$};
	% edges
	\edge{z0}{z1};
	\edge{z1}{z2};
	\edge{z2}{x1};
}
\caption{2nd model type.}
\end{subfigure}
\begin{subfigure}[b]{0.33\textwidth}
\centering
\tikz{
	% nodes
	\node[latent](z0){$z_0$};
	\node[latent, below=.5cm of z0, xshift=-.7cm](z1){$z_1$};
	\node[latent, below=.5cm of z0, xshift=.7cm, fill=red!30](z2){$z_2$};
	\node[obs, below=.5cm of z1](x1){$x_1$};
	\node[obs, below=.5cm of z2](x2){$x_2$};
	% edges
	\edge{z0}{z1};
	\edge{z0}{z2};
	\edge{z1}{x1};
	\edge{z2}{x2};
}
\caption{3rd model type.}
\end{subfigure}
\caption{Canonicalised dependency graphs for all three model types in the $\mulmod$ class.}
\label{fig:mulmod-types}
\end{figure}

\subsection{Test-time efficiency in comparison with alternatives}
\label{appendix:mulmod}
This section details the $\mulmod$ class in \S\ref{sec:mulmod}, which has three different model types. Fig.~\ref{fig:mulmod-types} shows the dependency 
graphs for all 
the model types. The red node in each graph represents the position of the $\mm$ variable. We used all the three types in training, applied the learnt
inference algorithm to programs in the third model type, and compared the results with those returned by HMC.

We similarly explain the generator only using the model type corresponding to the first dependency graph in Fig.~\ref{fig:mulmod-types}. To generate 
programs of this type, we use the following program template:
%The model type for the first dependency graph in Fig.~\ref{fig:mulmod-types} is described as follows:
\begin{align*}
& m_{z_0} := \theta_1;\ v_{z_0} := \theta_2';\ v_{z_1} := \theta_3';\ v_{x_1} := \theta_4'; \\
& z_0 \sim \mathcal{N}(m_{z_0},v_{z_0});\ z_1 \sim \mathcal{N}(z_0,v_{z_1});\ z_2 := \mm(z_1);\ \code{obs}(\mathcal{N}(z_2,v_{x_1}),o_1)
\end{align*}
where $\mm(x) \defeq 100 \times x^3 / (10 + x^4)$. For each program in this model type, our generator instantiates the parameter values as follows:
\begin{align*}
& \theta_1 \sim \mathrm{U}(-5,5),\ \theta_2 \sim \mathrm{U}(0,20),\ \theta_2' = (\theta_2)^2,\ \theta_3 \sim \mathrm{U}(0,20),\ \theta_3' = (\theta_3)^2 \\
& \theta_4 \sim \mathrm{U}(0.5,10),\ \theta_4' = (\theta_4)^2
\end{align*}
and synthesises the observation $o_1$ by running the program forward where the values for $z_{0:2}$ in this specific simulation were sampled (and fixed to
specific values) as follows:
\begin{align*}
& z_0 \sim \mathrm{U}(m_{z_0} - 2 \times \sqrt{v_{z_0}},\ m_{z_0} + 2 \times \sqrt{v_{z_0}}) \\
& z_1 \sim \mathrm{U}(z_0 - 2 \times \sqrt{v_{z_1}},\ z_0 + 2 \times \sqrt{v_{z_1}}) \\
& z_2 = \mm(z_1).
\end{align*}
The generator uses different templates for the other two model types in $\mulmod$, while
sharing the similar process for instantiation of the parameters and observations.
%The model descriptions and the random program generations are specified similarly for the other model types, except that they have different dependency graphs
%and positions of the $\mm$ variable.

\section{Detailed evaluation setup}
\label{appendix:empirical-setup}

In our evaluation, the dimension $s$ of the internal state $h$ was $10$ (i.e., $h \in \R^{10}$). We used the same neural network 
architecture for all the neural network components of our inference algorithm $\postinfer$.
Each neural network had three linear layers and used the $\tanh$ activation. The hidden dimension was
$10$ for each layer in all the neural networks except for $ \nn_\decode$ where the hidden dimensions were $50$.
The hyper-parameter in our optimisation objective (\S\ref{sec:learning}) was set to $\lambda = 2$ in the evaluation.
For HMC, we used the NUTS sampler~\citep{hoffman14pnut}. We did not use GPUs.

Before running our inference algorithm, we canonicalise the names of variables in a given program based on its dependency (i.e., data-flow) graph.  Although not perfect, this preprocessing %step 
removes a superficial difference between programs caused by different variable names, and enables us to avoid unnecessary complexity caused by variable-renaming symmetries at training and inference times.

%\begin{figure}[t]
%\centering
%\includegraphics[width=0.85\columnwidth]
%{ICML21_rb_losses.pdf}
%\caption{Training and test losses for $\rbk$ until 4K epochs. The $x$ and $y$ axes are log-scaled.}
%\label{fig:rb-loss}
%\end{figure}

%\section{Training and Test Losses for $\rbk$}
%\label{appendix:rb-losses}
%Fig.~\ref{fig:rb-loss} shows the training and test losses for $\rbk$ from three experiments with different random seeds, recorded
%until 4K epochs. Similar to the other models, the decrease of training losses led to reduction the test losses, and the surges of the
%test losses in the later epochs were due to a few test programs in each experiment while the majority of the test losses remained small.

%The rise of the average test losses in later epochs may indicate overfitting. One practical attempt to avoid overfitting is to
%further split the training set into training and validation sets, and use early stopping~\citep{goodfellow2016deep}.
%Alternative approaches include introducing a regularization term in the learning objective 
%explicitly~\citep{bishop2006pattern,goodfellow2016deep}. We will give more systematic treatment in future work.

\begin{figure*}[t]
	\centering
	\begin{subfigure}{0.45\textwidth}
		\includegraphics[width=\textwidth]
		{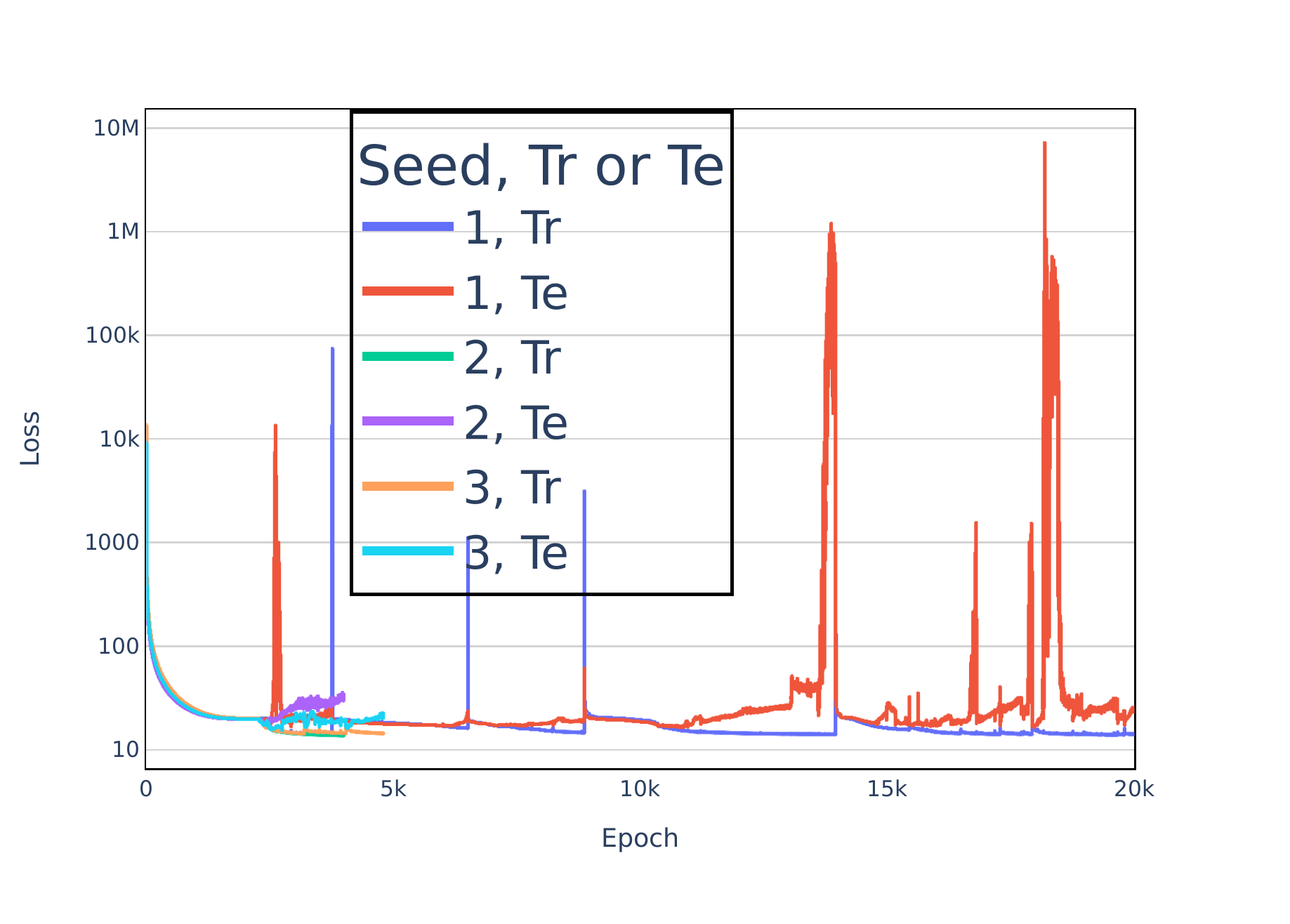}
		\caption{$\hierd$}
		\label{fig:hierd-loss}
	\end{subfigure}
	\begin{subfigure}{0.45\textwidth}
		\includegraphics[width=\textwidth]
		{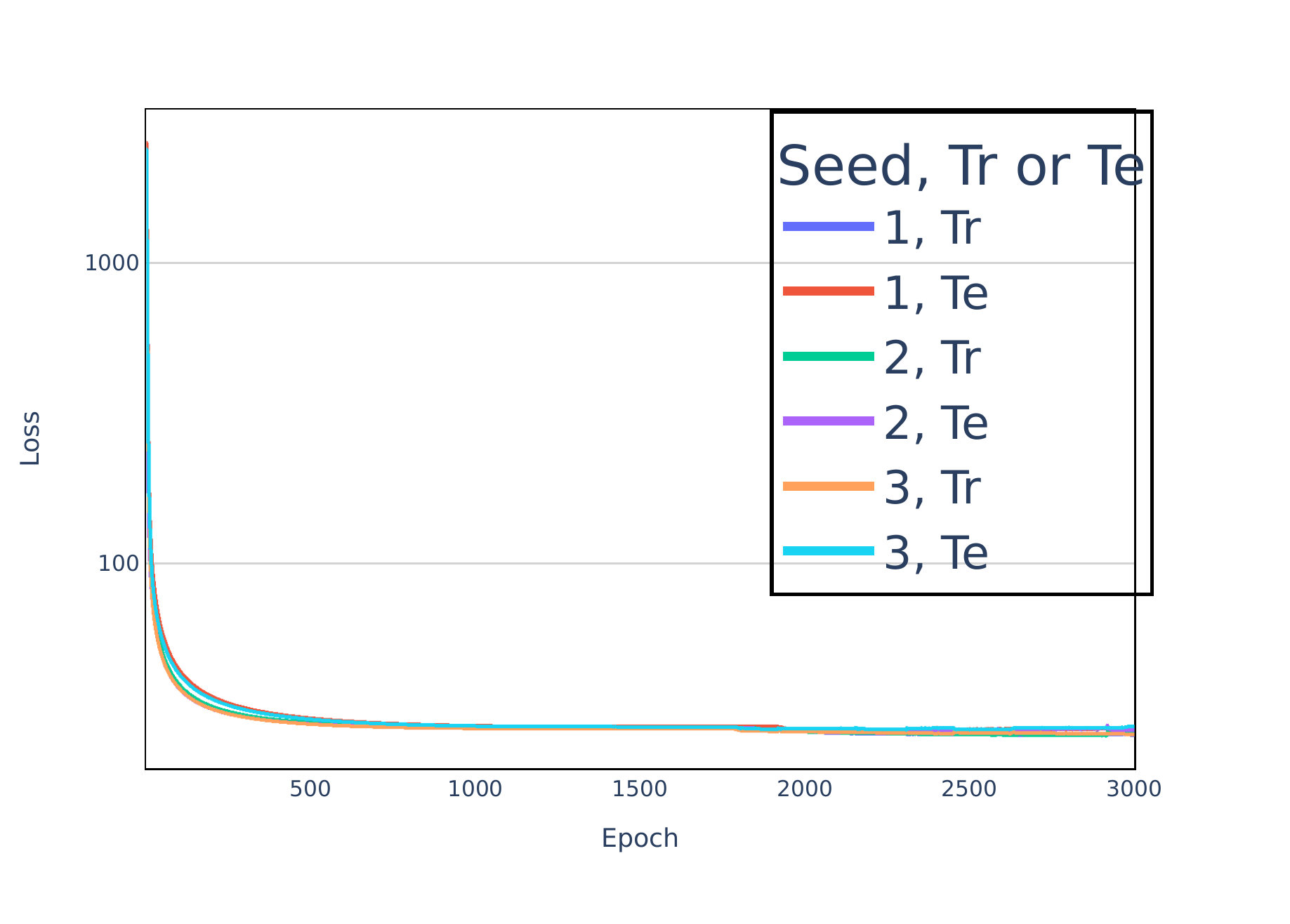}
		\caption{$\cluster$}
		\label{fig:cluster-loss}
	\end{subfigure}
	\begin{subfigure}{0.45\textwidth}
		\includegraphics[width=\textwidth]
		{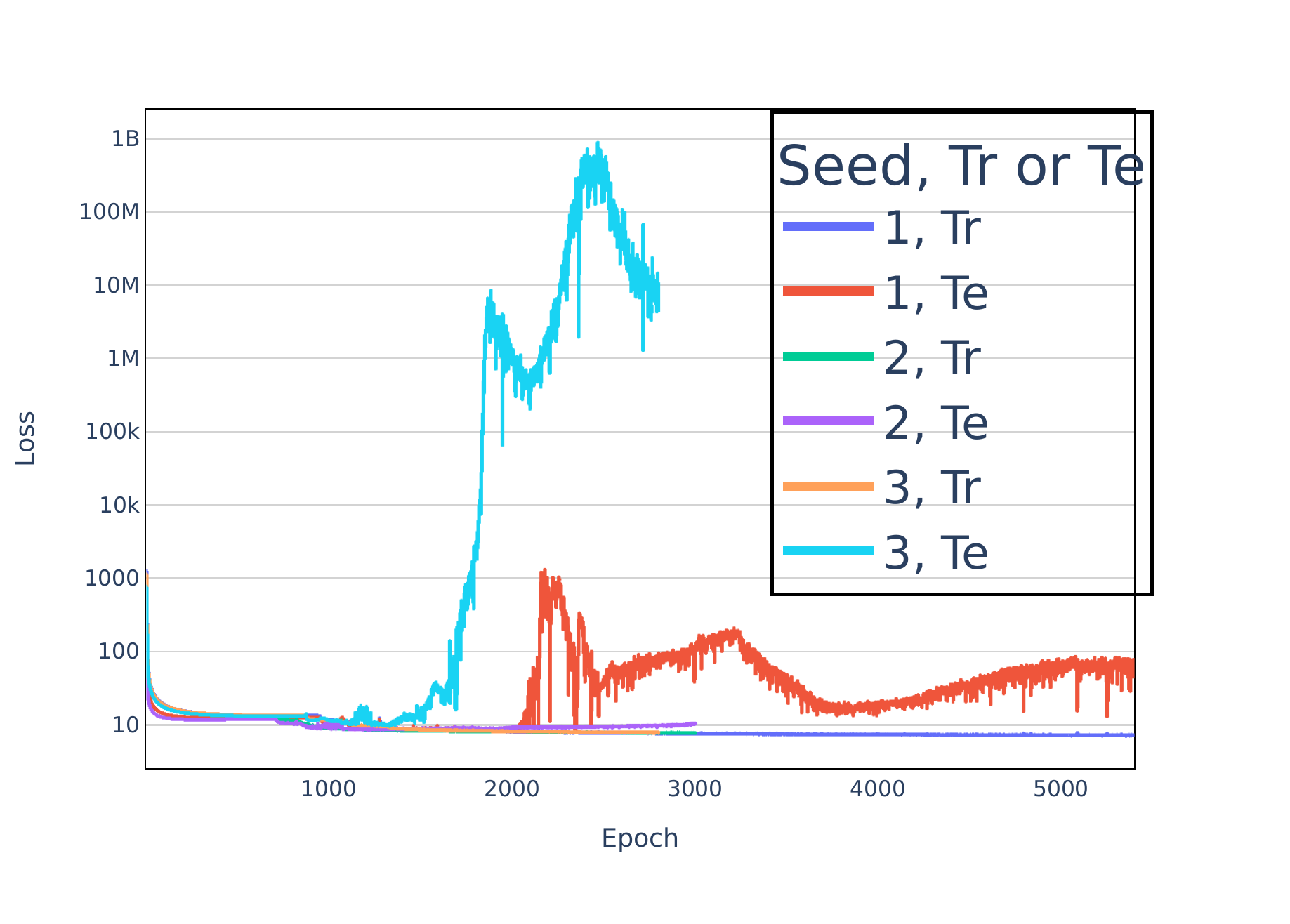}
		\caption{$\milkyo$}
		\label{fig:milkyo-loss}
	\end{subfigure}
	\begin{subfigure}{0.45\textwidth}
		\includegraphics[width=\textwidth]
		{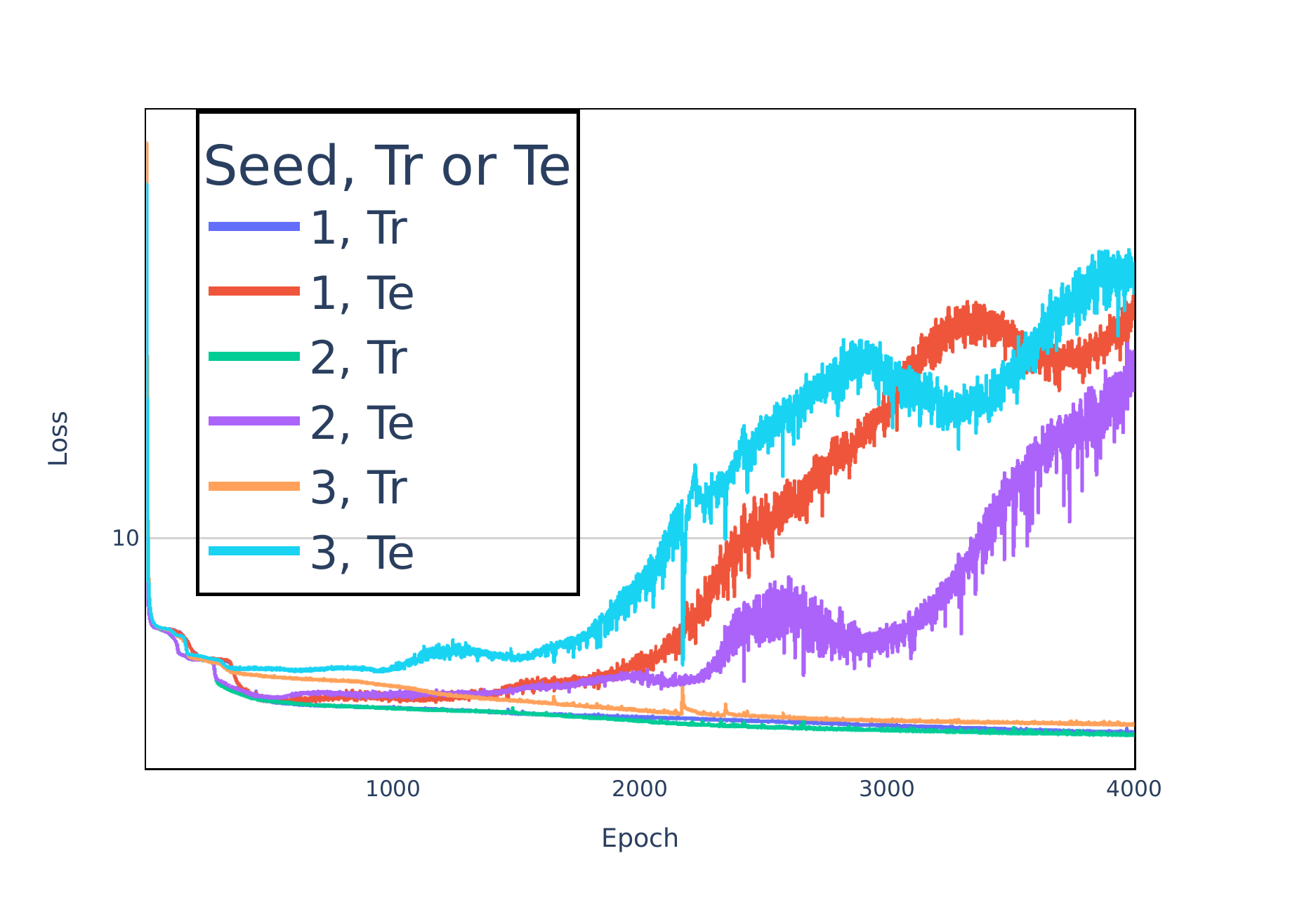}
		\caption{$\rbk$}
		\label{fig:rb-loss}
	\end{subfigure}
	\caption{
		Losses for $\hierd$, $\cluster$, $\milkyo$, and $\rbk$.
		The $y$-axes are log-scaled.
		The surges in later epochs of Fig.~\ref{fig:hierd-loss}, \ref{fig:milkyo-loss} and \ref{fig:rb-loss} were due to only a single or a few test programs out of $50$.}
	\label{fig:interpolation-losses-appendix}
\end{figure*}

\section{Losses for $\hierd$, $\cluster$, $\milkyo$, and $\rbk$}
\label{appendix:interpol-losses}
Fig.~\ref{fig:interpolation-losses-appendix} shows the average training and test losses under three random seeds for $\hierd$, $\cluster$, $\milkyo$, and $\rbk$.
The later part of Fig.~\ref{fig:hierd-loss}, \ref{fig:milkyo-loss} and \ref{fig:rb-loss} shows cases where the test loss surges. This was when the loss of only
a few programs in the test set (of $50$ programs) became large. Even in this situation, the losses of the rest remained small.
We give analyses for $\cluster$ and $\rbk$ separately in \S\ref{appendix:multimodality}.

% [GC: Removed this section and the figure, since the corresponding results was a bug.]
%\section{Program in $\hierd$ in moments estimation}
%\todo{Remove this section and Fig.~\ref{fig:hierd-program}.}
%\label{appendix:hierd-moments}

%Fig.~\ref{fig:hierd-program} shows the test program from $\hierd$ that were used in our moments estimation
%in the ``Application to Moments Estimation'' part of \S\ref{sec:empirical}.

%\begin{figure}
%	\hrule
%	\begin{align*}
%	& m_{z_1} := 4.6;\ v_{z_1} := 4045;\ v_{z_2} := 43.6;\ v_{z_3} := 8.4; \\
%	& m_{z_4} := 4.6;\ v_{z_4} := 23;\ d_1 := -2.5;\ d_2 := 0.07; \\
%	& v_{x_1} := 5.8;\ v_{x_2} := 31.4;\ o_1 := 60.8;\ o_2 := 69.3; \\
%	& z_1 \sim \mathcal{N}(m_{z_1},v_{z_1});\ z_2 \sim \mathcal{N}(z_1,v_{z_2});\ z_3 \sim \mathcal{N}(z_1,v_{z_3}); \\
%	& z_4 \sim \mathcal{N}(m_{z_4},v_{z_4}); \\
%	& t_1 := z_4 \times d_1;\ t_2 := z_2 + t_1;\ \code{obs}(\mathcal{N}(t_2,v_{x_1}),o_1); \\
%	& s_1 := z_4 \times d_2;\ s_2 := z_3 + s_1;\ \code{obs}(\mathcal{N}(s_2,v_{x_2}),o_2)
%	\end{align*}
%	\hrule
%	$\,$
%	\caption{Probabilistic program in $\hierd$ that was used in our moments evaluation.
%		The variables $z_{1:4}$ are latent.}
%	\label{fig:hierd-program}
%\end{figure}

\begin{figure}[!ht]
	\centering
	\includegraphics[width=0.45\textwidth]{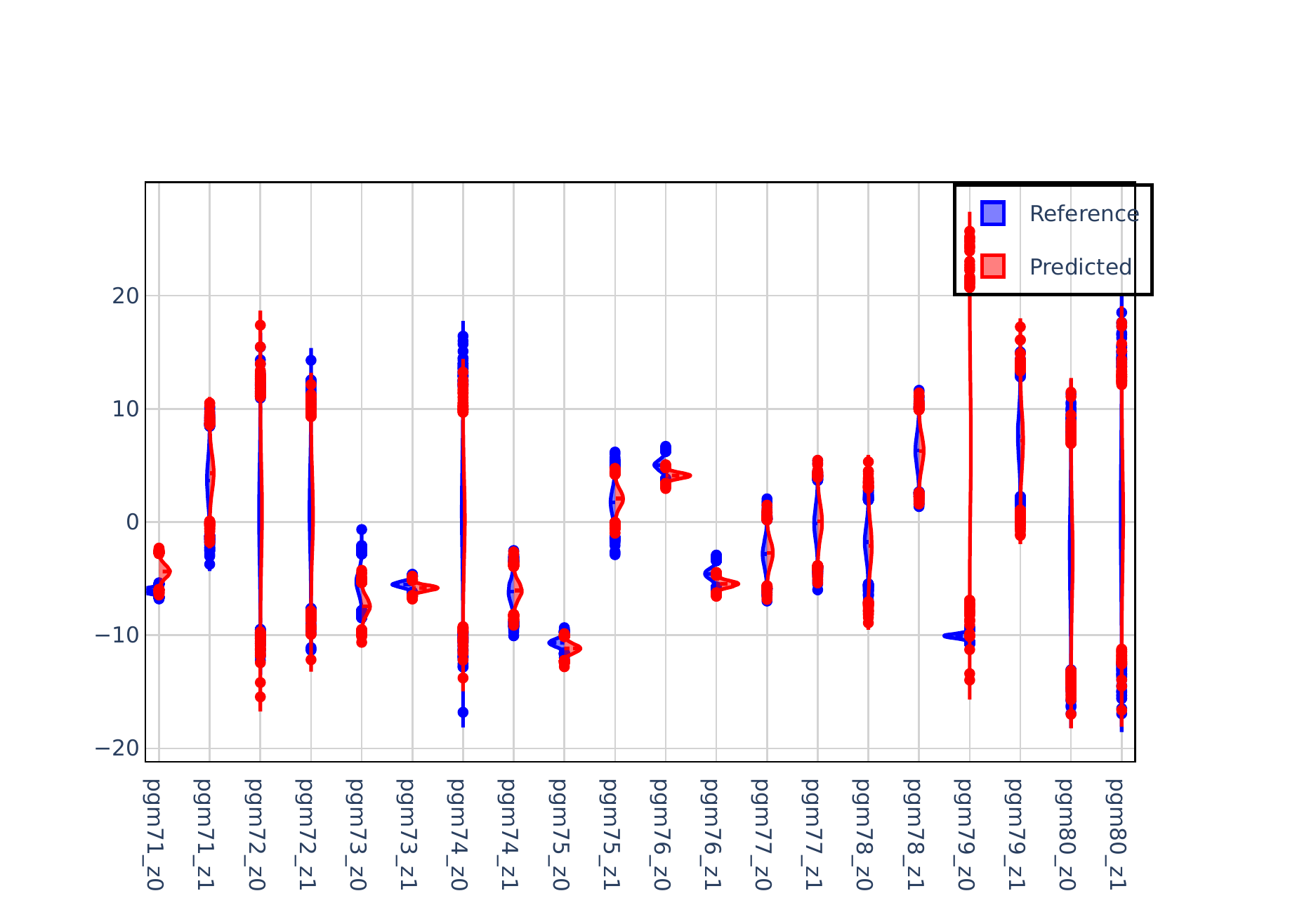}
	\caption{Comparisons of reference and predicted marginal posteriors for $10$ programs in the $\rbk$ test set.}
	\label{fig:rb}
\end{figure}

\section{Multimodal posteriors: $\cluster$ and $\rbk$}
\label{appendix:multimodality}
The $\cluster$ and $\rbk$ classes in \S\ref{sec:interpol} posed another challenge: the models often had multimodal posteriors, and it was
significantly harder for our meta-algorithm to learn an optimal inference algorithm. To make the evaluation partially feasible for $\rbk$,
we changed
two parts of our meta-algorithm slightly, as well as increasing the size of the test set from $50$ to $100$. First, we used importance samples instead of
samples by HMC, which often failed to converge, to learn an inference algorithm.
%we used an importance-sampling
%variant of the gradient estimator for our meta-learning objective so that the parameters for our inference algorithm can be optimised with
%importance samples, instead of posterior (i.e., HMC) samples, which are difficult to obtain in this case due to the multimodality issue.
Second, our random program generator placed some restriction on the programs it generated
(e.g., by using tight boundaries on some model parameters), guided by the analysis of the geometry of the Rosenbrock function \citep{pagani2019n}.
Consequently, HMC (with 500K samples after 50K warmups) failed to converge for only one fifth of the test programs.

Fig.~\ref{fig:rb} shows the similar comparison plots between reference and predicted marginal posteriors for $10$ test programs of the $\rbk$ type, 
after 52.4K epochs.
Our inference algorithm computed the posteriors precisely for most of the programs except two ($\mathsf{pgm75}$ and $\mathsf{pgm79}$) with significant 
multimodality. The latent variable
$\mathsf{pgm75\_z0}$ had at least two modes at around $-10$ (visible in the figure) and around $10$ (hidden in the figure)\footnote{The blue reference plots were drawn using an HMC chain, but the HMC chain got stuck in the mode around $-10$ for this variable.}. Our inference algorithm
showed a mode-seeking behavior for this latent variable. Similarly, the variable $\mathsf{pgm79\_z0}$ had
at least two modes in the similar domain region (one shown and one hidden), but this time our inference algorithm showed a 
mode-covering behavior.

The multimodality issue raises two questions. First, how can our meta-algorithm generate samples from the posterior more effectively
so that it can optimise the inference algorithm for classes of models with multimodal posteriors? For example, our current results for $\cluster$ 
suffer %are not free 
from the fact that the samples used in the training are often biased (i.e., only from a single mode of the posterior). 
One possible direction would be to use multiple Markov
chains simultaneously and apply ideas from the mixing-time research. Second, how can our white-box inference algorithm catch more information
from the program description and find non-trivial properties that may be useful for computing the posterior distributions having multiple modes?
We leave the answers for future work.

\begin{figure*}[t]
	\centering
	\begin{subfigure}{0.45\textwidth}
		\includegraphics[width=\textwidth]
		{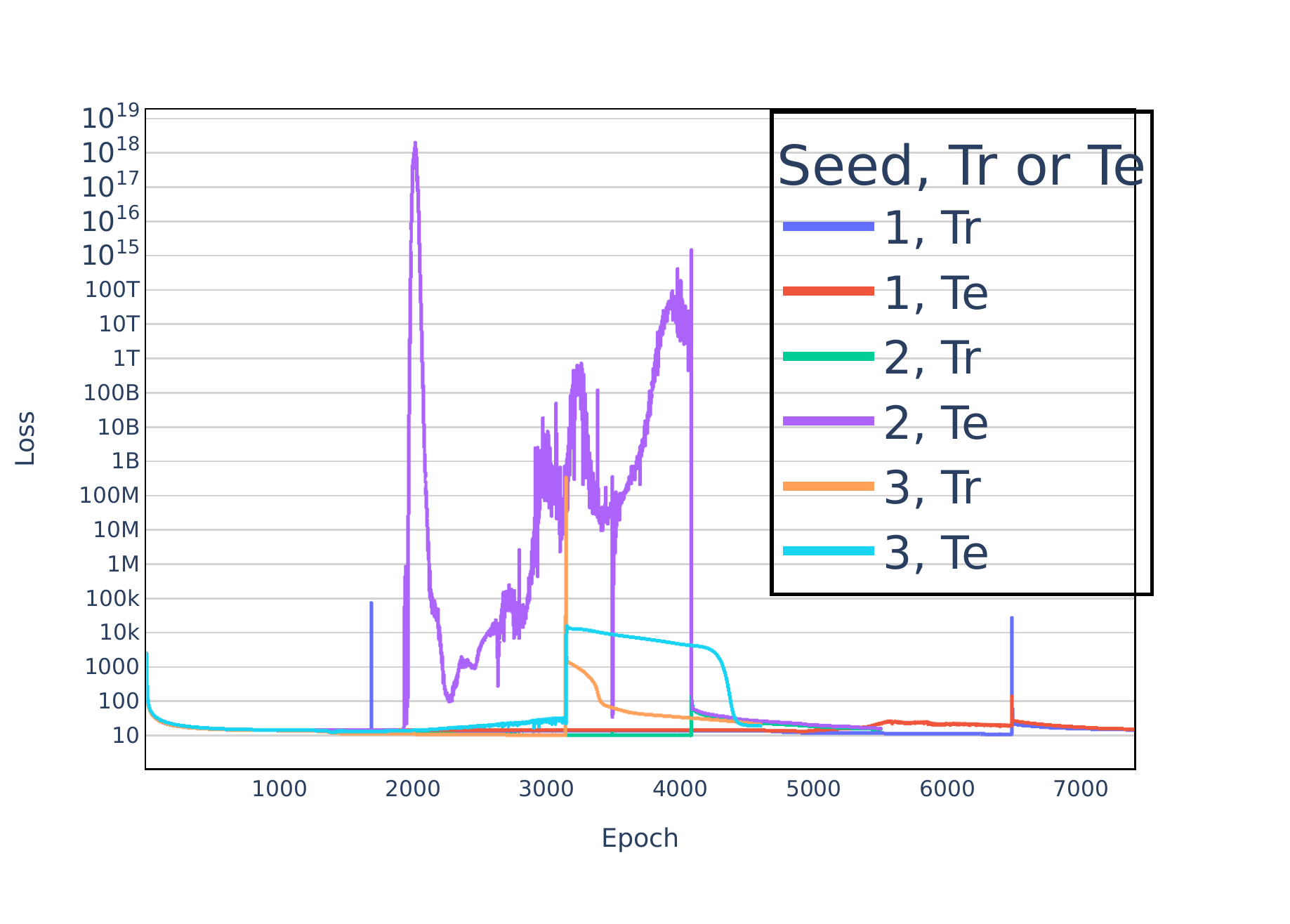}
		\caption{To 4th dep. graph.}
		%\label{}
	\end{subfigure}
	\begin{subfigure}{0.45\textwidth}
		\includegraphics[width=\textwidth]
		{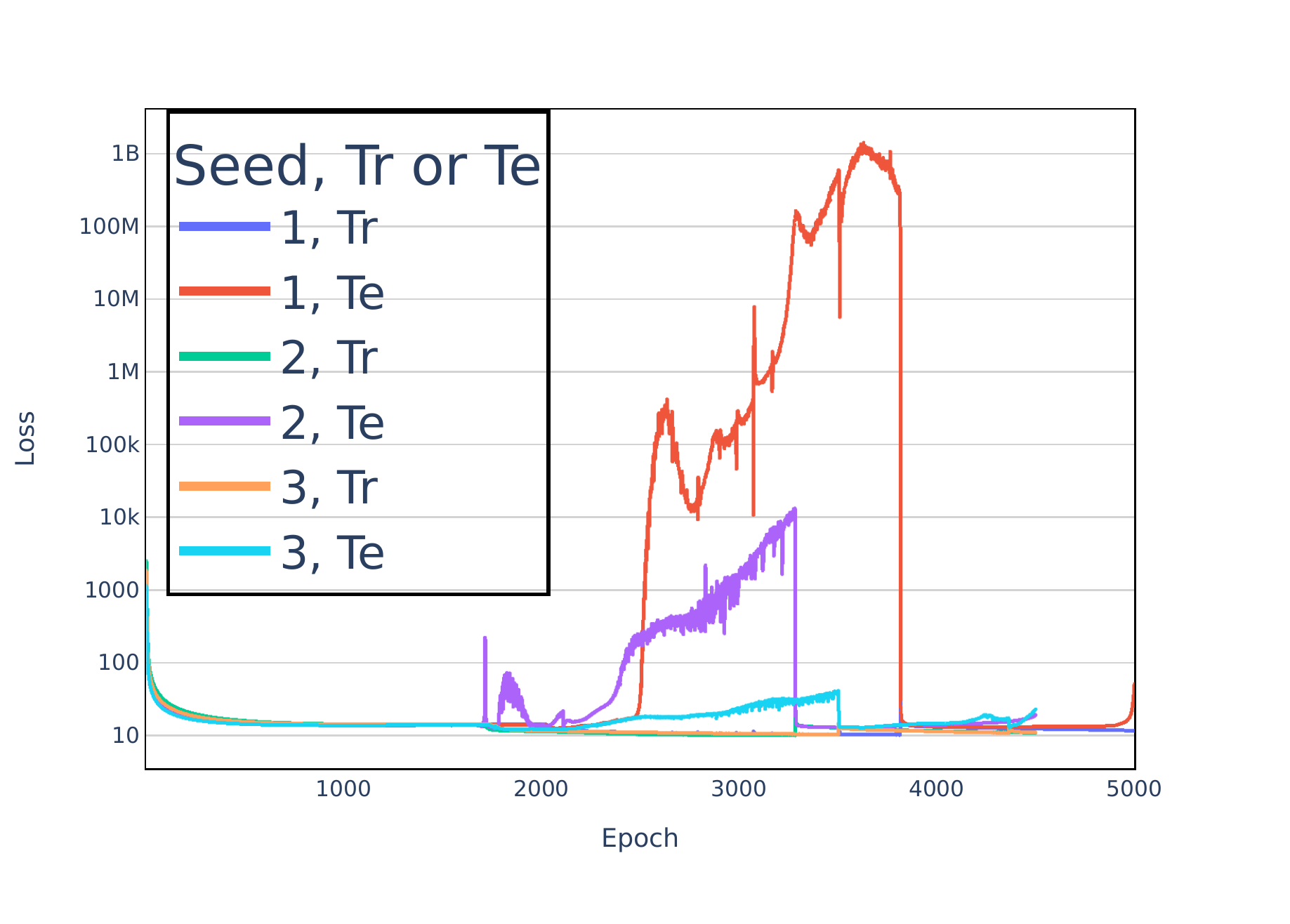}
		\caption{To 1st $\nl$ position.}
		%\label{}
	\end{subfigure}
	\begin{subfigure}{0.45\textwidth}
		\includegraphics[width=\textwidth]
		{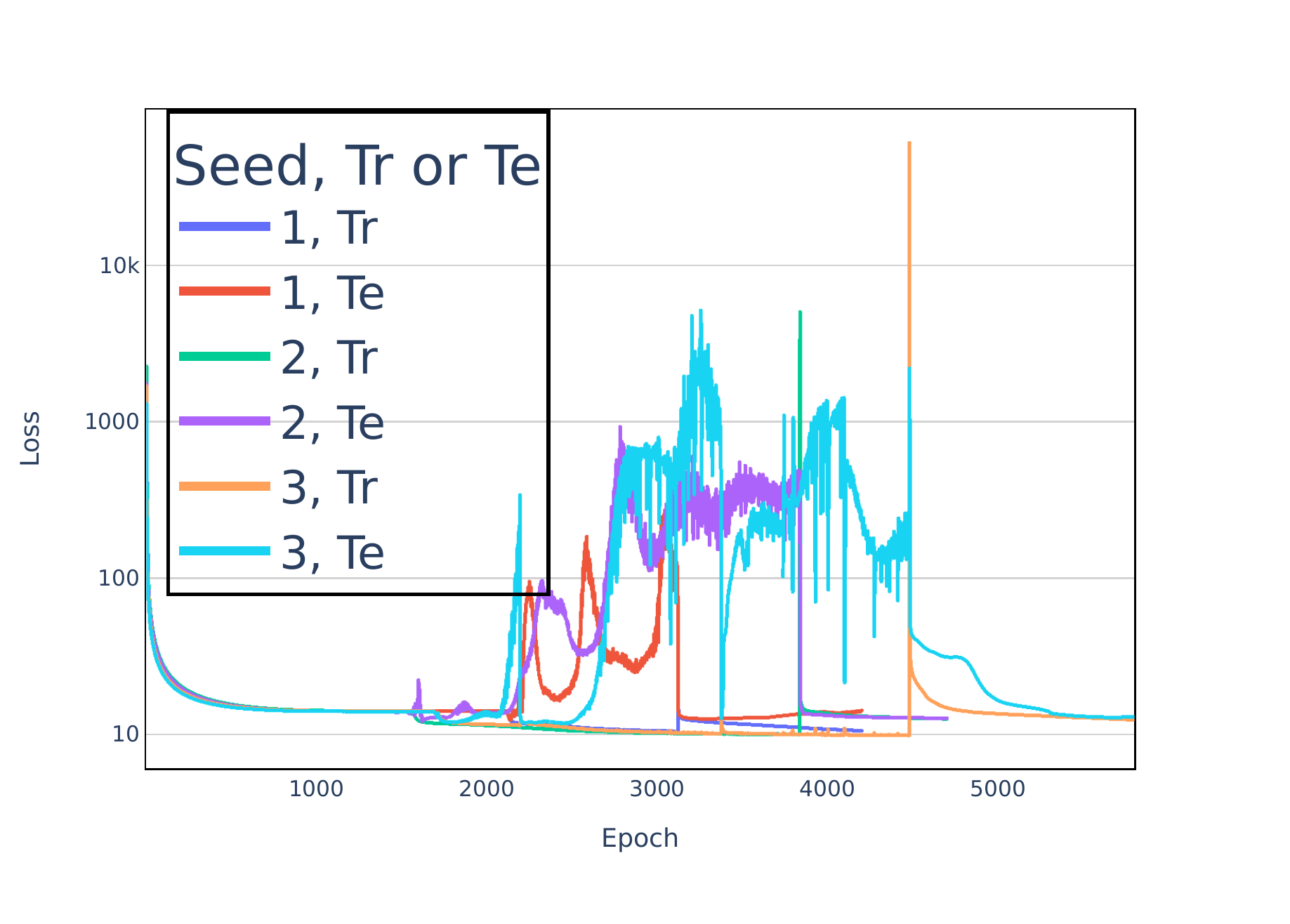}
		\caption{To 2nd $\nl$ position.}
		%\label{}
	\end{subfigure}
	\begin{subfigure}{0.45\textwidth}
		\includegraphics[width=\textwidth]
		{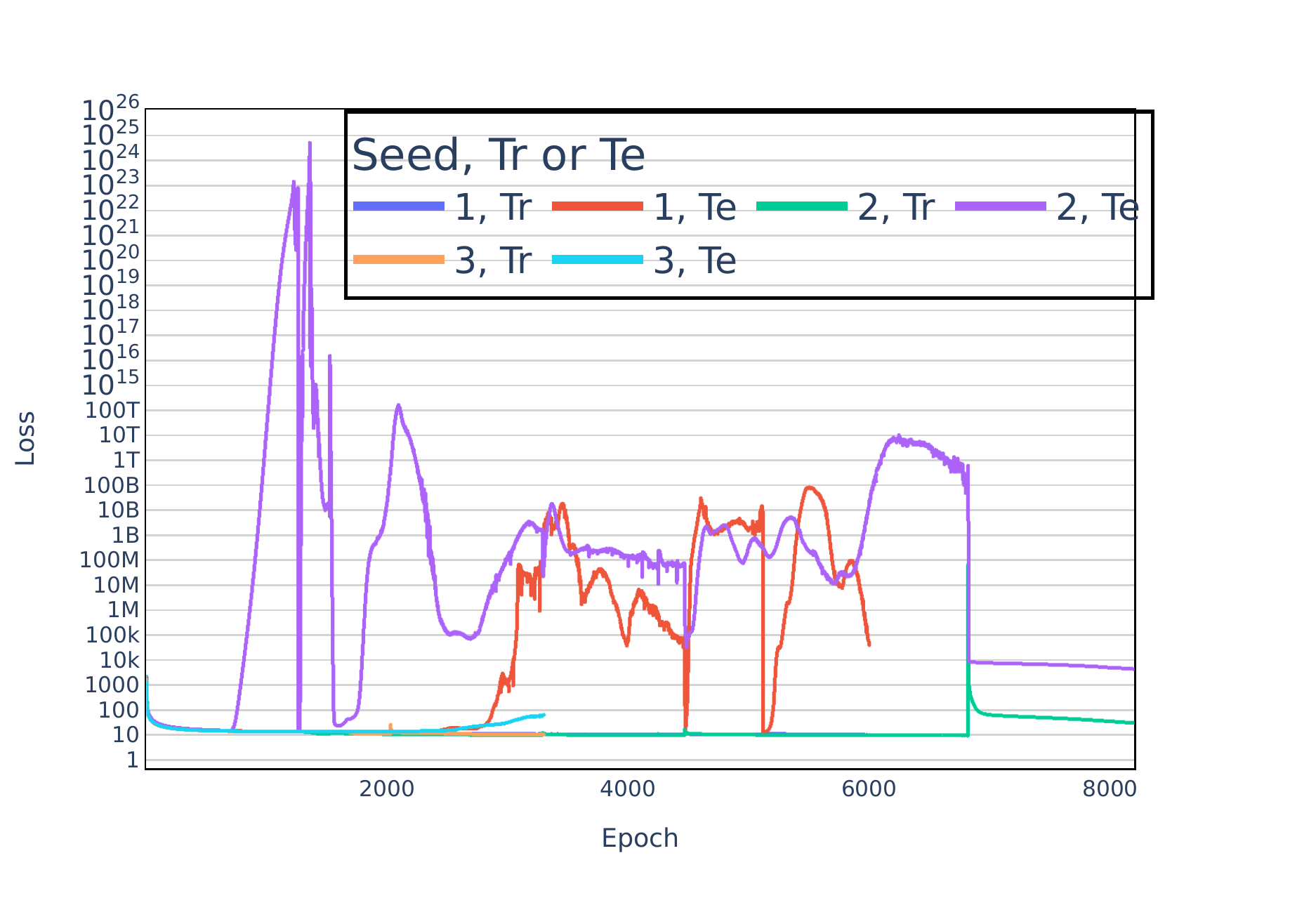}
		\caption{To 3rd $\nl$ position.}
		%\label{}
	\end{subfigure}
	\caption{
		Average training and test losses for generalisation to the last (4th) dependency graph and to all three positions of the $\nl$ variable in $\ext1$. The y-axes are log-scaled.
	}
	\label{fig:ext1-losses-appendix}
\end{figure*}

\section{Training and test losses for the other cases in $\ext1$}
\label{appendix:ext1-losses}
Fig.~\ref{fig:ext1-losses-appendix} shows the average training and test losses in the $\ext1$ experiment runs (under three different random seeds) for 
generalisation to the last (4th) dependency graph and to all three positions of the $\nl$ variable.

\begin{figure*}[t]
	\centering
	\begin{subfigure}{0.45\textwidth}
		\includegraphics[width=\textwidth]
		{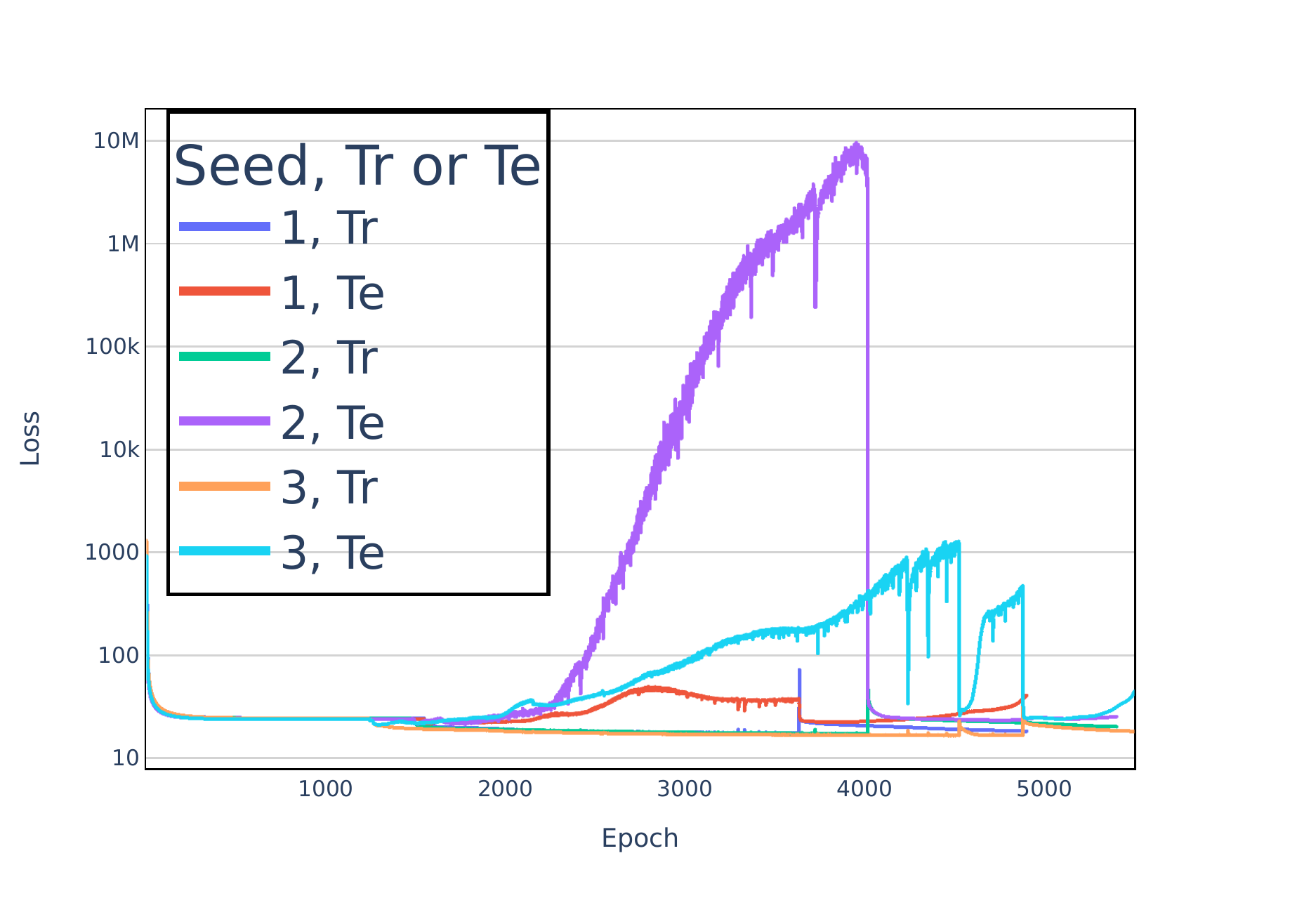}
		\caption{To $4$th dep. graph.}
		%\label{}
	\end{subfigure}
	\begin{subfigure}{0.45\textwidth}
	\includegraphics[width=\textwidth]
	{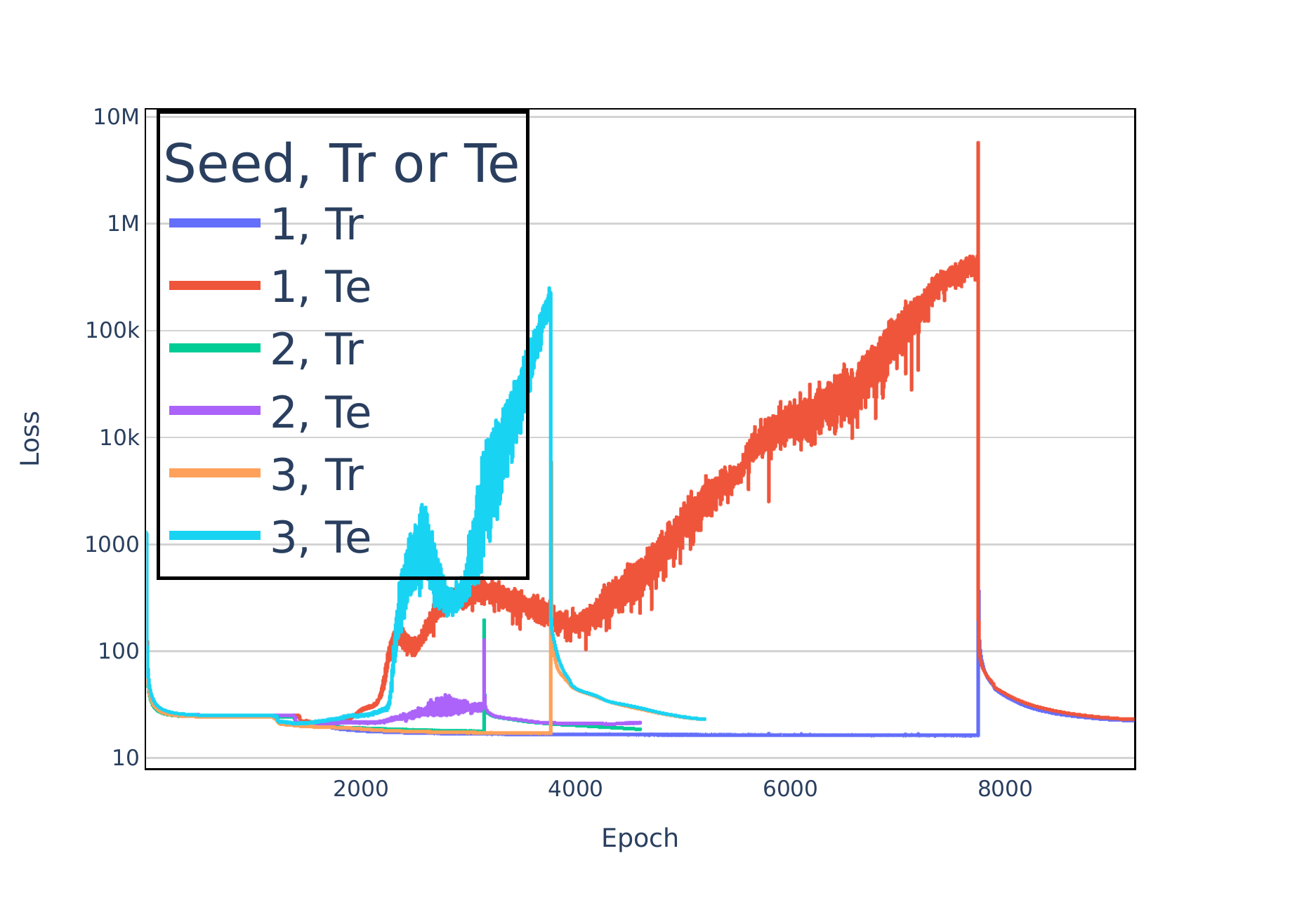}
	\caption{To $5$th dep. graph.}
	%\label{}
	\end{subfigure}
	\caption{
		Average training and test losses for generalisation to the $4$th and $5$th dependency graphs in $\ext2$. The y-axes are log-scaled.
	}
	\label{fig:ext2-losses-appendix}
\end{figure*}

\section{Training and test losses for the other cases in $\ext2$}
\label{appendix:ext2-losses}
Fig.~\ref{fig:ext2-losses-appendix} shows the average training and test losses in the $\ext2$ experiment runs (under three different random seeds) for 
generalisation to the $4$th and $5$th dependency graphs.

\begin{figure}[!ht]
	\hrule
	\begin{align*}
	& \mathit{a}:=3.93;\ \mathit{b}:=348.16;\ \mathit{c}:=57.5;\ \mathit{d}:=14.04;\ \mathit{e}:=40.34; \\
	& z_1 \sim \mathcal{N}(a,b);\ z_2 \sim \mathcal{N}(z_1,c);\ z_3 := \mm(z_1); \\
	& \code{obs}(\mathcal{N}(z_2,d), 53.97);\ \code{obs}(\mathcal{N}(z_3,e), 0.12)
	\end{align*}
	\hrule
	$\,$
	\caption{The program that is reported in \S\ref{sec:mulmod}, written in our probabilistic programming language.}
	\label{fig:mulmod-pgm19-program}
\end{figure}

\section{Quantified accuracy of predicted posteriors}
\label{appendix:predicted-posteriors-accuracy}
For accuracy, it would be ideal to report $\KL[p {||} q]$, where $p$ is the fully joint target posterior and $q$ is the predicted distribution. It is,
however, hard to compute this quantity since often we cannot compute the density of $p$. One (less convincing) alternative is to compute
$\KL[p'(z) {||} q(z)]$ for a latent variable $z$ where $p'(z)$ is the best Gaussian approximation (i.e., the best approximation using the mean and standard
deviation) for the true marginal posterior $p(z)$, and average the results over all the latent variables of interest.
We computed $\KL[p'(z) {||} q(z)]$ for the test programs from $\ext1$ and $\ext2$ in \S\ref{sec:extrapol}, and for the three from $\mulmod$ that are reported
in \S\ref{sec:mulmod}.

For $\ext1$ and $\ext2$, we measured the average $\KL[p'(z) {||} q(z)]$ over all the latent variables $z$ in the test programs. For instance, if there were 
$90$ test programs and each program had three latent variables, we averaged $90 \times 3 = 270$ $\KL$ measurements.
In an experiment run for $\ext1$ (which tested generalisation to an unseen dependency graph), the average $\KL$ was around $1.32$. In an experiment run for 
$\ext2$, the estimation was around $0.95$. When we replaced $q$ with a normal distribution that is highly flat (with mean $0$ and standard deviation $10$K), 
the estimation was $7.11$ and $7.57$, respectively. The results were similar in all the other experiment runs that were reported in \S\ref{sec:extrapol}.

For the three programs from $\mulmod$ in \S\ref{sec:mulmod}, $p'$ was the best Gaussian approximation whose mean and standard deviation were estimated by
the reference importance sampler (IS-ref), and $q$ was either the predicted marginal posterior by the learnt inference algorithm or the best Gaussian
approximation whose mean and standard deviation were estimated by HMC. The average $\KL$ was around $1.19$ when $q$ was the predicted posterior, while
the estimation was $40.9$ when $q$ was the best Gaussian approximation by HMC. The results demonstrate that the predicted posteriors were more accurate on
average than HMC at least in terms of $p'$.

\section{Program in \S\ref{sec:mulmod}}
\label{appendix:mulmod-pgm19}
Fig.~\ref{fig:mulmod-pgm19-program} shows the program that is reported in \S\ref{sec:mulmod}, written in our probabilistic programming language.

\section{Discussion of the cost of IS-pred vs. IS-prior}
\label{appendix:is-pred-cost}

Our approach (IS-pred) must scan the given program ``twice'' at test time, once for computing the proposal using the learnt neural networks
and another for running the importance sampler with the predicted proposal, while IS-prior only needs to scan the program once. Although it
may seem that IS-prior has a huge advantage in terms of saving the wall-clock time, our observation is that the effect easily disappears
as the sample size increases. In fact, going through the neural networks in our approach (i.e., the first scanning of the program) does not
depend on the sample size, and so its time cost remains constant given the program; the time cost was $0.6$ms for the reported test program 
($\mathsf{pgm19}$) in \S\ref{sec:mulmod}.

%\section{Limitations and future work}
%\label{appendix:limitations}

%Currently, a learnt inference algorithm in our work does not generalise to programs with different sizes~\citep{yan20execute}, e.g., from clustering models with
%two clusters to those with ten clusters. Each model class assumes a fixed number of variables, and the neural networks crucially exploit the assumption. Also, our meta-algorithm does not scale in practice. When applied to large programs, e.g., state-space models with a few hundred time steps, it cannot learn an
%optimal inference algorithm within a reasonable amount of time. Overcoming these limitations is a future work. Another direction that we are considering is to remove the strong independence assumption (via mean field Gaussian) on the 
%approximating distribution in our inference algorithm, and to equip the algorithm with the capability of generating an appropriate form of the approximation distribution with rich dependency structure, by, e.g., incorporating the ideas from
%\citet{ambrogioni2021automatic}. This direction is closely related to automatic guide generation in Pyro~\citep{bingham2018pyro}.

\end{document}